%% file: main.tex
\pdfoutput=1
\documentclass[10pt,twocolumn,letterpaper]{article}

\usepackage[pagenumbers]{cvpr} %

\input{packages.tex}

\usepackage[section]{placeins}

\begin{document}

\title{eDiff-I: Text-to-Image Diffusion Models with an Ensemble of Expert Denoisers}

\author{Yogesh Balaji, Seungjun Nah, Xun Huang, Arash Vahdat, Jiaming Song, Qinsheng Zhang,\\ Karsten Kreis, Miika Aittala, Timo Aila, Samuli Laine, Bryan Catanzaro, Tero Karras, Ming-Yu Liu\\
\\
NVIDIA Corporation\\
{\small \{ybalaji, snah, xunh, avahdat, jiamings, qinshengz, kkreis, maittala, taila, slaine, bcatanzaro, tkarras, mingyul\}@nvidia.com}}

\twocolumn[{%
\renewcommand\twocolumn[1][]{#1}%
\maketitle
    \input{figures/teaser/teaser.tex}
}]

\input{sections/abs}
\input{sections/intro}

\input{sections/related}
\input{sections/preliminary}

\input{sections/body}
\input{sections/expr}

\input{sections/conc}

{\small
\bibliographystyle{ieee_fullname}
\bibliography{egbib}
}
\input{sections/appendix}

\end{document}

%% file: packages.tex
\usepackage{graphicx}
\usepackage{amsmath}
\usepackage{amssymb}
\usepackage{booktabs}
\usepackage{color}
\usepackage{xcolor}
\usepackage{tabularx}
\usepackage{makecell}
\newcolumntype{Y}{>{\centering\arraybackslash}X}

\makeatletter
\@namedef{ver@everyshi.sty}{}
\makeatother
\usepackage{tikz}
\usepackage{pgfplots}
\usepackage{amsmath}
\usepackage{animate}
\usepackage{subcaption}
\usepackage{array}
\usepackage{multirow}
\usepackage{microtype}
\usepackage{enumitem}
\usepackage{xspace}
\usepackage{soul}
\usepackage{multicol}
\usepackage{colortbl}
\usepackage{tabularx,booktabs}
\usepackage{cancel}
\usepackage{float}
\usepackage{pifont}
\usepackage{adjustbox}
\usepackage{epigraph}
\usepackage{wrapfig}
\usepackage{textcomp}  %
\usepackage{scalerel}  %

\pgfplotsset{xtick style={draw=none}}
\pgfplotsset{ytick style={draw=none}}
\pgfplotsset{major grid style={gray!40}}
\pgfplotsset{every axis plot/.style={thick, mark size=1.0pt}}
\pgfplotsset{legend image code/.code={\draw[mark repeat=2, mark phase=2] plot coordinates {(0cm, 0cm) (0.2cm, 0cm) (0.4cm, 0cm)};}} %
\definecolor{C0}{rgb}{0.121569, 0.466667, 0.705882}
\definecolor{C1}{rgb}{1.000000, 0.498039, 0.054902}
\definecolor{C2}{rgb}{0.172549, 0.627451, 0.172549}
\definecolor{C3}{rgb}{0.839216, 0.152941, 0.156863}
\definecolor{C4}{rgb}{0.580392, 0.403922, 0.741176}
\newcommand{\plothh}{}\newcommand{\plotvv}{}

\newcommand{\ours}{\texttt{eDiff-I}\xspace}
\newcommand{\configa}{\texttt{eDiff-I-Config-A}\xspace}
\newcommand{\configb}{\texttt{eDiff-I-Config-B}\xspace}
\newcommand{\configc}{\texttt{eDiff-I-Config-C}\xspace}
\newcommand{\configd}{\texttt{eDiff-I-Config-D}\xspace}
\newcommand{\pww}[0]{\textit{paint-with-words}}
\usepackage[pagebackref,breaklinks,colorlinks]{hyperref}

\newcommand{\xx}{\boldsymbol{x}}
\newcommand{\ee}{\boldsymbol{e}}
\newcommand{\dx}{\mathrm{d}\xx}

\newcommand{\dsigma}{\mathrm{d}\sigma}
\newcommand{\nablaxx}{\nabla_{\hspace{-0.5mm}\xx}}
\newcommand{\normal}{\mathcal{N}}

\newcommand{\smax}{\sigma_\text{max}}
\newcommand{\smin}{\sigma_\text{min}}
\newcommand{\boldzero}{\mathbf{0}}
\newcommand{\boldI}{\mathbf{I}}
\newcommand{\sdata}{\sigma_\text{data}}
\newcommand{\sast}{\sigma^{*\!}}
\newcommand{\xxclean}{\xx_\text{clean}}
\newcommand{\expect}{\mathbb{E}}
\newcommand{\pmean}{P_\text{mean}}
\newcommand{\pstd}{P_\text{std}}
\newcommand{\pdata}{p_\text{data}}
\newcommand{\beps}{\boldsymbol{\epsilon}}
\newcommand{\s}{\hphantom{0}}
\newcommand{\sdot}{\hphantom{.}}

\newcolumntype{M}[1]{>{\centering\arraybackslash}m{#1}}

\usepackage[capitalize]{cleveref}
\crefname{section}{Sec.}{Secs.}
\Crefname{section}{Section}{Sections}
\Crefname{table}{Table}{Tables}
\crefname{table}{Tab.}{Tabs.}

%% file: figures/teaser/teaser.tex
    \setlength{\tabcolsep}{0.5pt}
    \renewcommand{\arraystretch}{0.5}
    \begin{tabularx}{\textwidth}{M{0.331\textwidth} M{0.331\textwidth} M{0.331\textwidth}}
        \includegraphics[width=0.331\textwidth]{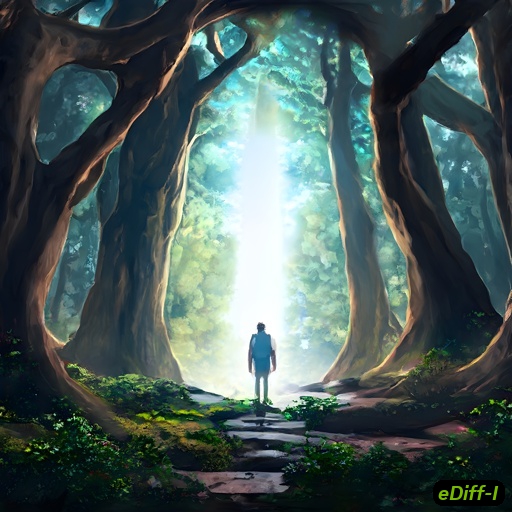} &
        \includegraphics[width=0.331\textwidth]{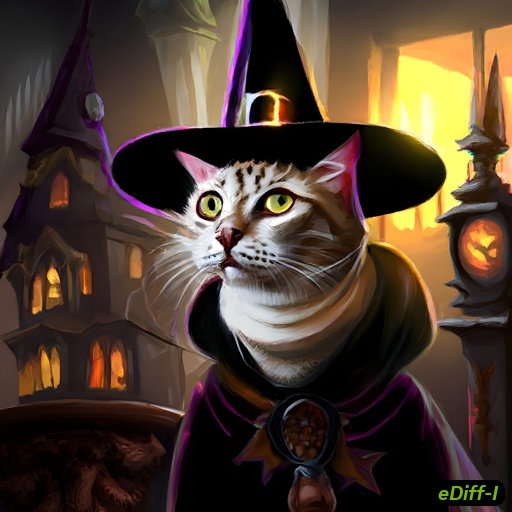} &
        \includegraphics[width=0.331\textwidth]{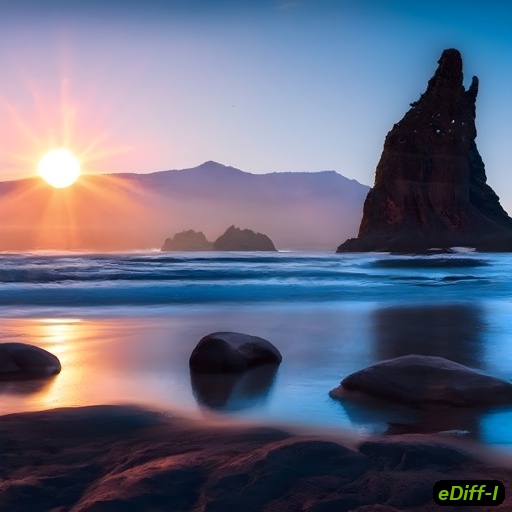} \\
    \end{tabularx}
    \begin{tabularx}{\textwidth}{M{0.30\textwidth} c M{0.30\textwidth} c M{0.30\textwidth}}
        \emph{A highly detailed digital painting of a portal in a mystic forest with many beautiful trees. A person is standing in front of the portal.} & \quad \quad \quad & \emph{A highly detailed zoomed-in digital painting of a cat dressed as a witch wearing a wizard hat in a haunted house, artstation.} & \quad \quad \quad& \emph{An image of a beautiful landscape of an ocean. There is a huge rock in the middle of the ocean. There is a mountain in the background. Sun is setting.} \\
    \end{tabularx}

    \begin{minipage}{\textwidth}
        {
        \setlength{\tabcolsep}{0.5pt}
        \begin{tabularx}{0.45\textwidth}{M{0.145\textwidth} M{0.29\textwidth}}
            \vspace{-1.45em}\includegraphics[width=0.145\textwidth]{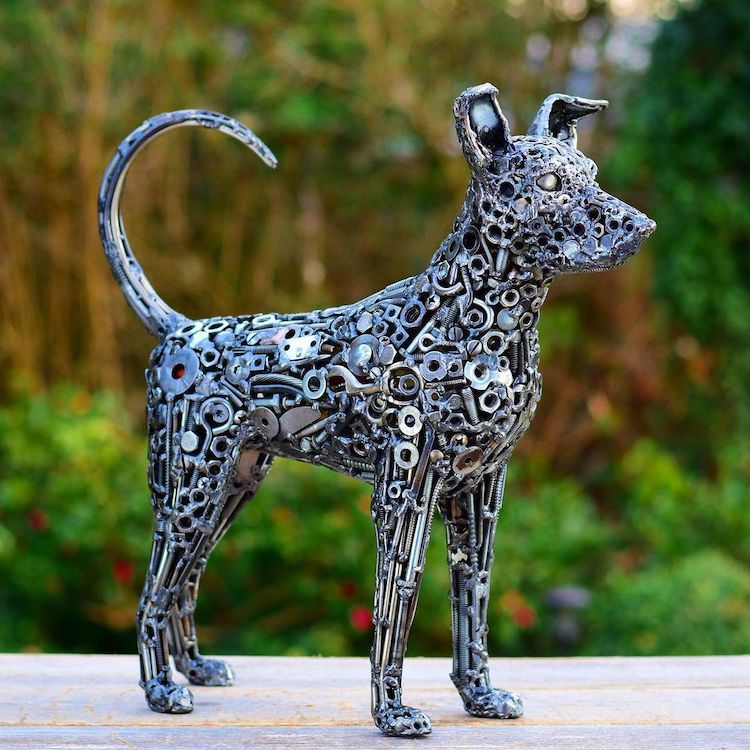} & \multirow{3}{*}[3.8em]{\centering \includegraphics[width=0.29\textwidth]{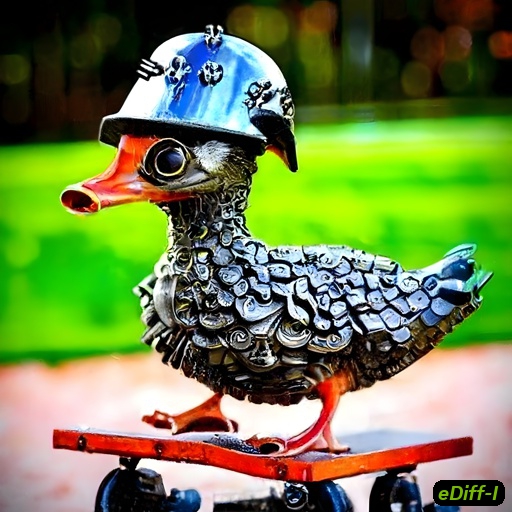}}\vspace{0.45em} \\
            {\footnotesize Style reference} & \\
            \vspace{0.5em}\emph{A photo of a duckling wearing a medieval soldier helmet and riding a skateboard.} & \\
        \end{tabularx}
        }
        \hspace{-6mm}
        {
        \setlength{\tabcolsep}{0pt}
        \begin{tabularx}{0.56\textwidth}{M{0.56\textwidth}}
            \includegraphics[width=0.273\textwidth]{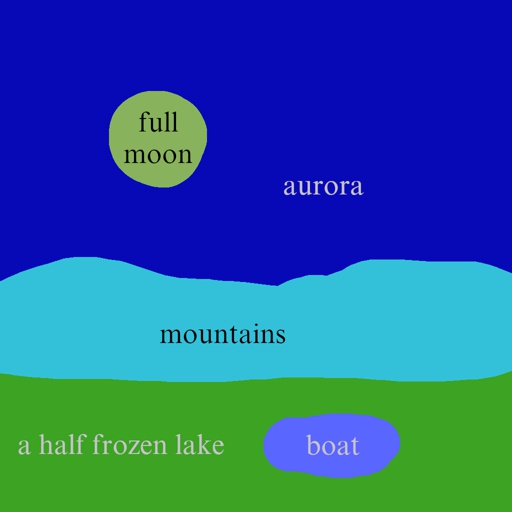} 
            \includegraphics[width=0.273\textwidth]{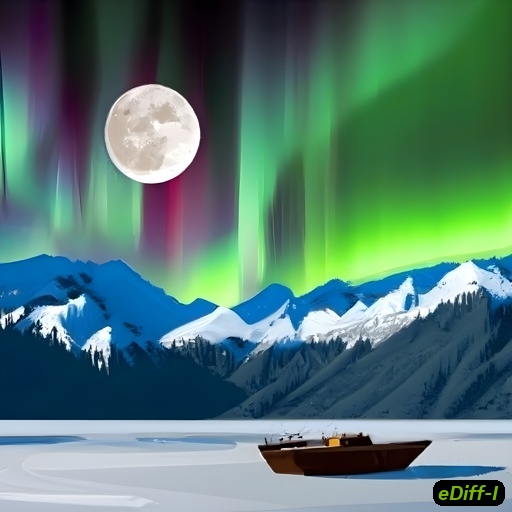} \\
            \emph{A digital painting of a half-frozen lake near mountains under a full moon and aurora. A boat is in the middle of the lake. Highly detailed.}            
        \end{tabularx}
        }        
    \end{minipage}
    
    \captionof{figure}{Example results and capabilities from our proposed method, \ours. The first row shows that \ours can faithfully turn complex input text prompts into artistic and photorealistic images. In the second row, we first show that \ours can combine the text input and a reference image for generating the target output image, where the reference image can be conveniently used to represent a style or concept that is difficult to describe in words, but a visual example exists. We also show the paint-by-word capability of \ours, where phrases in the input text can be painted on a canvas to control the specific layout of objects described in the input text. The \textit{paint-with-words} capability complements the text-to-image capability and provides an artist with more control over the generation outputs. More results are available at \url{https://deepimagination.cc/eDiff-I/}}
    \label{fig:teaser}
    \vspace{6mm}

%% file: sections/abs.tex
\begin{abstract}
Large-scale diffusion-based generative models have led to breakthroughs in text-conditioned high-resolution image synthesis, demonstrating complex text comprehension and outstanding zero-shot generalization. Starting from random noise, such text-to-image diffusion models gradually synthesize images in an iterative fashion while conditioning on text prompts. We find that their synthesis behavior qualitatively changes throughout this process: Early in sampling, generation strongly relies on the text prompt to generate text-aligned content, while later, the text conditioning is almost entirely ignored, and the task changes to producing outputs of high visual fidelity. This suggests that sharing model parameters throughout the entire generation process, the standard practice in the literature, may not be ideal to best capture these distinctly different modes of the generation process. Therefore, in contrast to existing works, we propose to train an ensemble of text-to-image diffusion models specialized for different synthesis stages. To maintain training efficiency, we initially train a single model, which is then progressively split into specialized models that are further trained for the specific stages of the iterative generation process. Our ensemble of diffusion models, called \ours, results in improved text alignment while maintaining the same inference computation cost and preserving high visual quality, outperforming previous large-scale text-to-image diffusion models on the standard benchmark. In addition, we train our model to exploit a variety of embeddings for conditioning, including the T5 text, CLIP text, and CLIP image embeddings. We show that these different embeddings lead to different image formation behaviors. Notably, the CLIP image embedding allows an intuitive and instant way of transferring the style of a reference image to the target text-to-image output. Lastly, we show a technique that enables \ours's ``paint-with-words'' capability. A user can select the word in the input text and paint it in a canvas to control the output, which is very handy for crafting the desired image in mind. The project page is available at \url{https://deepimagination.cc/eDiff-I/}
\end{abstract}

%% file: sections/intro.tex
\section{Introduction}\label{sec:intro}

Diffusion models~\cite{sohl-dickstein2015deep,ho2020denoising,song2020score} that generate images through iterative denoising, as visualized in Figure~\ref{fig:diffusion_models}, are revolutionizing the field of image generation. They are the core building block of recent text-to-image models, which have demonstrated astonishing capability in turning complex text prompts into photorealistic images, even for unseen novel concepts~\cite{ramesh2022hierarchical,saharia2022photorealistic,rombach2022high}. These models have led to the development of numerous interactive tools and creative applications and turbocharged the democratization of content creation.

\input{figures/method/motivation}

Arguably, this success is attributed largely to the great scalability of diffusion models because the scalability provides a clear pathway for practitioners to translate larger model capacity, compute, and datasets into better image generation quality. This is a great reminder of the bitter lessons~\cite{bitter_lesson}, which observe that a scalable model trained on vast data with massive computing power in the long term often outperforms its handcrafted specialized counterparts. A similar trend has been observed previously in natural language modeling~\cite{kaplan2020scaling}, image classification~\cite{hestness2017deep}, image reconstruction~\cite{klug2022scaling}, and autoregressive generative modeling~\cite{henighan2020scaling}.

We are interested in further scaling diffusion models in terms of model capability for the text-to-image generation task. We first note that simply increasing the capacity by using deeper or wider neural networks for each denoising step will negatively impact the test-time computational complexity of sampling, since sampling amounts to solving a reverse (generative) differential equation in which a denoising network is called many times. We aim to achieve the scaling goal without incurring the test-time computational complexity overhead.

Our key insight is that text-to-image diffusion models exhibit an intriguing temporal dynamic during generation. At the early sampling stage, when the input data to the denoising network is closer to the random noise, the diffusion model mainly relies on the text prompt to guide the sampling process. As the generation continues, the model gradually shifts towards visual features to denoise images, mostly ignoring the input text prompt as shown in Figure~\ref{fig:attention_maps} and \ref{fig:prompt_switching}.

\input{figures/method/attention_maps}
\input{figures/method/prompt_switching}

Motivated by this observation, we propose to increase the capacity of diffusion models by training an ensemble of \textit{expert denoisers}, each specialized for a particular stage in the generation process. While this does not increase the computational complexity of sampling per time step, it increases the training complexity since different denoising models should be trained for different stages. To remedy this, we propose pre-training a shared diffusion model for all stages. We then use this pre-trained model to initialize specialized models and finetune them for a smaller number of iterations. This scheme leads to state-of-the-art text-to-image generation results on the benchmark dataset.

We also explore using an ensemble of pretrained text encoders to provide inputs to our text-to-image model. We use both the CLIP text encoder\cite{radford2021learning}, which is trained to align text embedding to the corresponding image embedding, and the T5 text encoder\cite{raffel2020exploring}, which is trained for the language modeling task. Although prior works~\cite{nichol2021glide, saharia2022photorealistic, ramesh2022hierarchical, rombach2022high,stablediffusion1-4} have used these two encoders, they have not been used together in one model. As these two encoders are trained with different objectives, their embeddings favor formations of different images with the same input text. While CLIP text embeddings help determine the global look of the generated images, the outputs tend to miss the fine-grained details in the text. In contrast, images generated with T5 text embeddings alone better reflect the individual objects described in the text, but their global looks are less accurate. Using them jointly produces the best image-generation results in our model. In addition to text embeddings, we train our model to leverage the CLIP image embedding of an input image, which we find useful for style transfer. We call our complete model \textit{ensemble diffusion for images}, abbreviated \ours.

While text prompts are effective in specifying the objects to be included in the generated images, it is cumbersome to use text to control the spatial locations of objects. We devise a training-free extension of our model to allow \pww{}, a controllable generation approach with our text-to-image model that allows the user to specify the locations of specific objects and concepts by scribbling them in a canvas. The result is an image generation model that can take both texts and semantic masks as inputs to better assist users in crafting the perfect images in their minds.

\paragraph{Contributions} The main contributions of our work are
\begin{enumerate}
    \item Based on the observation that a text-to-image diffusion model has different behaviors at different noise levels, we propose the ensemble-of-expert-denoisers design to boost generation quality while maintaining the same inference computation cost. The expert denoisers are trained through a carefully designed finetuning scheme to reduce the training cost.
    \item We propose to use an ensemble of encoders to provide input information to the diffusion model. They include the T5 text encoder, the CLIP text encoder, and the CLIP image encoder. We show that the text encoders favor different image formations, and the CLIP image encoder provides a useful style transfer capability that allows a user to use a style reference photo to influence the text-to-image output.
    \item We devise a training-free extension that enables the \pww{} capability through a cross-attention modulation scheme, which allows users additional spatial control over the text-to-image output.
\end{enumerate}

%% file: figures/method/motivation.tex
\begin{figure}[t!]
    \centering
    \includegraphics[width=0.47\textwidth,trim={0.5cm 0cm 0cm 0cm}]{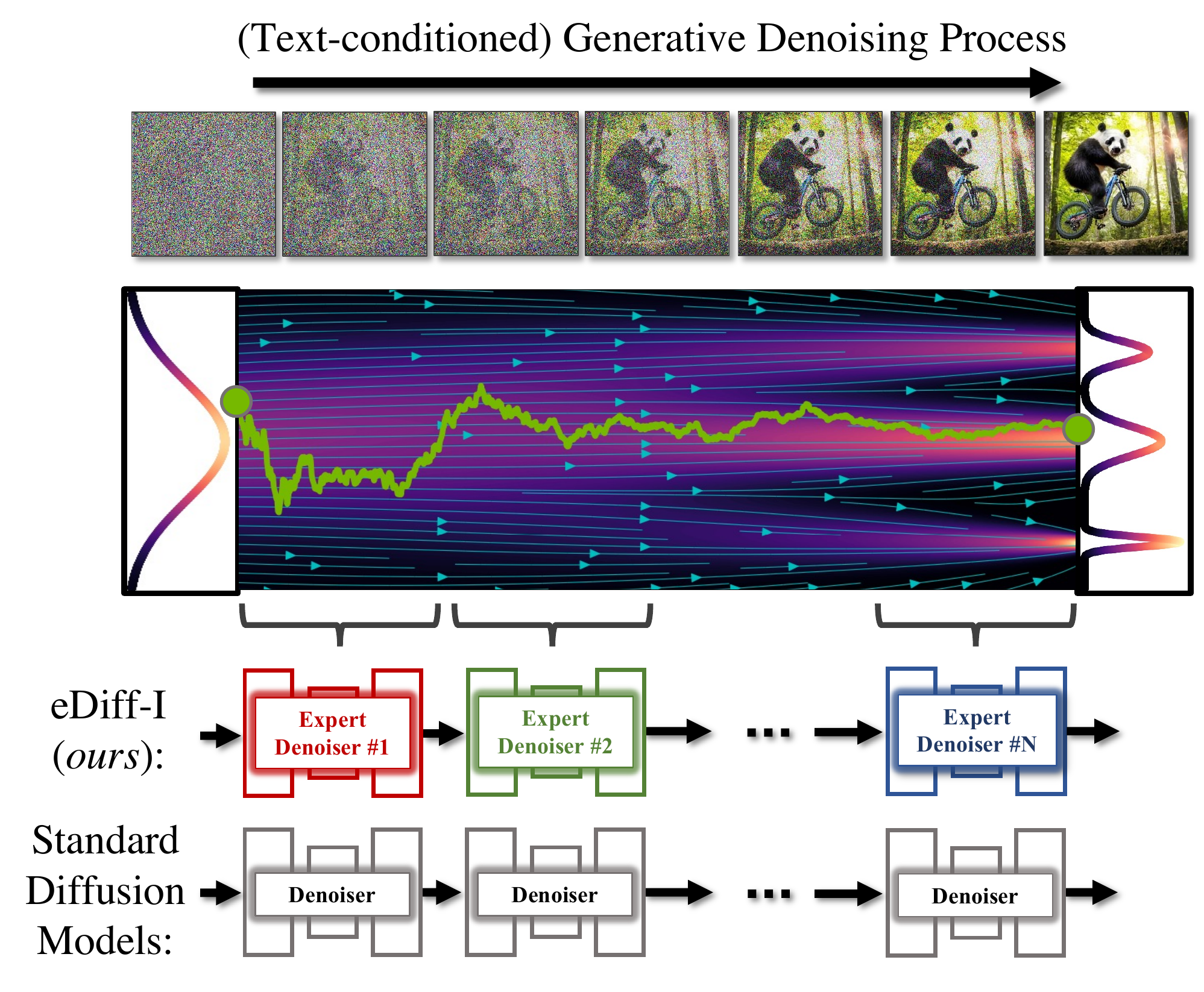}
    \caption{Synthesis in diffusion models corresponds to an iterative denoising process that gradually generates images from random noise; a corresponding stochastic process is visualized for a one-dimensional distribution. Usually, the same denoiser neural network is used throughout the entire denoising process. In \ours, we instead train an ensemble of expert denoisers that are specialized for denoising in different intervals of the generative process.}
    \label{fig:diffusion_models}
\end{figure}

%% file: figures/method/attention_maps.tex
\begin{figure*}[t!]
    \centering
    \includegraphics[width=\textwidth,trim={0cm 1cm 0cm 0cm}]{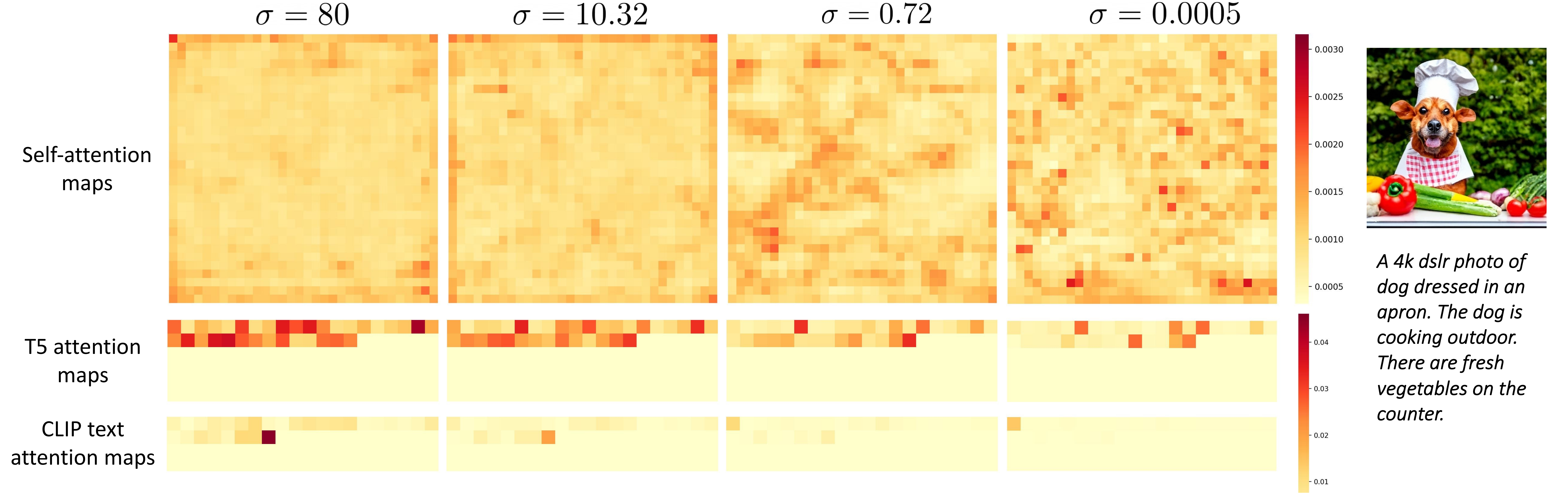}
    \caption{Visualization of attention maps at different noise levels. We plot the self-attention heat maps for visual features (top). Each value indicates how often the visual feature at this location is used and is the result of averaging over all attention queries. We also plot the cross-attention heat maps (bottom), where each value indicates how often the corresponding text token is used. We plot the cross-attention heat maps for both the T5 text tokens and the CLIP text tokens. Note that our cross-attention layer also includes a null token, which is not shown in the figure. When the attention values for all text tokens are low, the null token is mostly attended. The figure shows that the text attention value is strong at higher noise levels where the core image formation occurs. In these regions, the model relies on the text to generate a rough layout consistent with the text description. On the other hand, at lower noise levels, the text attention value is weak because the images are mostly formed at this point. The models need not rely on text information for denoising. Conversely, in the case of self-attention maps, the attention values are evenly distributed at higher noise levels, as the image content in these regions is not informative. At lower noise levels, the attention maps exhibit patterns better correlated with the image content.}
    \label{fig:attention_maps}
\end{figure*}

%% file: figures/method/prompt_switching.tex
\begin{figure*}
    \centering
    \includegraphics[width=\textwidth,trim={0cm 1cm 0cm 0cm}]{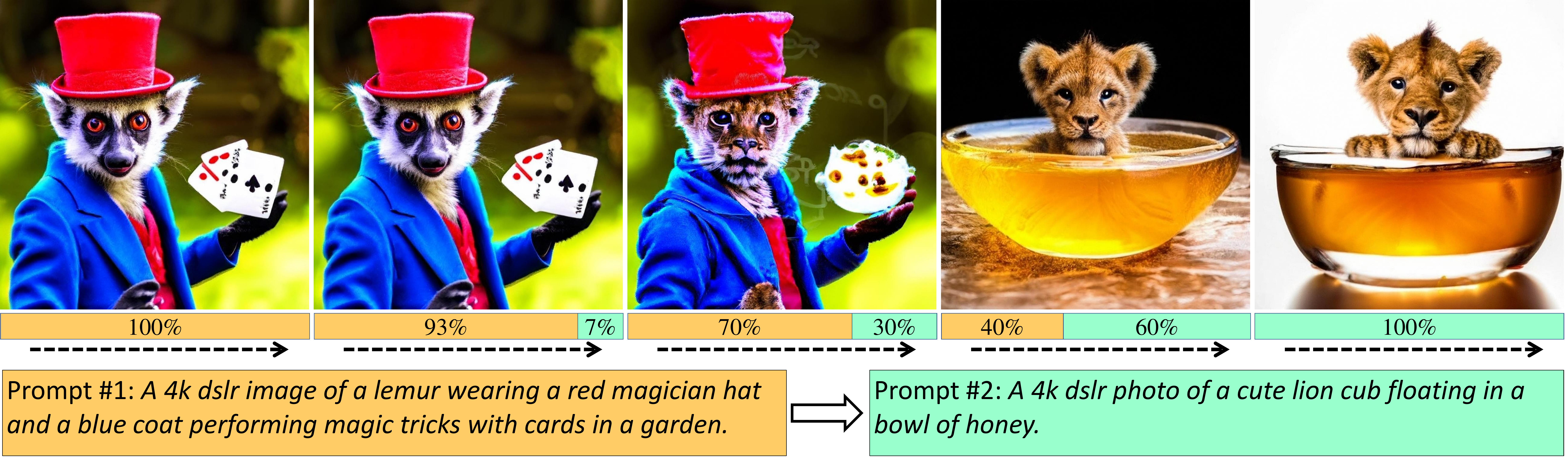}
    \caption{Impact of prompt switching during iterative denoising. We change the input text to Prompt \#2 after a fixed percentage of denoising steps have been performed using Prompt \#1. From left to right, the 5 images are produced with different transition percentages, which are 0\%, 7\%, 30\%, 60\%, and 100\%, as visualized in the figure. Comparing the first and second outputs, we note that the text inputs have no visible impact on the output when used in the last 7\% of denoising, which suggests text prompt is not used at the end of iterative denoising. The third output shows influences from both prompts, where the lemur and cards are replaced with lion and honey, respectively. The fourth output suggests the text input in the first 40\% of denoising is overridden by the text input in the remaining 60\%. From these results, we find the denoiser utilizes text input differently at different noise levels.}
    \label{fig:prompt_switching}
    \vspace{-4mm}
\end{figure*}

%% file: sections/related.tex
\section{Related Work}\label{sec:related}

Denoising Diffusion models~\cite{sohl-dickstein2015deep,ho2020denoising,song2020score} are a class of deep generative models that generate samples through an iterative denoising process. These models are trained with denoising score matching~\cite{Hyvarinen05,Vincent11} objectives at different noise levels and thus are also known as noise-conditioned score networks~\cite{song2019generative,song2020improved}. They have driven successful applications such as text-to-image generation~\cite{saharia2022photorealistic,ramesh2022hierarchical,rombach2022high}, natural language generation~\cite{li2022diffusion}, time series prediction~\cite{tashiro2021csdi}, audio synthesis~\cite{kong2020diffwave}, 3D shape generation~\cite{zhou2021shape,luo2021diffusion,zeng2022lion}, molecular conformation generation~\cite{xu2022geodiff}, protein structure generation~\cite{wu2022protein}, machine learning security~\cite{nie2022DiffPure}, and differentially private image synthesis~\cite{dockhorn2022differentially}.

\paragraph{Text-to-image diffusion models}
Some of the most high-quality text-to-image generative models are based on diffusion models. These models learn to perform the denoising task conditioned on text prompts, either on the image space (such as GLIDE~\cite{nichol2021glide} and Imagen~\cite{saharia2022photorealistic}) or on a separate latent space (such as DALL$\cdot$E 2~\cite{ramesh2022hierarchical}, Stable Diffusion~\cite{rombach2022high,stablediffusion1-4}, and VQ-Diffusion \cite{gu2022vector}). For computational efficiency, a diffusion model is often trained on low-resolution images or latent variables, which are then transformed into high-resolution images by super-resolution diffusion models~\cite{ho2022cascaded} or latent-to-image decoders~\cite{vahdat2021score,sinha2021d2c}. Samples are drawn from these diffusion models using classifier(-free) guidance~\cite{dhariwal2021diffusion,ho2022classifier} as well as various sampling algorithms that use deterministic~\cite{song2020denoising,lu2022dpm,liu2022pseudo,zhang2022fast,karras2022elucidating,dockhorn2022genie} or stochastic~\cite{bao2022analytic,bao2022estimating,dockhorn2021score,zhang2022gddim} iterative updates. 
Several works retrieve auxiliary images related to the text prompt from an external database and condition generation on them to boost performance~\cite{blattmann2022semi,sheynin2022knndiffusion,chen2022reimagen}.
Recently, several text-to-video diffusion models were proposed and achieved high-quality video generation results~\cite{ho2022video,ho2022imagen,yang2022diffusion,singer2022make,harvey2022flexible}.

\paragraph{Applications of text-to-image diffusion models}
Apart from serving as a backbone to be fine-tuned for general image-to-image translation tasks~\cite{wang2022pretraining}, text-to-image diffusion models have also demonstrated impressive capabilities in other downstream applications. Diffusion models can be directly applied to various inverse problems, such as super-resolution~\cite{saharia2021image,dhariwal2021diffusion}, inpainting~\cite{lugmayr2022repaint,chung2022improving}, deblurring~\cite{whang2021deblurring,kawar2022denoising}, and JPEG restoration~\cite{saharia2022palette,kawar2022jpeg}. For example, blended diffusion~\cite{avrahami2022blended,avrahami2022blendedlatent} performs inpainting with natural language descriptions. Text-to-image diffusion models can also perform other semantic image editing tasks. SDEdit~\cite{meng2021sdedit} enables re-synthesis, compositing, and editing of an existing image via colored strokes or image patches. DreamBooth~\cite{ruiz2022dreambooth} and Textual Inversion~\cite{gal2022image} allow the ``personalization'' of models by learning a subject-specific token from a few images. Prompt-to-prompt tuning can achieve image editing by modifying the textual prompt used to produce the same image without having the user provide object-specific segmentation masks~\cite{hertz2022prompt}. Similar image-editing capabilities can also be achieved by fine-tuning the model parameters~\cite{kawar2022imagic,valevski2022unitune} or automatically finding editing masks with the denoiser~\cite{couairon2022diffedit}.

\paragraph{Scaling up deep learning models}
The recent success of deep learning has primarily been driven by increasingly large models and datasets. It has been shown that simply scaling up model parameters and data size results in substantial performance improvement in various tasks, such as language understanding~\cite{brown2020language,chowdhery2022palm}, visual recognition~\cite{zhai2022scaling}, and multimodal reasoning~\cite{radford2021learning}. However, those high-capacity models also incur increased computational and energy costs during training and inference. Some recent works~\cite{shazeer2017outrageously,riquelme2021scaling, artetxe2021efficient} employ sparse expert models which route each input example to a small subset of network weights, thereby keeping the amount of computation tractable as scaling. Similarly, our proposed expert denoisers increase the number of trainable parameters without adding the computational cost at test time.

%% file: sections/preliminary.tex
\section{Background}\label{sec:background}

In text-to-image generative models, the input text is often represented by a text embedding, extracted from a pre-trained model such as CLIP~\cite{radford2021learning} or T5~\cite{raffel2020exploring} text encoders. In this case, the problem of generating images given text prompts simply boils down to learning a conditional generative model that takes text embeddings as input conditioning and generates images aligned with the conditioning.

Text-to-image diffusion models generate data by sampling an image from a noise distribution and iteratively denoising it using a denoising model $D(\xx;\ee, \sigma)$ where $\xx$ represents the noisy image at the current step, $\ee$ is the input embedding, and $\sigma$ is a scalar input indicating the current noise level. Next, we formally discuss how the denoising model is trained and used for sampling.

\paragraph{Training} The denoising model is trained to recover clean images given their corrupted versions, generated by adding Gaussian noise of varying scales. Following the EDM formulation of Karras~\etal~\cite{karras2022elucidating} and their proposed corruption schedule~\cite{song2020denoising,karras2022elucidating}, we can write the training objective as:
\begin{equation*}
    \expect_{\pdata(\xxclean, \ee),p(\beps), p(\sigma)}[\lambda(\sigma)\Vert D(\xxclean + \sigma \beps;\ee, \sigma) - \xxclean \Vert^2_2]
\end{equation*}
where $\pdata(\xxclean, \ee)$ represents the training data distribution that produces training image-text pairs, $p(\beps)=\normal(\boldzero, \boldI)$ is the standard Normal distribution, $p(\sigma)$ is the distribution in which noise levels are sampled from, and $\lambda(\sigma)$ is the loss weighting factor.

\paragraph{Denoiser formulation} Following Karras~\etal~\cite{karras2022elucidating}, we precondition the denoiser using:
\begin{equation}
\label{eq:denoiser}
D(\xx; \ee, \!\sigma)\!:=\!\Big(\frac{\sdata}{\sast}\Big)^2\xx + \frac{\sigma\cdot\sdata}{\sast} F_\theta\Big(\frac{\xx}{\sast}\,;\ee,\!\frac{\ln(\sigma)}{4}\Big)
\end{equation}
where \smash{$\sast=\sqrt{\sigma^2+\sdata^2}$} and $F_\theta$ is the trained neural network. We use $\sdata=0.5$ as an approximation for the standard deviation of pixel values in natural images. For $\sigma$, we use the log-normal distribution \mbox{$\ln(\sigma)\sim\normal(\pmean,\pstd)$} with $\pmean=-1.2$ and $\pstd=1.2$, and weighting factor \mbox{$\lambda(\sigma)=(\sast/(\sigma\cdot\sdata))^2$} that cancels the output weighting of $F_\theta$ in~(\ref{eq:denoiser}).

\paragraph{Sampling}
To generate an image with the diffusion models, an initial image is generated by sampling from the prior distribution \mbox{$\xx\sim\normal(\boldzero,\smax^2\boldI)$}, and then the generative ordinary differential equation (ODE) is solved using:
\begin{equation}\label{eq:gegenerative_ode}
\frac{\dx}{\dsigma} = -\sigma\,\nablaxx\log p(\xx|\ee, \sigma)
    = \frac{\xx - D(\xx;\ee,\sigma)}{\sigma}
\end{equation}
for $\sigma$ flowing backward from $\smax$ to $\smin \approx 0$. Above, $\nablaxx\log p(\xx|\ee, \sigma)$ represents \textit{the score function} of the corrupted data at noise level $\sigma$ which is obtained from the denoising model~\cite{Hyvarinen05,Vincent11}. Above $\smax$ represents a high noise level at which all the data is completely corrupted, and the mutual information between the input image distribution and the corrupted image distribution is approaching zero. Note that sampling can also be expressed as solving a stochastic differential equation as discussed in Song~\etal~\cite{song2020score}.

\paragraph{Super-resolution diffusion models} The training of text-conditioned super-resolution diffusion models largely follows the training of text-conditioned diffusion models described above. The major difference is that the super-resolution denoising model also takes the low-resolution image as a conditioning input. Following prior work~\cite{ramesh2022hierarchical}, we apply various corruptions to the low-resolution input image during training~\cite{wang2021real} to enhance the generalization capability of the super-resolution model.

%% file: sections/body.tex
\section{Ensemble of Expert Denoisers}

As we discussed in the previous section, text-to-image diffusion models rely on a denoising model to convert samples from a prior Gaussian distribution to images conditioned on an input text prompt. Formally, the generative ODE shown in~(\ref{eq:gegenerative_ode}) uses $D(\xx;\ee,\sigma)$ to guide the samples gradually towards images that are aligned with the input conditioning. 

The denoising model $D$ at each noise level $\sigma$ relies on two sources of information for denoising: the current noisy input image $\xx$ and the input text prompt $\ee$. Our key observation is that text-to-image diffusion models exhibit a unique temporal dynamic while relying on these two sources. At the beginning of the generation, when $\sigma$ is large, the input image $\xx$ contains mostly noise. Hence, denoising directly from the input visual content is a challenging and ambiguous task. At this stage, $D$ mostly relies on the input text embedding to infer the direction toward text-aligned images. However, as $\sigma$ becomes small towards the end of the generation, most coarse-level content is painted by the denoising model. At this stage, $D$ mostly ignores the text embedding and uses visual features for adding fine-grained details.

We validate this observation in Figure~\ref{fig:attention_maps} by visualizing the cross-attention maps between visual and text features compared to the self-attention maps on the visual features at different stages of generation. In Figure~\ref{fig:prompt_switching}, we additionally examine how the generated sample changes as we switch the input caption from one prompt to another at different stages of the denoising process. When the prompt switching happens at the last $7 \%$ of denoising, the generation output remains the same. On the other hand, when the prompt switching happens at the first $40 \%$ of the training, the output changes completely.

In most existing works on diffusion models, the denoising model is shared across all noise levels, and the temporal dynamic is represented using a simple time embedding that is fed to the denoising model via an MLP network. We argue that the complex temporal dynamics of the denoising diffusion may not be learned from data effectively using a shared model with a limited capacity. Instead, we propose to scale up the capacity of the denoising model by introducing an ensemble of expert denoisers; each expert denoiser is a denoising model specialized for a particular range of noise levels~(see Figure~\ref{fig:diffusion_models}). This way, we can increase the model capacity without slowing down the sampling since the computational complexity of evaluating $D$ at each noise level remains the same.

However, naively training separate denoising models for different stages can significantly increase the training cost since one needs to train each expert denoiser from scratch. To remedy this, we first train a shared model across all noise levels. We then use this model to initialize the denoising experts in the next stage. Next, we discuss how we formally create denoising experts from a pre-trained model iteratively.

\subsection{Efficient Training of Expert Denoisers}\label{sec:body_branching}

We propose a branching strategy based on a binary tree implementation for training the expert denoisers efficiently. We begin by training a model shared among all noise levels using the full noise level distribution denoted as $p(\sigma)$. Then, we initialize two experts from this baseline model. Let us call these models the level 1 experts since they are trained on the first level of the binary tree. These two experts are trained on the noise distributions $p^{1}_{0}(\sigma)$ and $p^{1}_{1}(\sigma)$, which are obtained by splitting $p(\sigma)$ equally by area. So, the expert trained on $p^{1}_{0}(\sigma)$ specializes in low noise levels, while the expert trained on $p^{1}_{1}(\sigma)$ specializes in high noise levels. In our implementation, $p(\sigma)$ follows a log-normal distribution (Sec.~\ref{sec:background}). Recently, Luhman~\etal\cite{luhman2022improving} have also trained two diffusion models for two-stage denoising for image generation, but their models are trained over images of different resolutions and separately from the beginning.

Once the level 1 expert models are trained, we split each of their corresponding noise intervals in a similar fashion as described above and train experts for each sub-interval. This process is repeated recursively for multiple levels. In general, at level $l$, we split the noise distribution $p(\sigma)$ into $2^{l}$ intervals of equal area given by $\{p^{l}_{i}(\sigma)\}_{i=0}^{2^l-1}$, with model $i$ being trained on the distribution $p^{l}_{i}(\sigma)$. We call such a model or node in the binary tree $E^l_i$.

\input{figures/method/sr_cond_pipeline}

Ideally, at each level $l$, we would have to train $2^{l}$ models. However, this is impractical as the model size grows exponentially with the depth of the binary tree. Also, in practice, we found that models trained at many of the intermediate intervals do not contribute much toward the performance of the final system. Therefore, we focus mainly on growing the tree from the left-most and the right-most nodes at each level of the binary tree: $E^l_{0}$ and $E^{l}_{2^l-1}$. The right-most interval contains samples at high noise levels. As shown in Figures~\ref{fig:attention_maps} and \ref{fig:prompt_switching}, good denoising at high noise levels is critical for improving text conditioning as core image formation occurs in this regime. Hence, having a dedicated model in this regime is desired. Similarly, we also focus on training the models at lower noise levels as the final steps of denoising happen in this regime during sampling. So, good models are needed to get sharp results. Finally, we train a single model on all the intermediate noise intervals that are between the two extreme intervals.

In a nutshell, our final system would have an ensemble of three expert denoisers: an expert denoiser focusing on the low noise levels (given by the leftmost interval in the binary tree), an expert denoiser focusing on high noise levels (given by the right-most interval in the binary tree), and a single expert denoiser for learning all intermediate noise intervals. A more detailed description of our branching strategy is described in Appendix~\ref{sec:app_branching}. In Sec.~\ref{sec:expr}, we also consider other types of ensemble experts for quantitative evaluation purposes.

\subsection{Multiple Conditional Inputs}\label{sec:arch}

To train our text-to-image diffusion models, we use the following conditional embeddings during training: (1) T5-XXL ~\cite{raffel2020exploring} text embeddings, (2) CLIP L/14 text embeddings and (3) CLIP L/14 image embeddings. We pre-compute these embeddings for the whole dataset since computing them online is very expensive. Similar to prior work~\cite{ho2022cascaded,saharia2022photorealistic,rombach2022high}, we add the projected conditional embeddings to the time embedding and additionally perform cross attention at multiple resolutions of the denoising model. We use random dropout~\cite{srivastava2014dropout} on each of these embeddings independently during training. When an embedding is dropped, we zero out the whole embedding tensor. When all three embeddings are dropped, it corresponds to unconditional training, which is useful for performing classifier-free guidance~\cite{ho2022classifier}. We visualize the input conditioning scheme in~ Figure~\ref{fig:sr_cond_pipeline}.

Our complete pipeline consists of a cascade of diffusion models. Specifically, we have a base model that can generate images of $64{\times}64$ resolution and two super-resolution diffusion models that can progressively upsample images to $256{\times}256$ and $1024{\times}1024$ resolutions, respectively (see Figure~\ref{fig:sr_cond_pipeline}). To train the super-resolution models, we condition on the ground-truth low-resolution inputs that are corrupted by random degradation~\cite{wang2021real}. Adding degradation during training allows the models to better generalize to remove artifacts that can exist in the outputs generated by our base model. For the base model, we use a modified version of the U-net architecture proposed in Dhariwal~\etal~\cite{dhariwal2021diffusion}, while for super-resolution models, we use a modified version of the Efficient U-net architecture proposed in~Saharia~\etal~\cite{saharia2022photorealistic}. More details on the architectures can be found in Appendix~\ref{sec:network_archtecture}.

\input{figures/method/paint_with_words.tex}

\subsection{Paint-with-words}\label{sec:paint-with-words}

We propose a training-free method, named \textit{paint-with-words}, that enables users to specify the spatial locations of objects. As shown in Figure~\ref{fig:paint_with_words}, users can select an arbitrary phrase from the text prompt and doodle on a canvas to create a binary mask corresponding to that phrase. The masks are input to all cross-attention layers and are bilinearly downsampled to match the resolution of each layer. We use those masks to create an input attention matrix $A\in \mathbb{R}^{N_{i}\times N_{t}}$ where $N_{i}$ and $N_{t}$ are the number of image and text tokens, respectively. Each column in $A$ is generated by flattening the mask corresponding to the phrase that contains the text token of that column. The column is set to zero if the corresponding text token is not contained in any phrases selected by the user. We add the input attention matrix to the original attention matrix in the cross-attention layer, which now computes the output as $\mathrm{softmax}\big(\frac{QK^T+wA}{\sqrt{d_k}}\big)V$, where $Q$ is query embeddings from image tokens, $K$ and $V$ are key and value embeddings from text tokens, $d_k$ is the dimensionality of $Q$ and $K$, and $w$ is a scalar weight that controls the strength of user input attention. Intuitively, when the user paints a phrase on a region, image tokens in that region are encouraged to attend more to the text tokens contained in that phrase. As a result, the semantic concept corresponding to that phrase is more likely to appear in the specified area.

We find it beneficial to use a larger weight at higher noise levels and to make the influence of $A$ irrelevant to the scale of $Q$ and $K$, which corresponds to a schedule that works well empirically: 
\begin{equation*}
w=w^{\prime}\cdot\log(1+\sigma)\cdot\max(QK^T),    
\end{equation*}
where $w^\prime$ is a scalar specified by the user. 

%% file: figures/method/sr_cond_pipeline.tex
\begin{figure}[t!]
    \centering
    \includegraphics[width=0.48\textwidth,trim={0cm 0.5cm 0.3cm 0cm}]{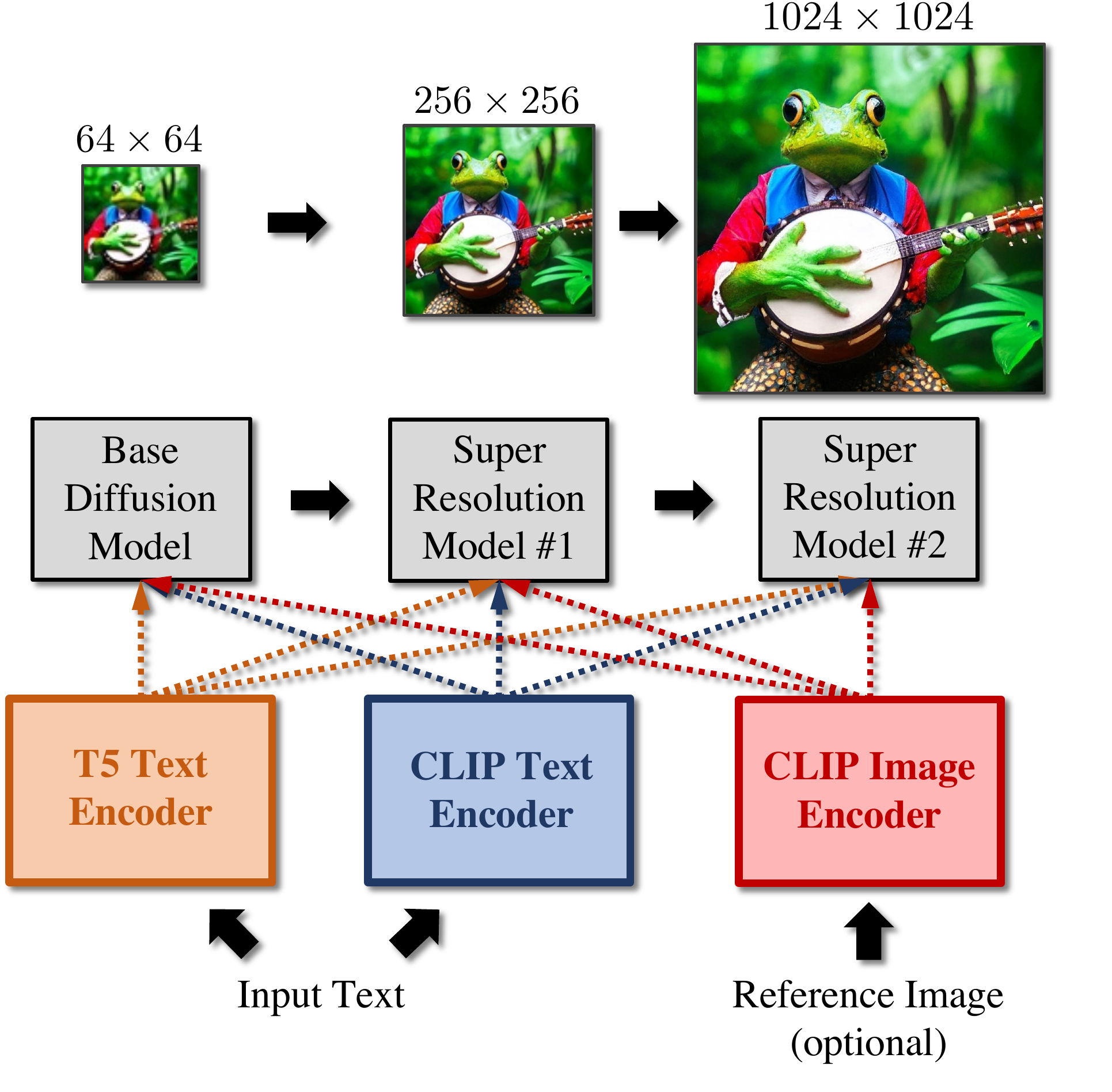}
    \caption{\ours consists of a base diffusion model that generates images in $64{\times}64$ resolution. This is followed by two super-resolution diffusion models that upsample the images to $256{\times}256$ and $1024{\times}1024$ resolution, respectively, referred to as SR256 and SR1024 through the paper. All models are conditioned on text through both T5 and CLIP text embeddings. \ours also allows the user to optionally provide an additional CLIP image embedding. This can enable detailed stylistic control over the output (see Figures~\ref{fig:teaser}~and~\ref{fig:clip_image_conditioning}).}
    \label{fig:sr_cond_pipeline}
    \vspace{-4mm}
\end{figure}

%% file: figures/method/paint_with_words.tex
\begin{figure}[t!]
    \centering
    \includegraphics[width=0.48\textwidth]{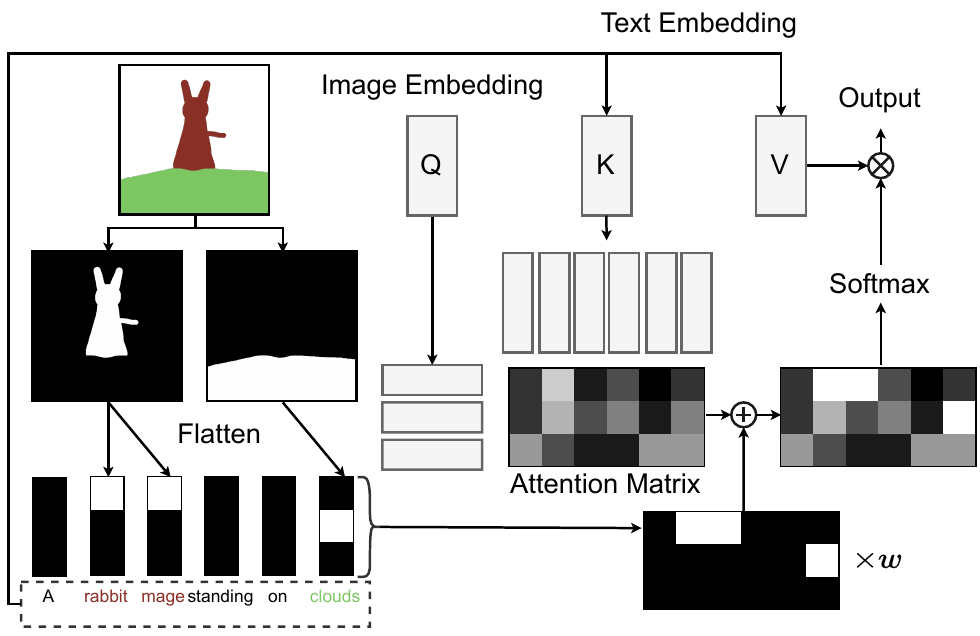}
    \caption{Illustration of the proposed \pww{} method. The user can control the location of objects by selecting phrases (here ``rabbit mage'' and ``clouds''), and painting them on the canvas. The user-specified masks increase the value of corresponding entries of the attention matrix in the cross-attention layers.}
    \label{fig:paint_with_words}
    \vspace{-4mm}
\end{figure}

%% file: sections/expr.tex
\section{Experiments}\label{sec:expr}

\input{figures/curve_ensemble_baseline_new}
\input{figures/curve_input_change_new}
\input{figures/model_size_comparison}

First, we discuss optimization, dataset, and evaluation. We then show improved quantitative results compared to previous methods in Sec.~\ref{sec:main-results}. Then, we perform two sets of ablation studies as well as study the effect of increasing the number of experts over standard image quality metrics. In Sec~\ref{sec:ablation-clip-t5}, we evaluate image generation performance based on CLIP and/or T5 text embeddings, where we show that having both embeddings lead to the best image generation quality. Finally, we discuss two novel applications that are enabled by \ours. In Sec~\ref{sec:style-transfer}, we illustrate style transfer applications that are enabled by CLIP image embeddings, and in Sec~\ref{sec:paint-with-words-results}, we present image generation results with the introduced \textit{paint-with-words} method.

\paragraph{Optimization}
Both the base and super-resolution diffusion models are trained using the AdamW~\cite{loshchilov2019adamw} optimizer with a learning rate of $0.0001$, weight decay of $0.01$, and a batch size of $2048$. The base model was trained using $256$ NVIDIA A100 GPUs, while the two super-resolution models were trained with $128$ NVIDIA A100 GPUs each. Our implementation is based on the Imaginaire library~\cite{imaginaire} written in PyTorch~\cite{pytorch}. More details on other hyperparameters settings are presented in Appendix~\ref{sec:hyperparams}.

\paragraph{Datasets}
We use a collection of public and proprietary datasets to train our model. To ensure high-quality training data, we apply heavy filtering using a pretrained CLIP model to measure the image-text alignment score as well as an aesthetic scorer to rank the image quality. We remove image-text pairs that fail to meet a preset CLIP score threshold and a preset aesthetic score. The final dataset to train our model contains about one billion text-image pairs. All the images have the shortest side greater than 64 pixels. We use all of them to train our base model. We only use images with the shortest side greater than 256 and 1024 pixels to train our SR256 and SR1024 models, respectively. For training our base and SR256 models, we perform resize-central crop. Images are first resized so that the shortest side has the same number of pixels as the input image side. For training the SR1024 model, we randomly crop $256{\times}256$ regions during training and apply it on $1024{\times}1024$ resolution during inference. We use COCO and Visual Genome datasets for evaluation, which are excluded from our training datasets for measuring zero-shot text-to-image generation performance.

\paragraph{Evaluation}
We use MS-COCO~\cite{lin2014microsoft} dataset for most of the evaluation. Consistent with prior work~\cite{ramesh2022hierarchical, saharia2022photorealistic}, we report zero-shot FID-30K~\cite{heusel2017gans} in which 30K captions are drawn randomly from the COCO validation set. We use the captions as inputs for synthesizing images. We compute the FID between these generated samples and the reference 30K ground truth images. We also report the CLIP score, which measures the average similarity between the generated samples and the corresponding input captions using features extracted from a pre-trained CLIP model. In addition to the MS-COCO validation set, we also report the CLIP and FID scores on Visual Genome dataset~\cite{krishna2016visualgenome}, which is a challenging dataset containing images and paired long captions.

\subsection{Main Results}\label{sec:main-results}

\input{tables/comparison_fid.tex}

First, we study the effectiveness of our ensemble model by plotting the FID-CLIP score trade-off curve and comparing it with our baseline model, which does not use our proposed ensemble-of-expert-denoisers scheme but shares all the remaining design choices. The trade-off curves are generated by performing a sweep over classifier-free guidance values in the range $\{0, 0.5, 1.0, \hdots 9.5, 10.0\}$. In this experiment, we train an ensemble of four expert models by splitting the baseline model trained to $500K$ iterations and training each child model for $50K$ iterations. We further split each of these child models at $550K$ iterations and train the resulting four models for another $50K$ steps. This results in four expert models, $E^2_0$, $E^2_1$, $E^2_3$, and $E^2_4$, trained for $600K$ steps. To make a fair comparison, we evaluate the ensemble model against our baseline model that is trained for $800K$ steps, as this would correspond to both models seeing the same number of training samples. As shown in Figure~\ref{fig:fig_ensemble_curves}, our ensemble model outperforms the baseline model by a significant margin on the entire trade-off curve.

Next, we report the FID-30K results of \ours computed on $256{\times}256$ resolution images on the MS-COCO dataset and compared it with the state-of-the-art methods in Table~\ref{tab:comparison_fid}. We experiment with the following model settings: 
\begin{itemize}
    \item \configa \medspace Our baseline model for the base model generates images at $64{\times}64$ resolution. The outputs are upsampled with our baseline SR256 model.
    \item \configb The same base model as in \configa. The ensemble SR256 model consists of two experts: $E^1_0$ and $E^1_1$.
    \item \configc Our $2$-expert ensemble base model generates images at $64{\times}64$ resolution. This ensemble model consists of an expert model trained at leaf nodes $E^9_{511}$ and a complement model trained on all other noise levels except $E^9_{511}$. The outputs are upsampled with our ensemble SR256 model as in \configb.
    \item \configd Our $3$-expert ensemble base model generates images at $64{\times}64$ resolution. The $3$-expert ensemble model consists of $E^9_{511}$ model (high noise regime model), $E^3_{0}$ (low noise regime model), and an expert denoiser model covering the noise levels in-between. The outputs are upsampled with our ensemble SR256 model as in \configb.
\end{itemize}

\input{figures/comparison_multiple_entities}
\input{figures/comparison_text}
\input{figures/comparison_long_captions}

We observe that our \configa model outperforms GLIDE~\cite{nichol2021glide}, DALL$\cdot$E 2~\cite{ramesh2022hierarchical}, Make-a-Scene~\cite{gafni2022make}, and Stable Diffusion~\cite{rombach2022high}, and achieves FID slightly higher than that of Imagen~\cite{saharia2022photorealistic} and Parti~\cite{yu2022scaling}. By applying the proposed ensemble scheme to the SR256 model (\configb), we achieve an FID score of 7.26, which is slightly better than Imagen. As we apply the proposed $2$-expert ensemble scheme to build \configc, which has roughly the same model size as Imagen, we outperform both Imagen and Parti by an FID score of 0.16 and 0.12, respectively. Our \configd model achieves the best FID of 7.04. In Figure~\ref{fig:comparison_ensemble}, we qualitatively compare the results of \configa with those of \configc. We observe that our ensemble of expert denoisers generates improved results compared with the baseline.

\input{figures/curve_speed_comparison_ensmble_depth.tex}

We now report qualitative comparison results using our best \ours configuration with two publicly available text-to-image generative models\,---\,Stable Diffusion~\cite{rombach2022high} and DALL$\cdot$E~2~\cite{ramesh2022hierarchical} in Figures~\ref{fig:comparison_multiple_entities}, \ref{fig:comparison_text}, and \ref{fig:comparison_long_captions}. In the presence of multiple entities (Figure~\ref{fig:comparison_multiple_entities}), Stable Diffusion and DALL$\cdot$E~2 tend to mix the attributes from different entities or ignore some of the attributes, while \ours can accurately model attributes from all entities. In generating texts (Figure~\ref{fig:comparison_text}), both Stable Diffusion and DALL$\cdot$E~2 often produce misspellings or ignore words, while \ours correctly generates the texts. Even in the case of long descriptions, \ours can handle long-range dependencies better and perform better than DALL$\cdot$E~2 and Stable Diffusion. 

In Figure~\ref{fig:style_variants}, we show that \ours can generate images with a variety of styles by using the proper text prompts. Our model can also generate many variations for a given text prompt, as shown in Figure~\ref{fig:variations}.

In Figure~\ref{fig:fig_speed_comparison_curves}, we conduct an experiment to illustrate that the proposed ensemble-of-expert-denoisers scheme helps scale the model size without incurring additional computation in the inference time.

\subsection{CLIP Text and T5 Text}\label{sec:ablation-clip-t5}

As explained in Sec.~\ref{sec:arch}, we use both CLIP text embeddings and T5 text embeddings to train our models. Since we perform random dropout independently on the individual embeddings during training, the model has the capability to generate images when each of the embeddings is used in isolation. In Figure~\ref{fig:t5_vs_clip}, we examine the effect of the individual text embeddings in our model. We observe that images generated using the CLIP text embeddings alone typically have the correct foreground object, but lack in terms of compositionality, counting, and generating text. On the other hand, images generated by using only the T5 text embeddings obtain better compositions but are inferior in generating the foreground objects, such as the breeds of dogs. Using both T5 and CLIP text embeddings, we get the best of both worlds, where our model can use the provided attributes from each of the text embeddings.

Next, we quantitatively evaluate the effect of individual embeddings by plotting the CLIP-FID trade-off curve on MS-COCO and Visual Genome datasets in Figure~\ref{fig:fig_input_variation_curves}. We observe that, on the MS-COCO dataset, using CLIP and T5 embeddings in isolation results in a similar performance, while using CLIP+T5 embeddings leads to much better trade-off curves. On the visual genome dataset, using T5 embeddings in isolation leads to better performance than using CLIP text embeddings. A closer look at the dataset statistics reveals that the average number of words in each caption of the MS-COCO dataset is $10.62$, while it is $61.92$ for Visual Genome. So, when the text is more descriptive, the use of T5 embeddings performs better than the CLIP text embeddings. Again, the best performance is obtained by using CLIP+T5 embeddings.

\subsection{Style transfer}\label{sec:style-transfer}

In addition to the T5 and CLIP text embeddings, our model is also conditioned on CLIP image embeddings during training. We find that the use of CLIP image embeddings gives us the ability to do style transfer during synthesis. In Figure~\ref{fig:clip_image_conditioning}, we show some of our style transfer results. From a given reference image, we first obtain its CLIP image embedding. We then sample outputs conditioned on both the reference CLIP image embedding and the corresponding input text. We find that when CLIP image embeddings are not used, images are obtained in the natural style. On the other hand, when the CLIP image embeddings are active, images are generated in accordance with the style given by the reference image.

\subsection{Paint-with-words}\label{sec:paint-with-words-results}

We show some results produced by our ``paint-with-words'' approach in \cref{fig:paint_with_words_res}. Although the doodles are very coarse and do not contain the exact shape of objects, our method is still able to synthesize high-quality images that have the same rough layout. In most scenarios, this is more convenient than segmentation-to-image methods~\cite{isola2017image,park2019semantic,huang2022multimodal}, which are likely to fail when user-drawn shapes are different from shapes of real objects. Compared with text-conditioned inpainting methods~\cite{bau2021paint,avrahami2022blended,nichol2021glide} that apply a single concept to an image region, ``paint-with-words'' can generate the whole image containing multiple concepts in a single pass from scratch, without the need to start from an input image.

%% file: figures/curve_ensemble_baseline_new.tex
\input{figures/fid_clip_curves/data/coco_base}
\input{figures/fid_clip_curves/data/coco_ensemble}
\input{figures/fid_clip_curves/data/vg_base}
\input{figures/fid_clip_curves/data/vg_ensemble}
\renewcommand{\plothh}{94mm}
\renewcommand{\plotvv}{55mm}
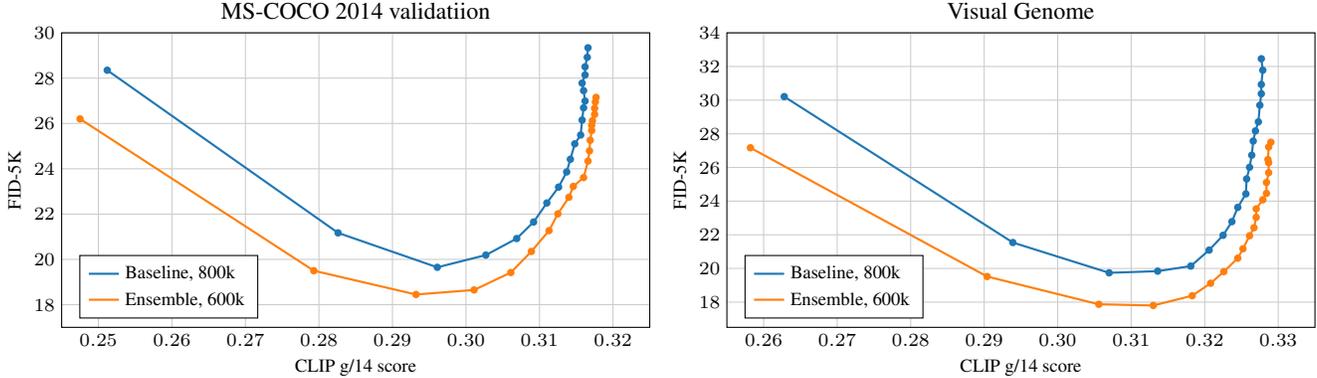
\begin{figure*}[ht!]
\centering%
\small%
\scriptsize%
\begin{tikzpicture}%
\begin{axis}[
  title={\small MS-COCO 2014 validatiion}, title style={yshift=-1ex},
  width=\plothh, height=\plotvv,
  xmin={0.245}, xmax={0.325}, 
  xmode={linear}, xtick={0.25, 0.26, 0.27, 0.28, 0.29, 0.30, 0.31, 0.32}, xticklabels={$0.25$, $0.26$, $0.27$, $0.28$, $0.29$, $0.30$, $0.31$, $0.32$},
  ymin={17}, ymax={30}, ymode={linear}, ytick={18, 20, 22, 24, 26, 28, 30}, yticklabels={$18$, $20$, $22$, $24$, $26$, $28$, $30$},
  xlabel={CLIP g/14 score}, xlabel near ticks,
  ylabel={FID-5K}, ylabel near ticks,
  grid={major}, legend pos={south west}, legend cell align={left}, legend style={font=\scriptsize},
]
\addplot[C0] coordinates {\genCocoBase};
\addplot[C1] coordinates {\genCocoEnsemble};
\addplot[C0, only marks, forget plot] coordinates {\genCocoBase};
\addplot[C1, only marks, forget plot] coordinates {\genCocoEnsemble};
\legend{
  {Baseline, 800k},
  {Ensemble, 600k},
}
\end{axis}
\end{tikzpicture}%
\hfill%
\begin{tikzpicture}%
\begin{axis}[
  title={\small Visual Genome}, title style={yshift=-1ex},
  width=\plothh, height=\plotvv,
  xmin={0.255}, xmax={0.335}, xmode={linear}, xtick={0.26, 0.27, 0.28, 0.29, 0.30, 0.31, 0.32, 0.33}, xticklabels={$0.26$, $0.27$, $0.28$, $0.29$, $0.30$, $0.31$, $0.32$, $0.33$},
  ymin={16.5}, ymax={34}, ymode={linear}, ytick={18, 20, 22, 24, 26, 28, 30, 32, 34}, yticklabels={$18$, $20$, $22$, $24$, $26$, $28$, $30$, $32$, $34$},
  xlabel={CLIP g/14 score}, xlabel near ticks,
  ylabel={FID-5K}, ylabel near ticks,
  grid={major}, legend pos={south west}, legend cell align={left}, legend style={font=\scriptsize},
]
\addplot[C0] coordinates {\genVgBase};
\addplot[C1] coordinates {\genVgEnsemble};
\addplot[C0, only marks, forget plot] coordinates {\genVgBase};
\addplot[C1, only marks, forget plot] coordinates {\genVgEnsemble};
\legend{
  {Baseline, 800k},
  {Ensemble, 600k},
}
\end{axis}
\end{tikzpicture}%
\vspace*{-.5ex}\\
\caption{\label{fig:fig_ensemble_curves}%
\textbf{2-Expert-Ensemble vs Baseline.}
We compare the FID-CLIP score trade-off curves between our 2-Expert-Ensemble model with our baseline model trained without any denoising expert. We sample 5K captions from each benchmark dataset. To make a fair comparison, we compare the baseline model trained for 800K iterations with our 2-Expert-Ensemble model trained for 600K iterations, as both models see the same number of training samples at this point. We show the results on the COCO dataset on the left panel and the results on the visual genome dataset on the right panel. In general, having a lower FID score and a higher CLIP score is more desirable. We observe that our ensemble model consistently outperforms the baseline model on the entire FID-CLIP score trade-off curve. }
\end{figure*}

%% file: figures/curve_input_change_new.tex
\input{figures/fid_clip_curves/data/coco_input_all}
\input{figures/fid_clip_curves/data/coco_input_clip}
\input{figures/fid_clip_curves/data/coco_input_tfive}
\input{figures/fid_clip_curves/data/vg_input_all}
\input{figures/fid_clip_curves/data/vg_input_clip}
\input{figures/fid_clip_curves/data/vg_input_tfive}
\renewcommand{\plothh}{94mm}
\renewcommand{\plotvv}{55mm}
\begin{figure*}[!bht]
\centering%
\small%
\scriptsize%
\hspace*{-1mm}%
\vspace{-2mm}
\begin{tikzpicture}%
\begin{axis}[
  width=\plothh, height=\plotvv,
  title={\small MS-COCO 2014 validatiion}, title style={yshift=-1ex},
  xmin={0.225}, xmax={0.325}, xmode={linear}, xtick={0.23, 0.24, 0.25, 0.26, 0.27, 0.28, 0.29, 0.30, 0.31, 0.32}, xticklabels={$0.23$, $0.24$, $0.25$, $0.26$, $0.27$, $0.28$, $0.29$, $0.30$, $0.31$, $0.32$},
  ymin={17}, ymax={30}, ymode={linear}, ytick={18, 20, 22, 24, 26, 28, 30}, yticklabels={$18$, $20$, $22$, $24$, $26$, $28$, $30$},
  xlabel={CLIP g/14 score}, xlabel near ticks,
  ylabel={FID-5K}, ylabel near ticks,
  grid={major}, legend pos={south west}, legend cell align={left}, legend style={font=\scriptsize},
]
\addplot[C0] coordinates {\genCocoInputAll};
\addplot[C1] coordinates {\genCocoInputTfive};
\addplot[C2] coordinates {\genCocoInputClip};
\addplot[C0, only marks, forget plot] coordinates {\genCocoInputAll};
\addplot[C1, only marks, forget plot] coordinates {\genCocoInputTfive};
\addplot[C2, only marks, forget plot] coordinates {\genCocoInputClip};
\legend{
  {T5 + CLIP},
  {T5 only},
  {CLIP only},
}
\end{axis}
\end{tikzpicture}%
\hfill%
\begin{tikzpicture}%
\begin{axis}[
  width=\plothh, height=\plotvv,
  title={\small Visual Genome}, title style={yshift=-1ex},
  xmin={0.245}, xmax={0.338}, xmode={linear}, xtick={0.25, 0.26, 0.27, 0.28, 0.29, 0.30, 0.31, 0.32, 0.33}, xticklabels={$0.25$, $0.26$, $0.27$, $0.28$, $0.29$, $0.30$, $0.31$, $0.32$, $0.33$},
  ymin={16.5}, ymax={30}, ymode={linear}, ytick={18, 20, 22, 24, 26, 28, 30}, yticklabels={$18$, $20$, $22$, $24$, $26$, $28$, $30$},
  xlabel={CLIP g/14 score}, xlabel near ticks,
  ylabel={FID-5K}, ylabel near ticks,
  grid={major}, legend pos={south west}, legend cell align={left}, legend style={font=\scriptsize},
]
\addplot[C0] coordinates {\genVgInputAll};
\addplot[C1] coordinates {\genVgInputTfive};
\addplot[C2] coordinates {\genVgInputClip};
\addplot[C0, only marks, forget plot] coordinates {\genVgInputAll};
\addplot[C1, only marks, forget plot] coordinates {\genVgInputTfive};
\addplot[C2, only marks, forget plot] coordinates {\genVgInputClip};
\legend{
  {T5 + CLIP},
  {T5 only},
  {CLIP only},
}
\end{axis}
\end{tikzpicture}%
\vspace*{-.5ex}\\
\caption{\label{fig:fig_input_variation_curves}%
\textbf{Effect of conditional embeddings.}
We plot the FID-CLIP score trade-off curves on the samples generated by our approach using (a) T5 + CLIP encoders, (b) T5 encoder only, and (c) CLIP text encoder only. On the COCO dataset, using T5 and CLIP encoders in isolation gives a similar performance, while using CLIP+T5 gives the best results. On the visual genome dataset, using the T5 encoder in isolation gives better performance than using the CLIP text encoder. This is because the visual genome dataset contains longer and more descriptive captions than the COCO dataset, and the use of a more powerful language model such as T5 becomes more beneficial. Nevertheless, using T5+CLIP performs the best even in this dataset with our model. 
}
\vspace{-2mm}
\end{figure*}
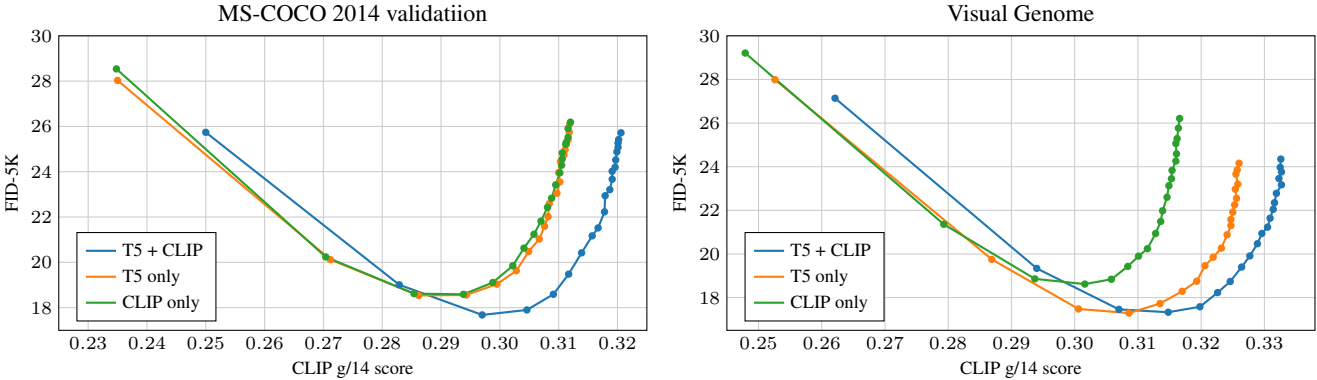

%% file: figures/model_size_comparison.tex
\setlength{\tabcolsep}{0.5pt}
\renewcommand{\arraystretch}{0.5}
\begin{figure*}[ht!]
    \centering
    \begin{tabularx}{\textwidth}{M{0.25\textwidth} M{0.25\textwidth} M{0.25\textwidth} M{0.25\textwidth}}
         \multicolumn{2}{c}{ (a) Baseline (no expert)} \vspace*{0.5mm} & \multicolumn{2}{c}{ (b) \ours } \vspace*{0.5mm} \\
        \includegraphics[width=0.248\textwidth]{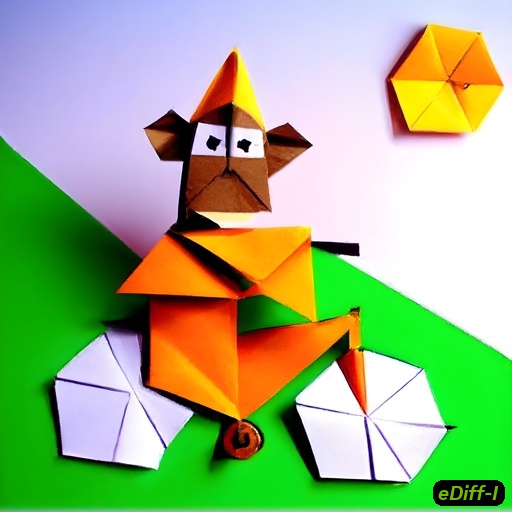} &
        \includegraphics[width=0.248\textwidth]{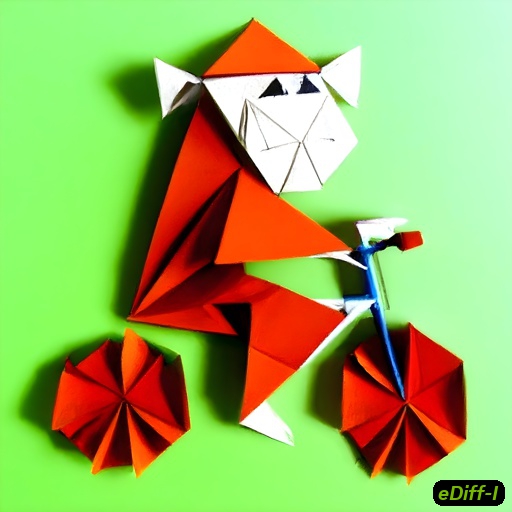} &
        \includegraphics[width=0.248\textwidth]{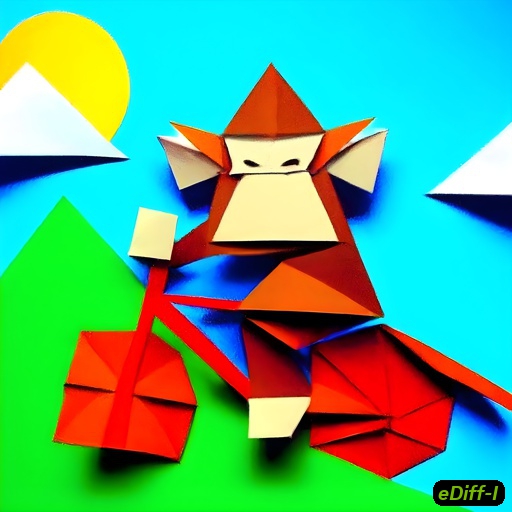} &
        \includegraphics[width=0.248\textwidth]{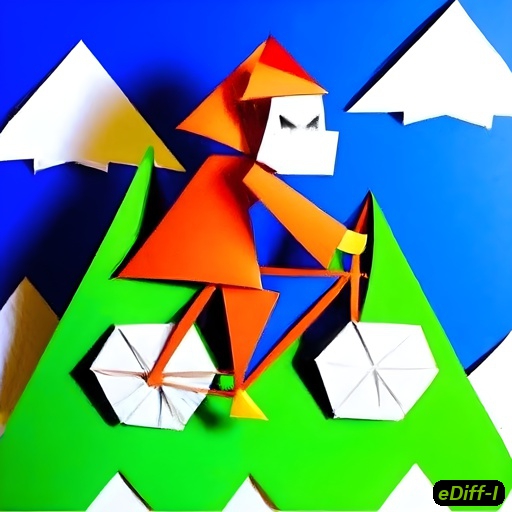} \\
        \multicolumn{4}{c}{\emph{An origami of a monkey dressed as a monk riding a bike on a mountain.}} \vspace*{2mm} \\
        \includegraphics[width=0.248\textwidth]{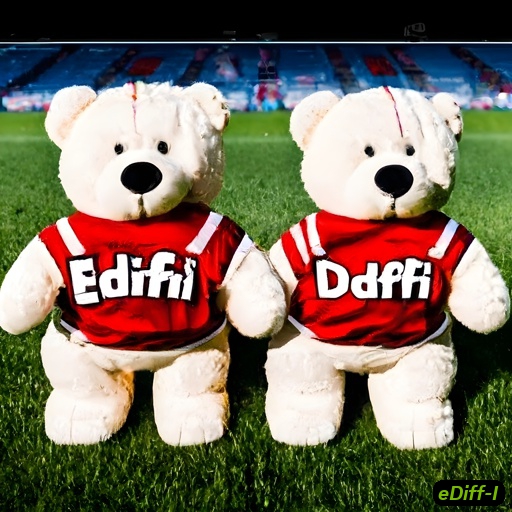} &
        \includegraphics[width=0.248\textwidth]{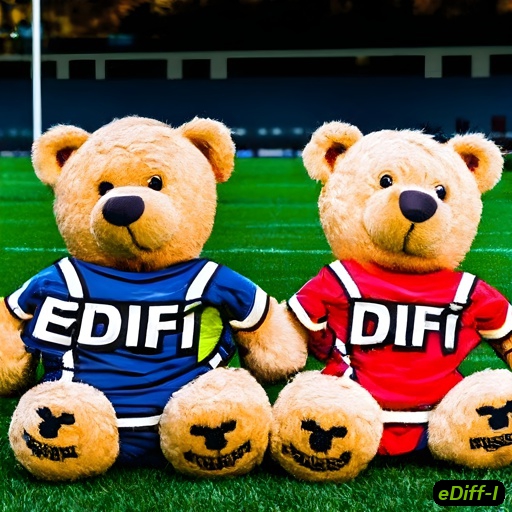} &
        \includegraphics[width=0.248\textwidth]{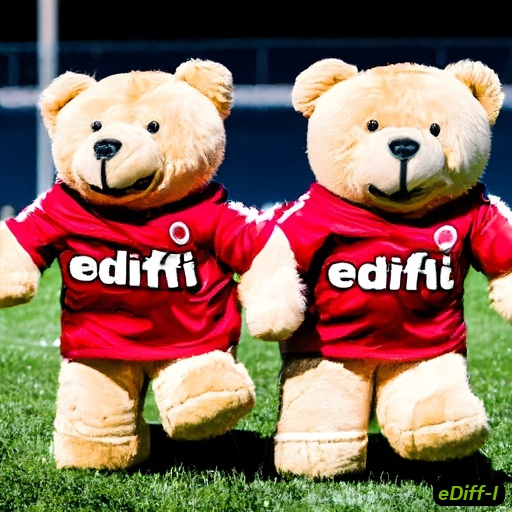} &
        \includegraphics[width=0.248\textwidth]{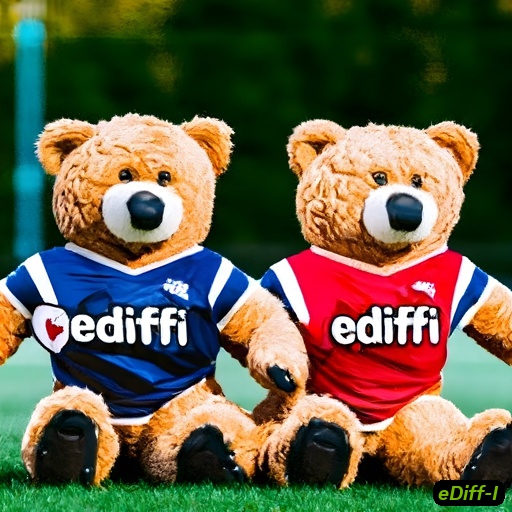} \\
        \multicolumn{4}{c}{ \emph{A 4k dslr photo of two teddy bears wearing a sports jersey with the text ``eDiffi'' written on it. They are on a soccer field.}}

    \end{tabularx}
    \caption{\textbf{2-Expert-Ensemble vs Baseline.}. We compare the image synthesis capability of our model (two rightmost panels) with our baseline (two left panels). We find that the baseline model does not generate mountains in one of the examples on the top row and produces incorrect text in the bottom row. Our 2-expert-ensemble improves the generation capability in both these cases.}
    \label{fig:comparison_ensemble}
\end{figure*}
\setlength\tabcolsep{12 pt}

%% file: tables/comparison_fid.tex
\begin{table}[t]
    \centering
    \caption{Zero-shot FID comparison with recent state-of-the-art methods on the COCO 2014 validation dataset. We include the text encoder size in our model parameter size calculation.}
    \label{tab:comparison_fid}

    \begin{tabular}{lcc}
        \toprule
        \multirow{2}{*}{Model} & $\#$ of & Zero-shot \\
            & params & FID $\downarrow$ \\
        \midrule
        GLIDE~\cite{nichol2021glide} & \s\sdot5B & 12.24\\
        Make-A-Scene~\cite{gafni2022make} & \s\sdot4B & 11.84\\
        DALL$\cdot$E~2~\cite{ramesh2022hierarchical} & 6.5B & 10.39\\
        Stable Diffusion~\cite{rombach2022high} & 1.4B & \s8.59\\
        Imagen~\cite{saharia2022photorealistic} & 7.9B & \s7.27\\
        Parti~\cite{yu2022scaling} & \sdot20B & \s7.23\\
        \midrule
        \configa & 6.8B & \s7.35\\
        \configb & 7.1B & \s7.26\\
        \configc & 8.1B & \s7.11\\
        \configd & 9.1B & \s\textbf{6.95}\\
        \bottomrule
    \end{tabular}
    \vspace{-4mm}
\end{table}

%% file: figures/comparison_multiple_entities.tex
\setlength{\tabcolsep}{0.5pt}
\renewcommand{\arraystretch}{0.5}
\begin{figure*}[ht!]
    \centering
    \begin{tabularx}{\textwidth}{c c M{0.27\textwidth} M{0.27\textwidth} M{0.27\textwidth}}
         & & (a) Stable Diffusion & (b) DALL$\cdot$E~2 & (c) Ours \\
        \makecell[X]{
            \emph{A photo of two cute teddy bears sitting on top of a grizzly bear in a beautiful forest. Highly detailed fantasy art, 4k, artstation}
        } & {\;\;\;} &
        \includegraphics[width=0.27\textwidth]{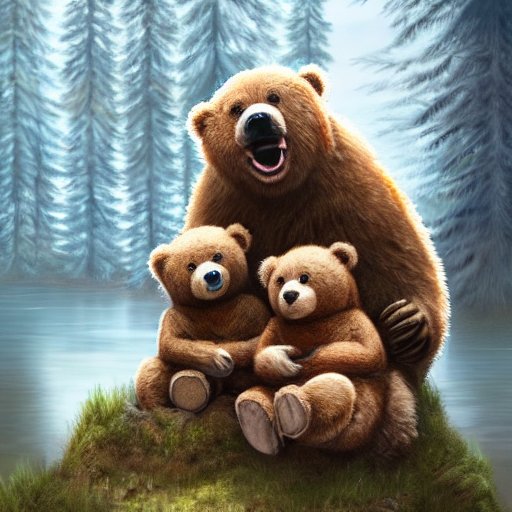} &
        \includegraphics[width=0.27\textwidth]{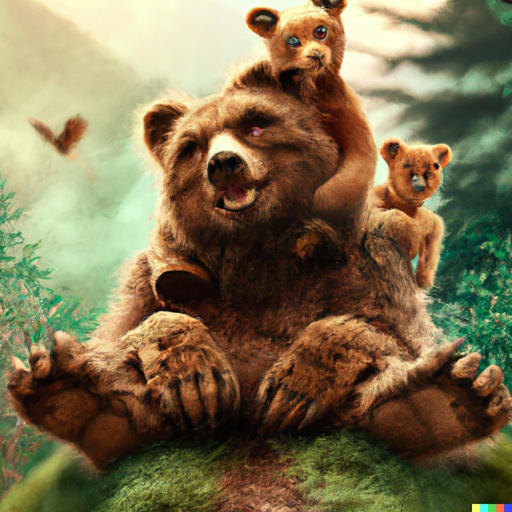} &
        \includegraphics[width=0.27\textwidth]{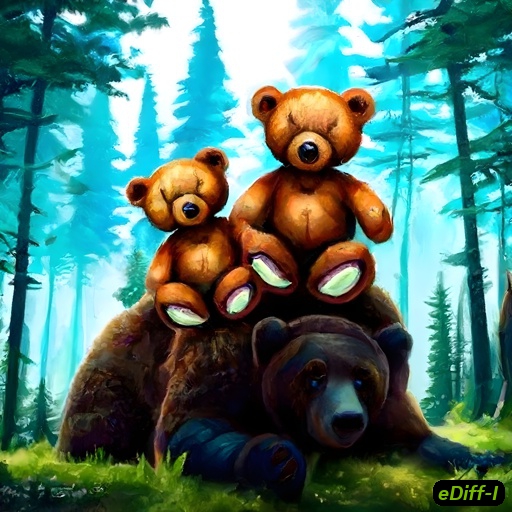} \\

        \makecell[X]{
            \emph{There are two Chinese teapots on a table. One pot has a painting of a dragon, while the other pot has a painting of a panda.}
        } & {\;\;\;} &
        \includegraphics[width=0.27\textwidth]{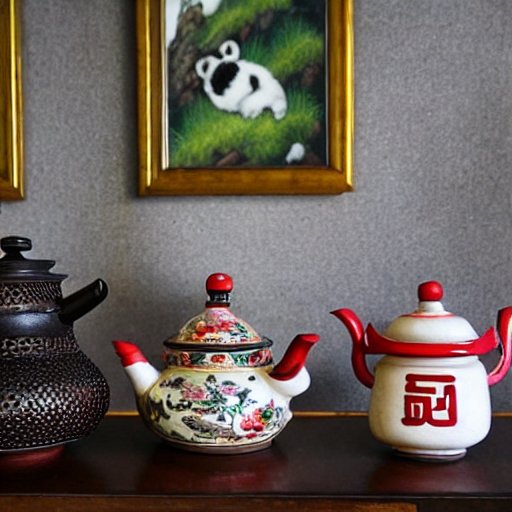} &
        \includegraphics[width=0.27\textwidth]{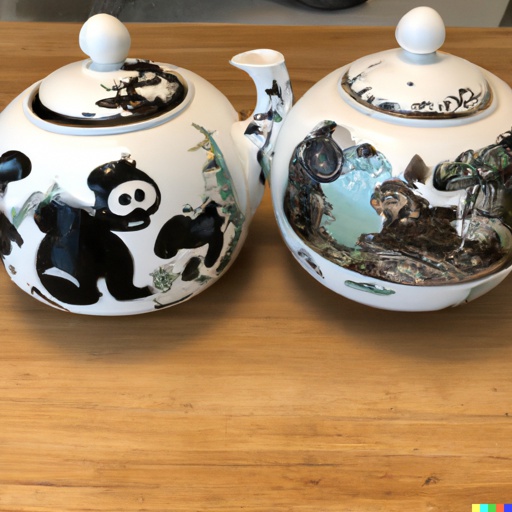} &
        \includegraphics[width=0.27\textwidth]{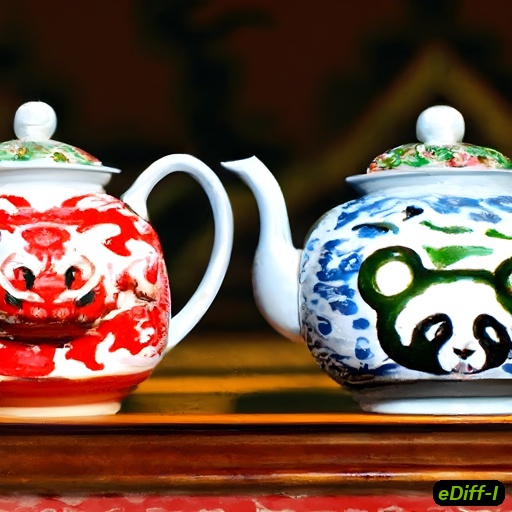} \\

        \makecell[X]{
            \emph{A photo of two squirrel warriors dressed as knights fighting on a battlefield. The squirrel on the left holds a stick, while the squirrel on the right holds a long sword. Gray clouds.}
        } & {\;\;\;} &
        \includegraphics[width=0.27\textwidth]{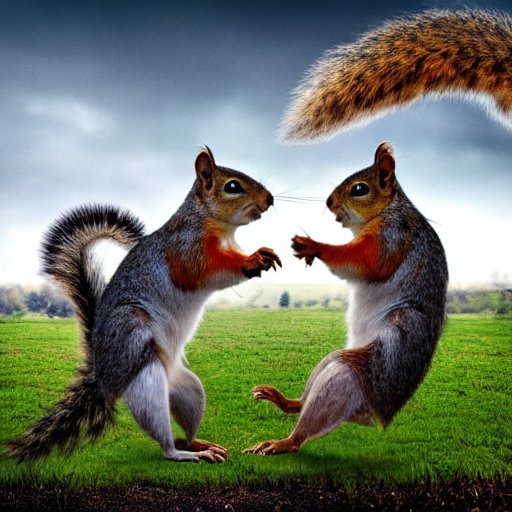} &
        \includegraphics[width=0.27\textwidth]{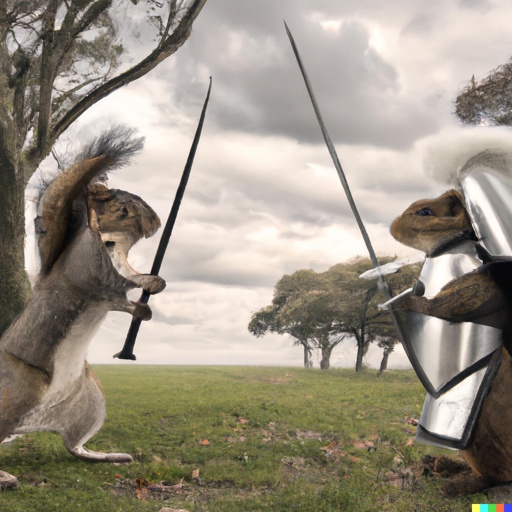} &
        \includegraphics[width=0.27\textwidth]{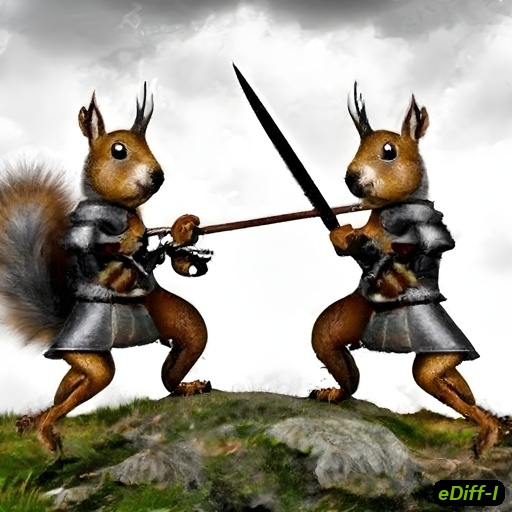} \\

        \makecell[X]{
            \emph{A photo of a dog wearing a blue shirt and a cat wearing a red shirt sitting in a park, photorealistic dslr.}
        } & {\;\;\;} &
        \includegraphics[width=0.27\textwidth]{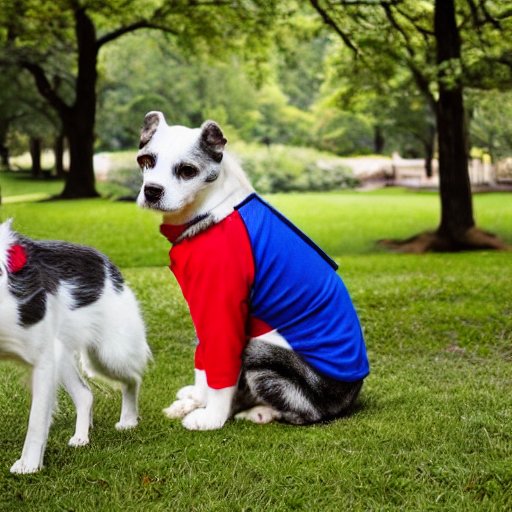} &
        \includegraphics[width=0.27\textwidth]{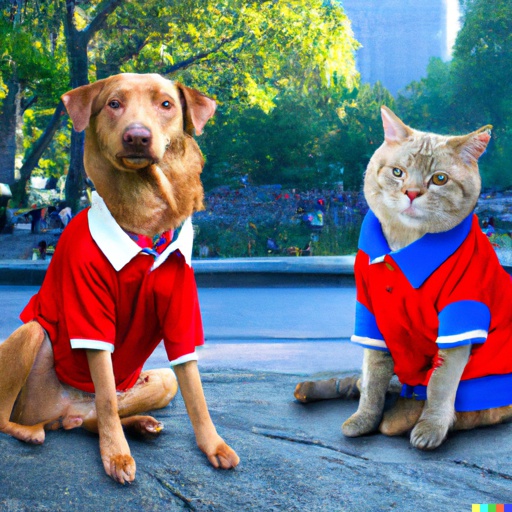} &
        \includegraphics[width=0.27\textwidth]{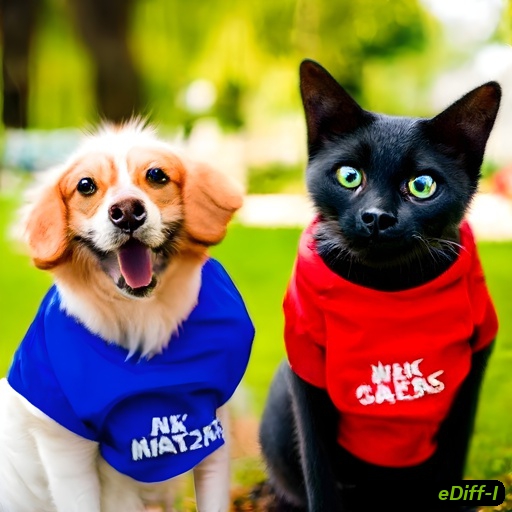} \\

    \end{tabularx}
    \caption{Comparison between samples generated by our approach with DALL$\cdot$E~2 and Stable Diffusion. Category: multiple entities. In this set of results, we generate images with multiple entities. Stable Diffusion and DALL$\cdot$E~2 tend to mix the attributes corresponding to different entities. For instance, in the last row, the dog and the cat both wear red shirts in DALL$\cdot$E~2 outputs, while Stable Diffusion does not even generate a cat. Our model, on the other hand, produces all entities consistent with the captions. Note that we generated multiple outputs from each method and picked the best one to include in the figure.}
    \label{fig:comparison_multiple_entities}
\end{figure*}

%% file: figures/comparison_text.tex
\setlength{\tabcolsep}{0.5pt}
\renewcommand{\arraystretch}{0.5}
\begin{figure*}[ht!]
    \centering
    \begin{tabularx}{\textwidth}{c c M{0.27\textwidth} M{0.27\textwidth} M{0.27\textwidth}}
         & & (a) Stable Diffusion & (b) DALL$\cdot$E~2 & (c) Ours \\
        \makecell[X]{
            \emph{A photo of two monkeys sitting on a tree. They are holding a wooden board that says ``Best friends'', 4K dslr.}
        } & {\;\;\;} &
        \includegraphics[width=0.27\textwidth]{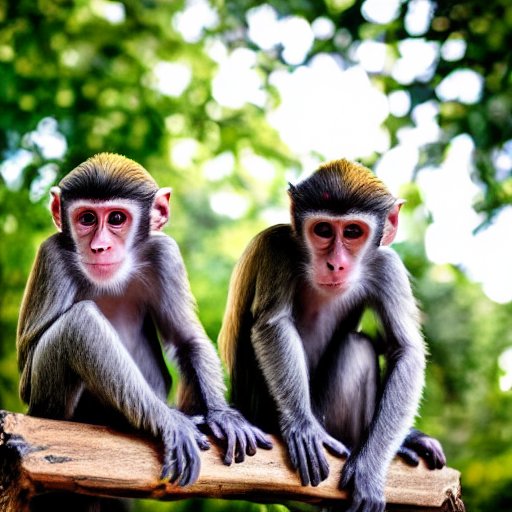} &
        \includegraphics[width=0.27\textwidth]{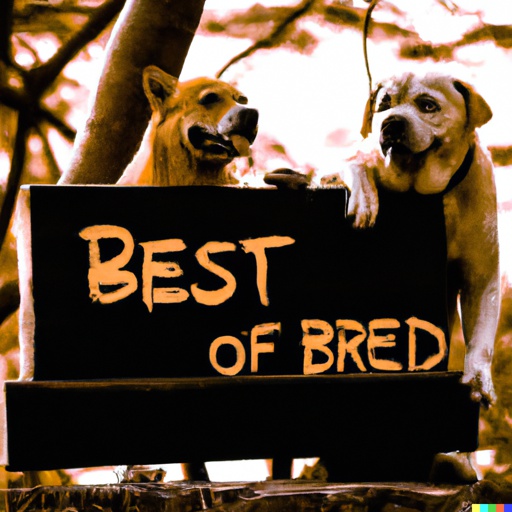} &
        \includegraphics[width=0.27\textwidth]{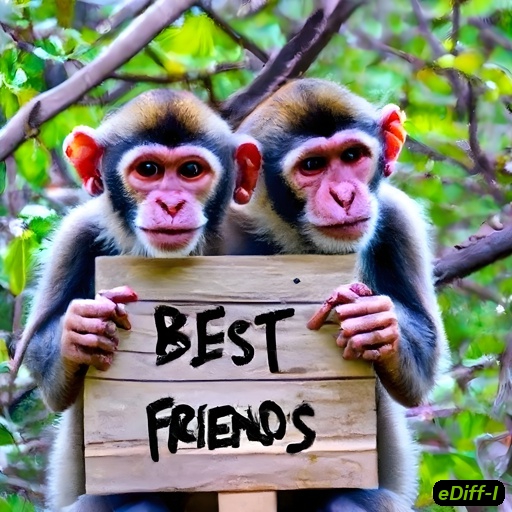} \\

        \makecell[X]{
            \emph{A photo of a golden retriever puppy wearing a green shirt. The shirt has text that says ``NVIDIA rocks''. Background office. 4k dslr.}
        } & {\;\;\;} &
        \includegraphics[width=0.27\textwidth]{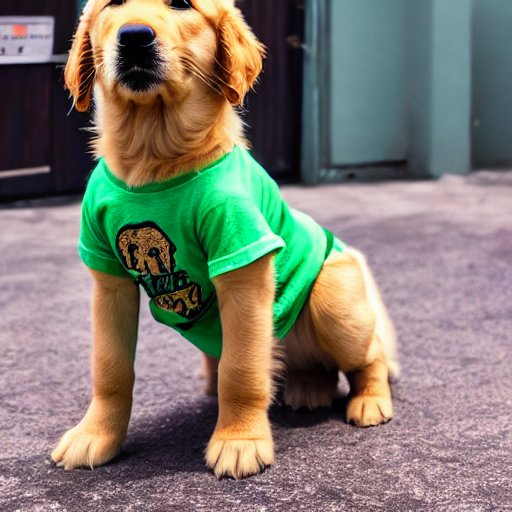} &
        \includegraphics[width=0.27\textwidth]{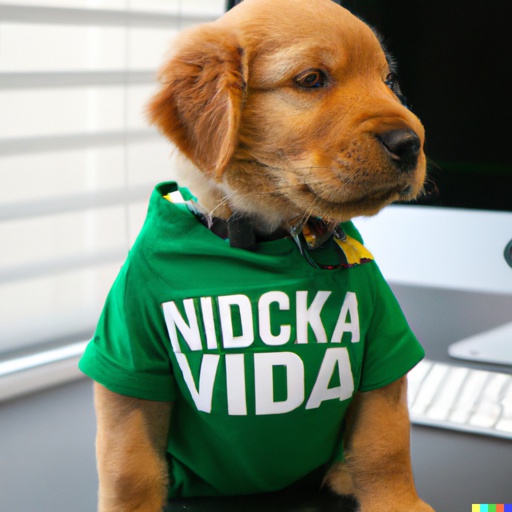} &
        \includegraphics[width=0.27\textwidth]{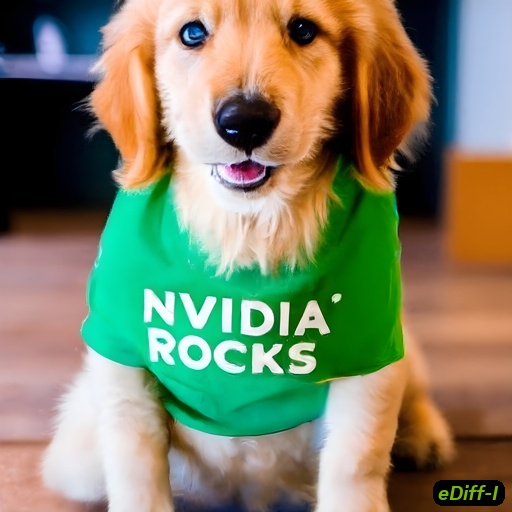} \\

        \makecell[X]{
            \emph{An ice sculpture is made with the text ``Happy Holidays''. Christmas decorations in the background. Dslr photo.}
        } & {\;\;\;} &
        \includegraphics[width=0.27\textwidth]{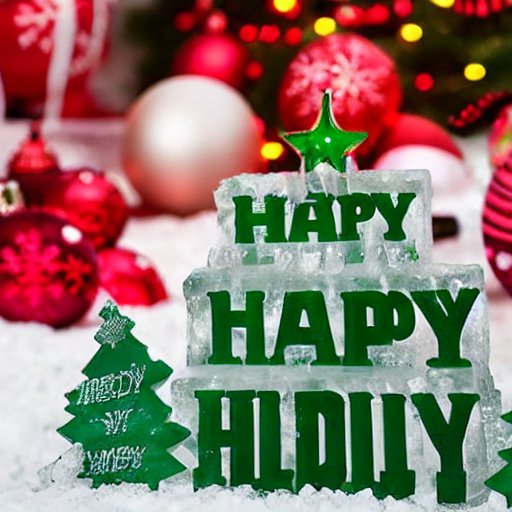} &
        \includegraphics[width=0.27\textwidth]{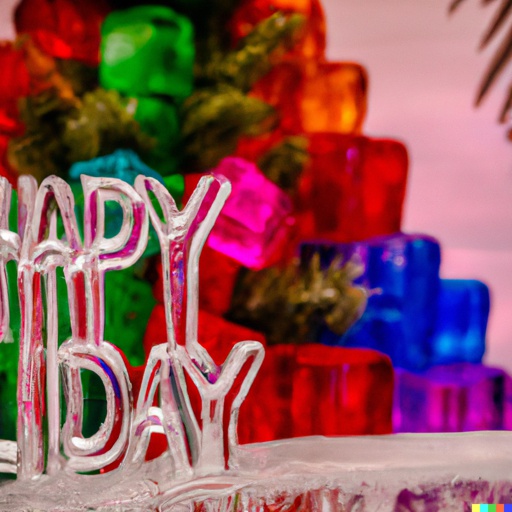} &
        \includegraphics[width=0.27\textwidth]{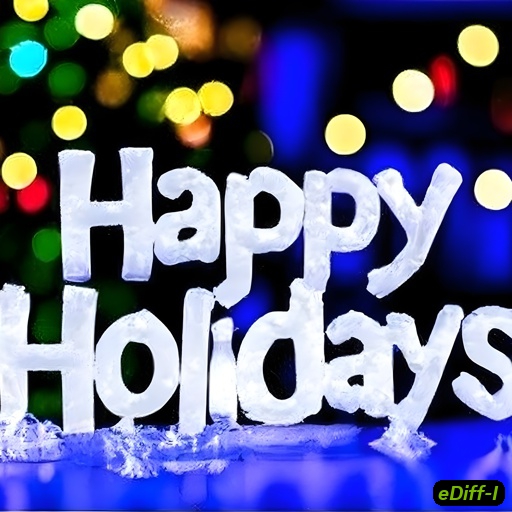} \\

        \makecell[X]{
            \emph{A dslr photo of a colorful Graffiti on a wall with the text ``Peace love''. There is a painting of a bulldog with sunglasses next to it.}
        } & {\;\;\;} &
        \includegraphics[width=0.27\textwidth]{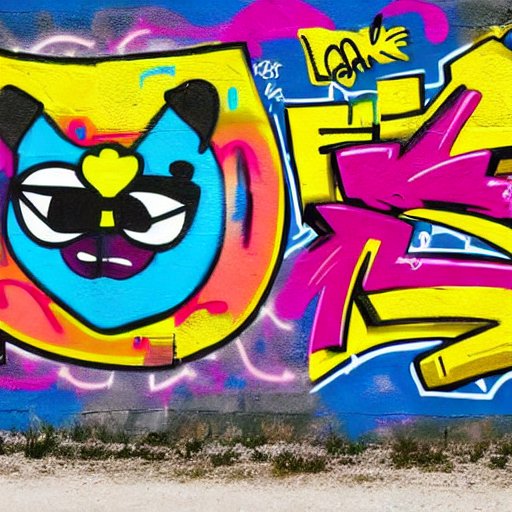} &
        \includegraphics[width=0.27\textwidth]{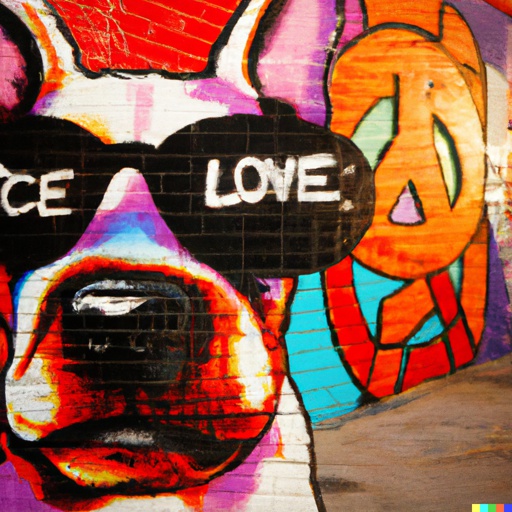} &
        \includegraphics[width=0.27\textwidth]{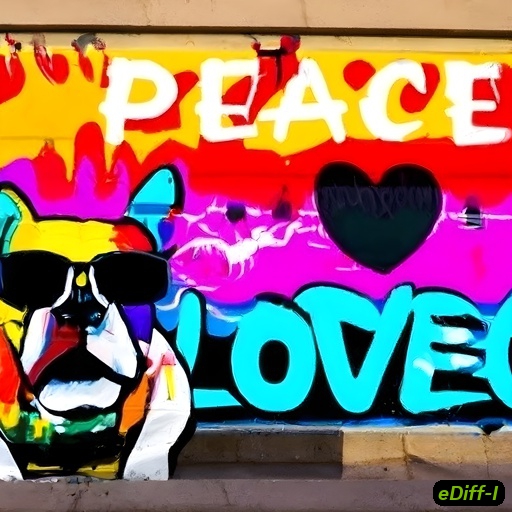} \\

    \end{tabularx}
    \caption{Comparison between samples generated by our approach with DALL$\cdot$E~2 and Stable Diffusion. Category: text. We observe that both DALL$\cdot$E~2 and Stable Diffusion fail at generating text. These models either produce misspellings or do not even generate text. For example, as spelling ``NVIDIA rocks'', Stable Diffusion failed to generate any text, while DALL$\cdot$E~2 only generated ``NIDCKA VIDA''. Only our model succeeds in generating the target. Overall, we observe that our model can generate English text on a wide range of samples. Note that we generated multiple outputs from each method and picked the best one to include in the figure.}
    \label{fig:comparison_text}
\end{figure*}

%% file: figures/comparison_long_captions.tex
\setlength{\tabcolsep}{0.5pt}
\renewcommand{\arraystretch}{0.5}
\begin{figure*}[ht!]
    \centering
    \begin{tabularx}{\textwidth}{c c M{0.27\textwidth} M{0.27\textwidth} M{0.27\textwidth}}
         & & (a) Stable Diffusion & (b) DALL$\cdot$E~2 & (c) Ours \\
        \makecell[X]{
            \emph{A 4K dslr photo of a hedgehog sitting in a small boat in the middle of a pond. It is wearing a Hawaiian shirt and a straw hat. It is reading a book. There are a few leaves in the background.}
        } & {\;\;\;} &
        \includegraphics[width=0.27\textwidth]{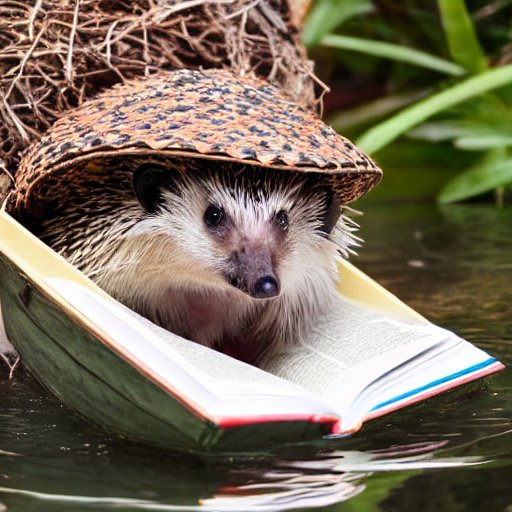} &
        \includegraphics[width=0.27\textwidth]{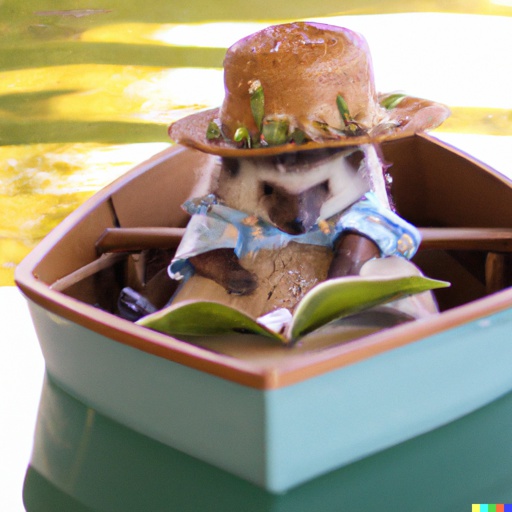} &
        \includegraphics[width=0.27\textwidth]{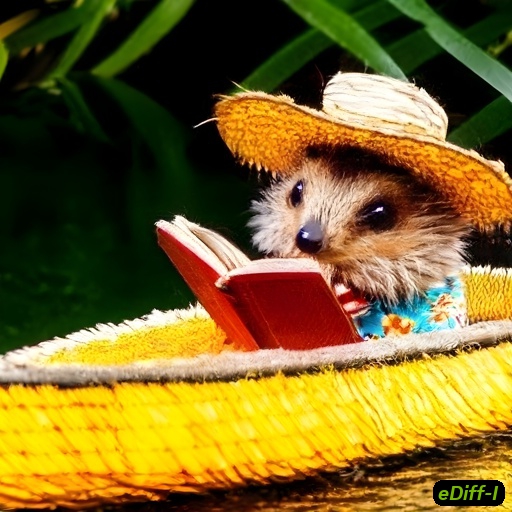} \\

        \makecell[X]{
            \emph{A fantasy landscape on an alien planet in which there are many buildings. There is a beautiful bridge with a pond in the center. There is one large moon in the sky. The sky is orange. Digital art, artstation.}
        } & {\;\;\;} &
        \includegraphics[width=0.27\textwidth]{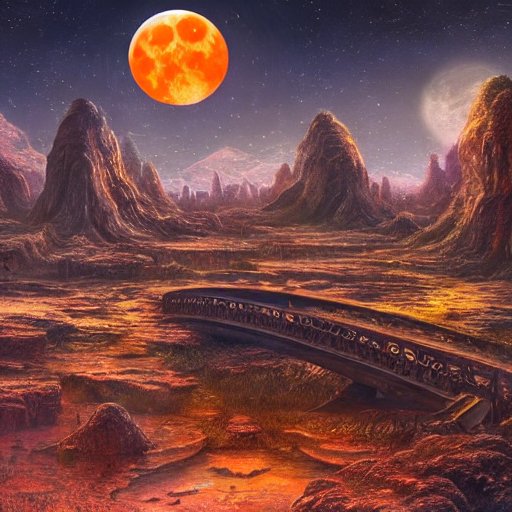} &
        \includegraphics[width=0.27\textwidth]{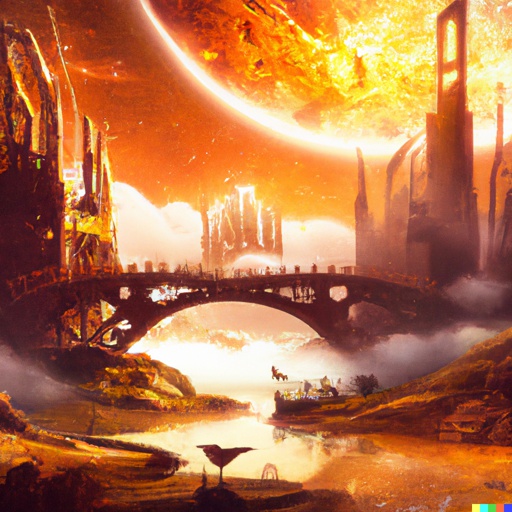} &
        \includegraphics[width=0.27\textwidth]{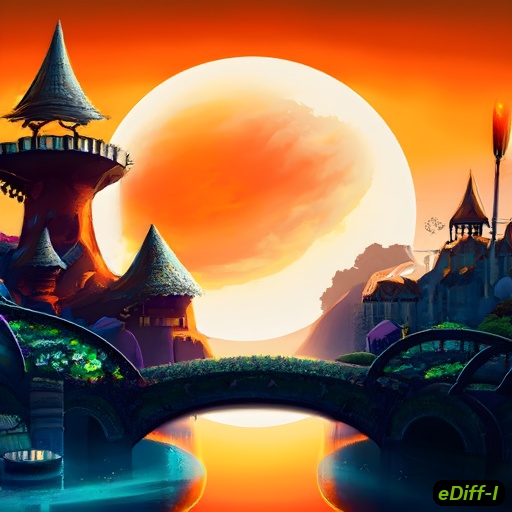} \\

        \makecell[X]{
            \emph{A close-up 4k dslr photo of a cat riding a scooter. It is wearing a plain shirt and has a bandana around its neck. It is wearing a scooter helmet. There are palm trees in the background.}
        } & {\;\;\;} &
        \includegraphics[width=0.27\textwidth]{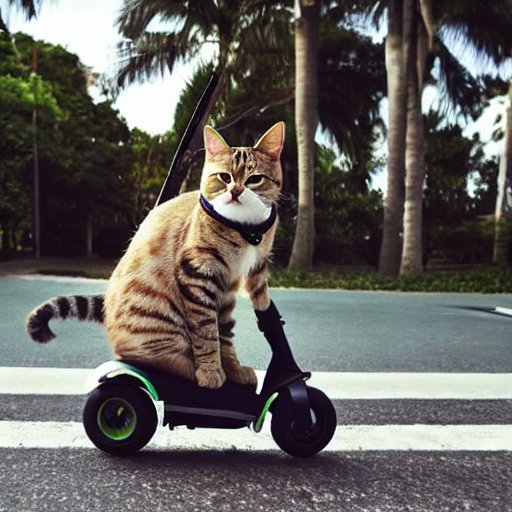} &
        \includegraphics[width=0.27\textwidth]{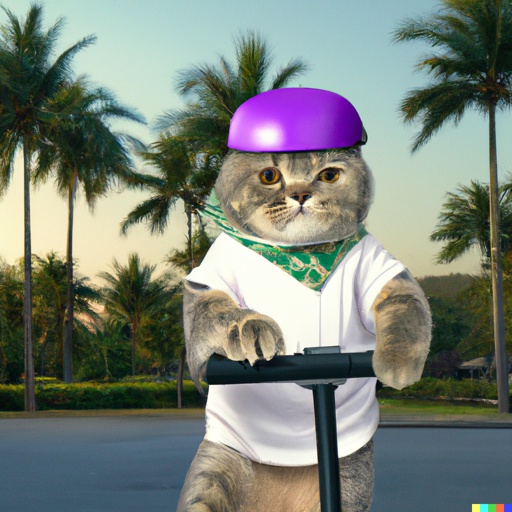} &
        \includegraphics[width=0.27\textwidth]{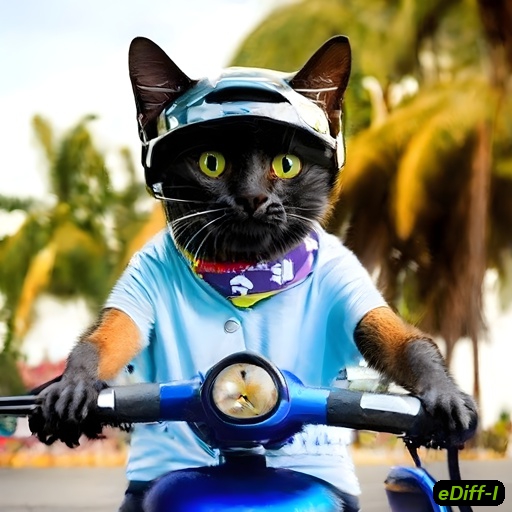} \\

        \makecell[X]{
            \emph{A photo of a plate at a restaurant table with spaghetti and red sauce. There is sushi on top of the spaghetti. The dish is garnished with mint leaves. On the side, there is a glass with a purple drink, photorealistic, dslr.}
        } & {\;\;\;} &
        \includegraphics[width=0.27\textwidth]{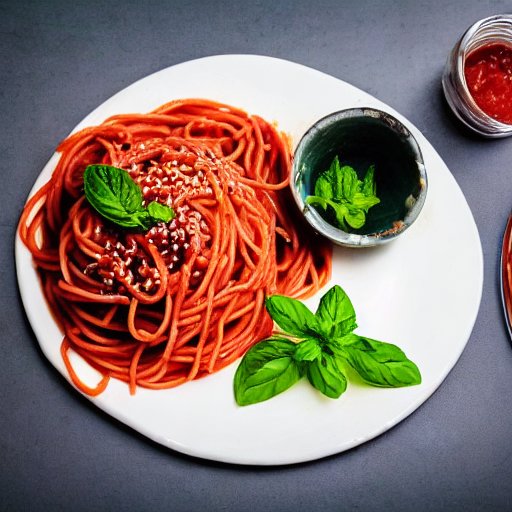} &
        \includegraphics[width=0.27\textwidth]{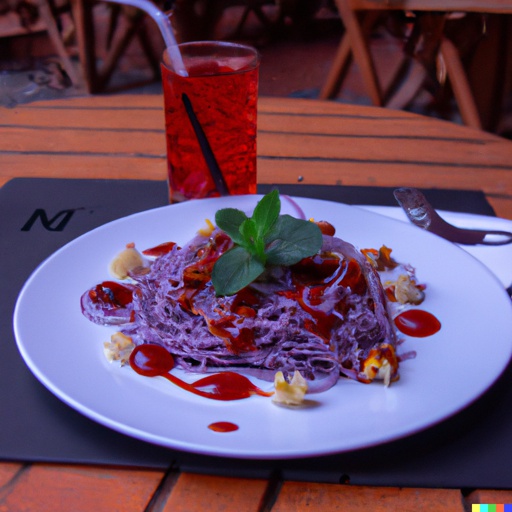} &
        \includegraphics[width=0.27\textwidth]{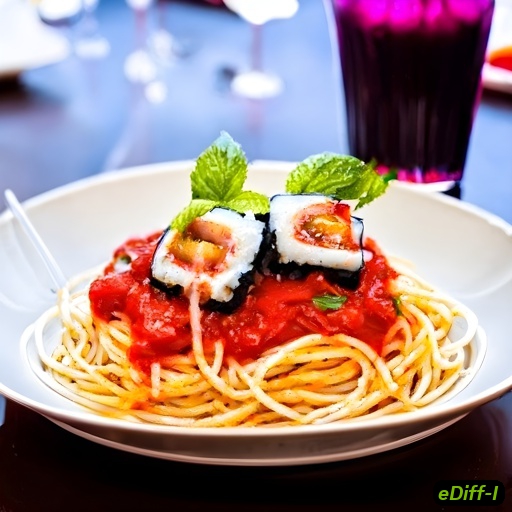} \\

    \end{tabularx}
    \caption{Comparison between samples generated by our approach with DALL$\cdot$E~2 and Stable Diffusion. Category: long detailed captions. On long descriptions, Stable Diffusion and DALL$\cdot$E~2 sometimes fail to create images with all attributes mentioned in the caption. For instance, in the last example, neither method generates sushi on top of pasta. Our method can handle long detailed descriptions better than other methods. Note that we generated multiple outputs from each method and picked the best one to include in the figure.}
    \label{fig:comparison_long_captions}
\end{figure*}

%% file: figures/curve_speed_comparison_ensmble_depth.tex
\input{figures/model_speed_curves/data/depth_increment.tex}
\input{figures/model_speed_curves/data/ensemble.tex}
\renewcommand{\plothh}{90mm}
\renewcommand{\plotvv}{55mm}
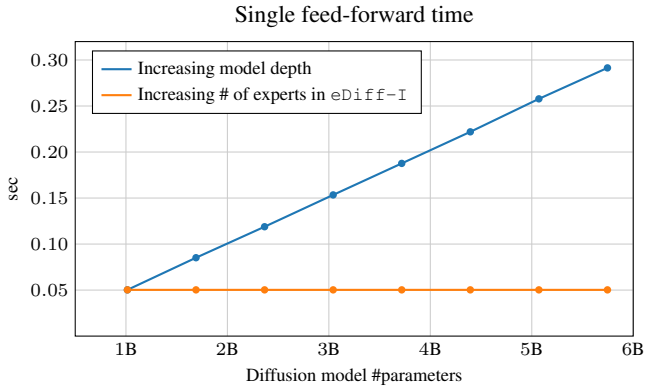
\begin{figure}[t!]
\vspace{-4mm}
\centering%
\small%
\scriptsize%
\hspace*{-1mm}%
\pgfplotsset{scaled x ticks=false}
\begin{tikzpicture}%
\begin{axis}[
  width=\plothh, height=\plotvv,
  title={\small Single feed-forward time}, title style={yshift=-1ex},
  xmin={500000000}, xmax={6000000000}, xmode={linear}, xtick={1000000000, 2000000000, 3000000000, 4000000000, 5000000000, 6000000000}, xticklabels={$1$B, $2$B, $3$B, $4$B, $5$B, $6$B},
  ymin={0}, ymax={0.32}, ymode={linear}, ytick={0.05, 0.10, 0.15, 0.20, 0.25, 0.30}, yticklabels={$0.05$, $0.10$, $0.15$, $0.20$, $0.25$, $0.30$},
  xlabel={Diffusion model \#parameters}, xlabel near ticks,
  ylabel={sec}, ylabel near ticks,
  grid={major}, legend pos={north west}, legend cell align={left}, legend style={font=\scriptsize},
]
\addplot[C0] coordinates {\genDepthIncrement};
\addplot[C1] coordinates {\genEnsembleIncrement};
\addplot[C0, only marks, forget plot] coordinates {\genDepthIncrement};
\addplot[C1, only marks, forget plot] coordinates {\genEnsembleIncrement};
\legend{
    {Increasing model depth},
    {Increasing \# of experts in \ours},
}
\end{axis}
\end{tikzpicture}%
\vspace*{-.5ex}\\
\vspace{-1mm}
\caption{\label{fig:fig_speed_comparison_curves}%
\textbf{Model inference time by base model capacity.} In this figure, we compare the network feed-forward evaluation time for two schemes of increasing network capacity in diffusion models. The first scheme is through increasing the depth of the model. We use the same base model architecture but increase the number of residual blocks there. We perform feed-forward evaluations 100 times on an NVIDIA A100 GPU and report the average feed-forward time in the curve. As expected, the feed-forward evaluation time increases with the network capacity. The second scheme is our proposed scheme, where we simply use different numbers of denoising experts for different noise levels, and hence the per-network feed-forward evaluation time remains constant.}
\vspace{-4mm}
\end{figure}

%% file: figures/model_speed_curves/data/depth_increment.tex
\newcommand{\genDepthIncrement}{%
(1014514947, 0.050232151)
(1690457091, 0.08516350612)
(2366399235, 0.11883393)
(3042341379, 0.1534057923)
(3718283523, 0.1876480108)
(4394225667, 0.2219442295)
(5070167811, 0.2577783167)
(5746109955, 0.2914138855)
}

%% file: figures/model_speed_curves/data/ensemble.tex
\newcommand{\genEnsembleIncrement}{%
(1014514947, 0.050232151)
(1690457091, 0.050232151)
(2366399235, 0.050232151)
(3042341379, 0.050232151)
(3718283523, 0.050232151)
(4394225667, 0.050232151)
(5070167811, 0.050232151)
(5746109955, 0.050232151)
}

%% file: sections/conc.tex
\section{Conclusions}\label{sec:conclusion}

In this paper, we proposed \ours, a state-of-the-art text-to-image diffusion model that consists of a base diffusion model and two super-resolution modules, producing $1024\times1024$ high-definition outputs. \ours utilizes an ensemble of expert denoisers to achieve superior performance compared to previous work. We found that the generation process in text-to-image diffusion models qualitatively changes throughout synthesis: Initially, the model focuses on generating globally coherent content aligned with the text prompt, while later in the process, the model largely ignores the text conditioning and its primary goal is to produce visually high-quality outputs. Our different expert denoiser networks allow us to specialize the model for different behaviors during different intervals of the iterative synthesis process. 
Moreover, we showed that by conditioning on both T5 text, CLIP text, and CLIP image embeddings, \ours not only enjoys improved performance but also enables rich controllability. In particular, the T5 and CLIP text embeddings capture complementary aspects of the generated images, and the CLIP image embedding further can be used for stylization according to reference images. Finally, we demonstrated expressive spatial control using \ours's ``paint-with-words'' capability.

\paragraph{Societal impact} We hope that \ours can serve as a powerful tool for digital artists for content creation and to express their creativity freely. Modern text-to-image diffusion models like ours have the potential to democratize artistic expression by offering the user the ability to produce detailed and high-quality imagery without the need for specialized skills. We envision that \ours can benefit designers, photographers, and content creators.

However, state-of-the-art text-to-image generative models like \ours need to be applied with an abundance of caution. For instance, they can also be used for advanced photo manipulation for malicious purposes or to create deceptive or harmful content. In fact, the recent progress of generative models and AI-driven image editing has profound implications for image authenticity and beyond. Such challenges can potentially be tackled, for instance, by methods that automatically validate real images and detect manipulated or fake content. Moreover, the extremely large, mostly unfiltered training data sets of current large-scale text-to-image generative models include biases that are captured by the model and also reflected in the generated data. It is, therefore, important to be aware of such biases in the underlying data and counteract them, for example, by actively collecting more representative data or by using bias correction methods.

\paragraph{Acknowledgements}

We would like to thank Qinsheng Zhang, Robin Rombach, Chen-Hsuan Lin, Mohammad Shoeybi, Tsung-Yi Lin, Wei Ping, Mostofa Patwary, Andrew Tao, Guilin Liu, Vijay Korthikanti, Sanja Fidler, David Luebke, and Jan Kautz for useful discussions. We would also like to thank Margaret Albrecht and Greg Estes for helping iterate our presentation. Thanks also go to Gabriele Leone and Ashlee Martino-Tarr, who are our valuable early testers of \ours. Finally, we would like to thank Amanda Moran, John Dickinson, and Sivakumar Arayandi Thottakara for the computing infrastructure support.

\input{figures/style_variants}
\input{figures/variations}
\input{figures/clip_img_conditioning}
\input{figures/paint_with_words_results}

\input{figures/t5_vs_clip}

%% file: figures/style_variants.tex
\setlength{\tabcolsep}{0.5pt}
\renewcommand{\arraystretch}{0.5}
\begin{figure*}[ht!]
    \centering
    \begin{tabular}{cccc}
        \includegraphics[width=0.33\textwidth]{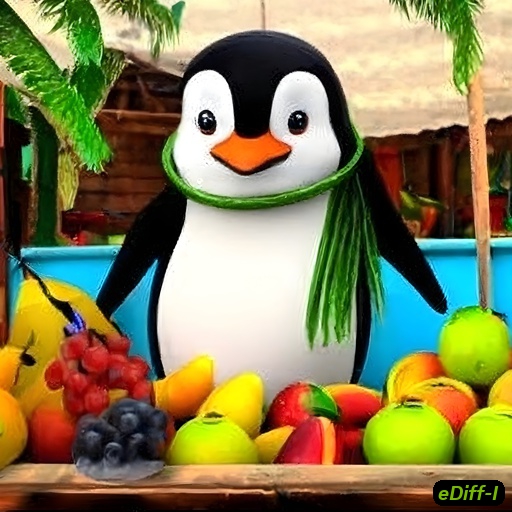} &
        \includegraphics[width=0.33\textwidth]{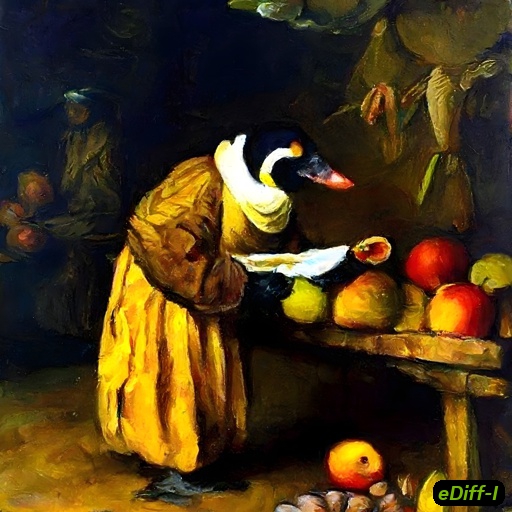} &
        \includegraphics[width=0.33\textwidth]{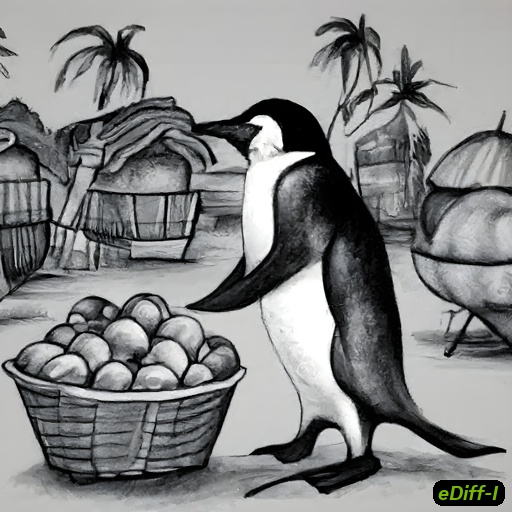} \\
        {\centering Real} & {\centering Rembrandt} & {\centering Pencil sketch} \\[3mm]

        \includegraphics[width=0.33\textwidth]{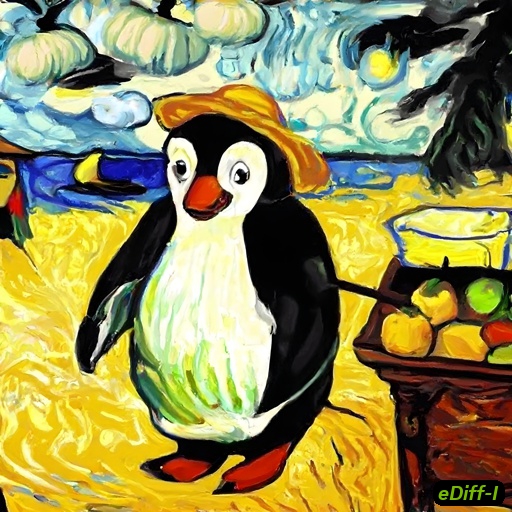} &
        \includegraphics[width=0.33\textwidth]{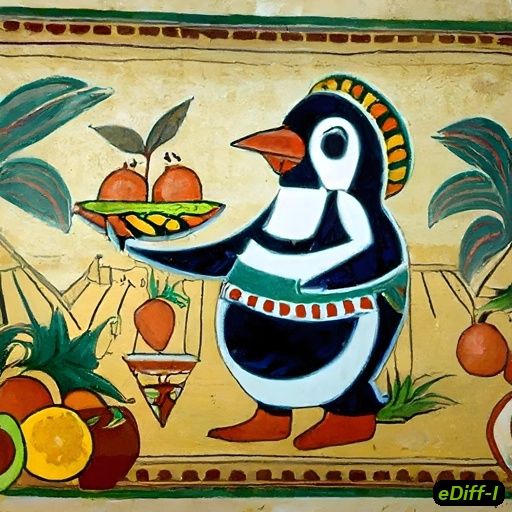} &
        \includegraphics[width=0.33\textwidth]{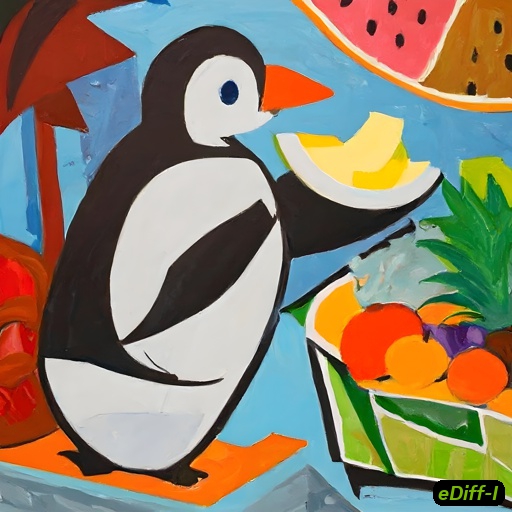} \\
        {\centering Vincent van Gogh} & {\centering Egyptian tomb hieroglyphics} & {\centering Abstract cubism} \\[3mm]

        \includegraphics[width=0.33\textwidth]{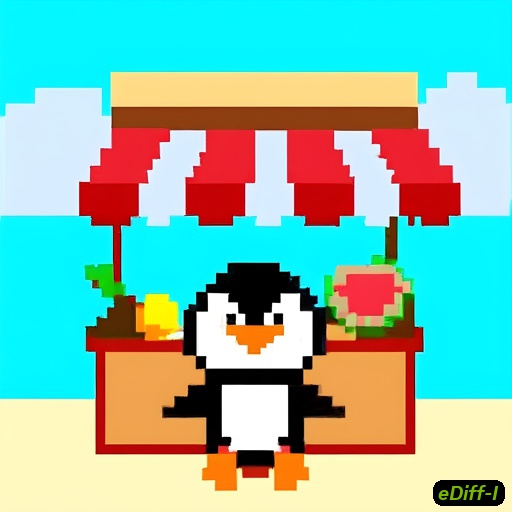} &
        \includegraphics[width=0.33\textwidth]{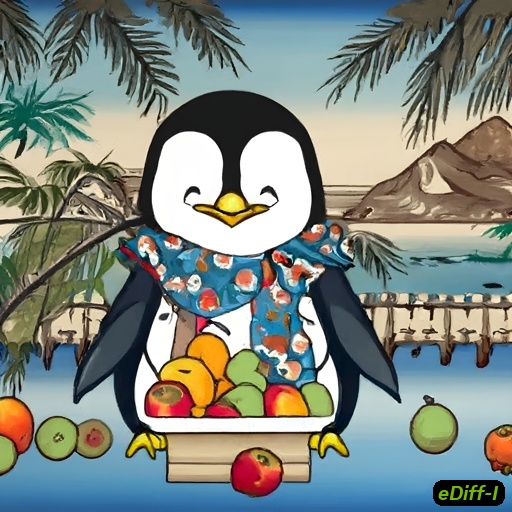} &
        \includegraphics[width=0.33\textwidth]{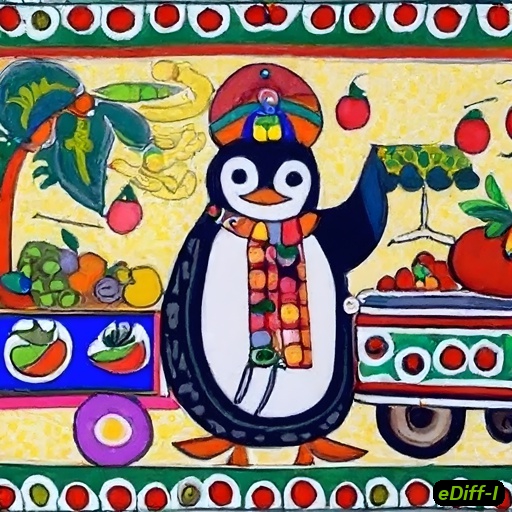} \\
        {\centering Pixel art} & {\centering Hokusai} & {\centering Madhubani} \\

    \end{tabular}
    \caption{Samples generated by our approach for the caption ``A \{X\} photo / painting of a penguin working as a fruit vendor in a tropical village'', where X denotes one of the nine styles listed above. Our model can generate images from a wide range of artistic styles while being consistent with the input text description.}
    \label{fig:style_variants}
    \vspace{25mm}
\end{figure*}

%% file: figures/variations.tex
\setlength{\tabcolsep}{0.5pt}
\renewcommand{\arraystretch}{0.5}
\begin{figure*}[ht!]
    \centering
    \begin{tabularx}{\textwidth}{c c M{0.21\textwidth} M{0.21\textwidth} M{0.21\textwidth} M{0.21\textwidth}}
        \makecell[X]{
            \emph{\footnotesize A photo of a dog wearing a blue shirt and a cat wearing a red shirt sitting in a park, photorealistic dslr.}
        } & {\;\;\;} &
        \includegraphics[width=0.21\textwidth]{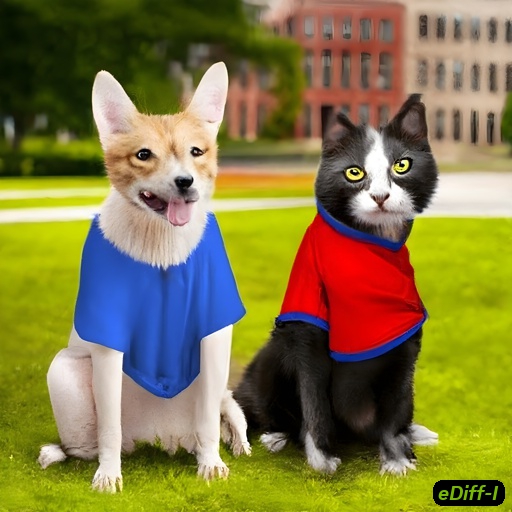} &
        \includegraphics[width=0.21\textwidth]{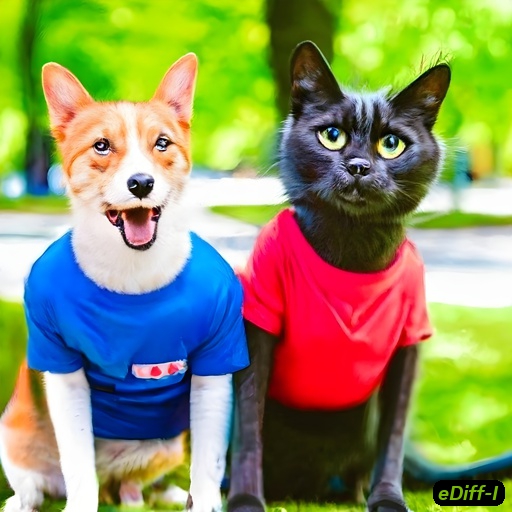} &
        \includegraphics[width=0.21\textwidth]{figures/variations/dog_cat/dog_cat2_4k_512_cr95_watermark.jpg} &
        \includegraphics[width=0.21\textwidth]{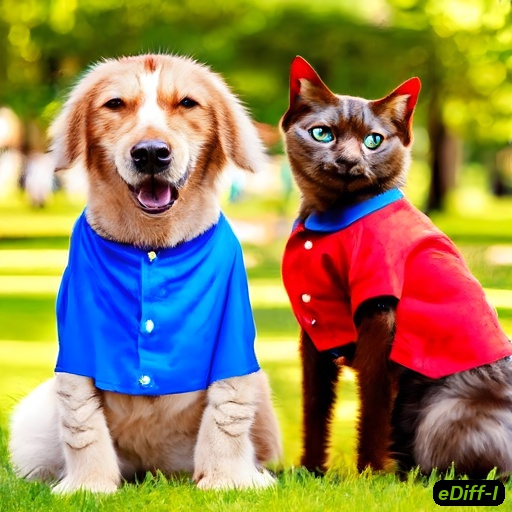} \\

        \makecell[X]{
            \emph{\footnotesize A photo of a hedgehog sitting in a small boat in the middle of a pond. It is wearing a Hawaiian shirt and a straw hat. It is reading a book.}
        } & {\;\;\;} &
        \includegraphics[width=0.21\textwidth]{figures/comparison_long/ours_hedgehog_4k_512_cr95_watermark.jpg} &
        \includegraphics[width=0.21\textwidth]{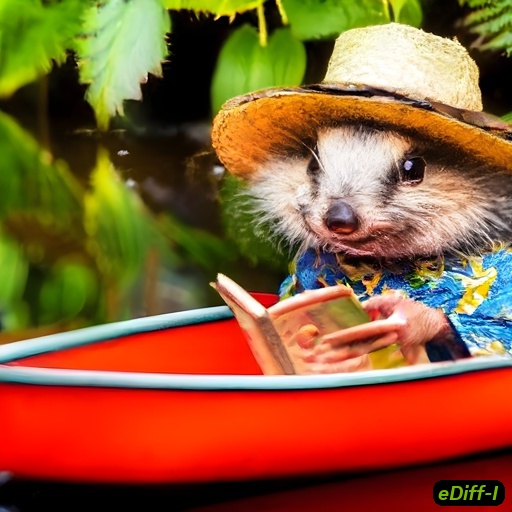} &
        \includegraphics[width=0.21\textwidth]{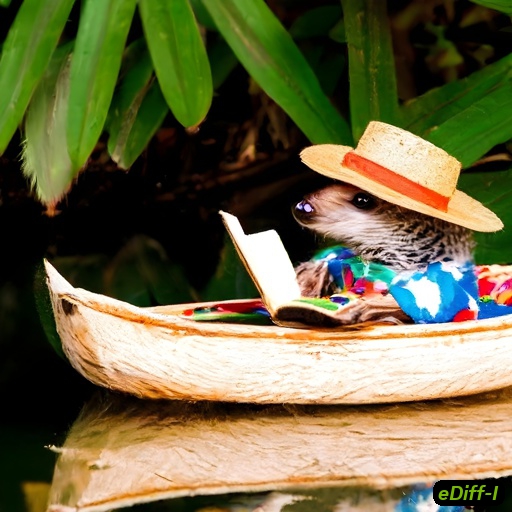} &
        \includegraphics[width=0.21\textwidth]{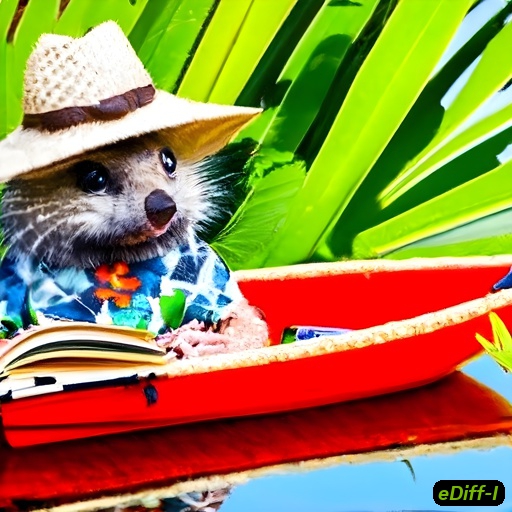} \\

        \makecell[X]{
            \emph{\footnotesize An ice sculpture is made with text ``Happy Holidays''. Christmas decorations in the background. Dslr photo.}
        } & {\;\;\;} &
        \includegraphics[width=0.21\textwidth]{figures/comparison_text/ours_happy_holidays_4k_512_cr95_watermark.jpg} &
        \includegraphics[width=0.21\textwidth]{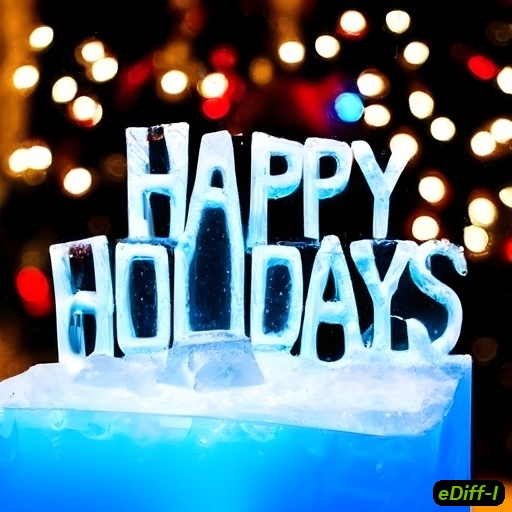} &
        \includegraphics[width=0.21\textwidth]{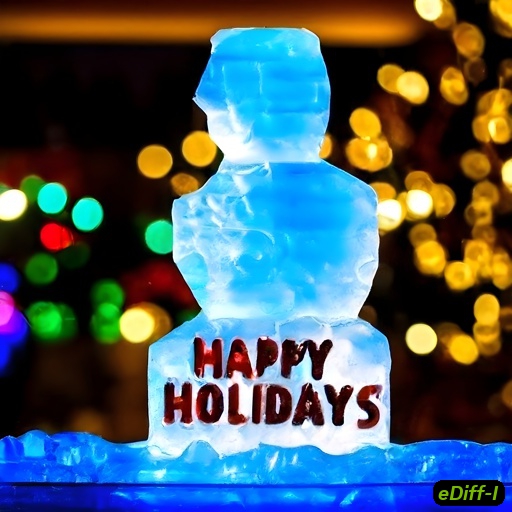} &
        \includegraphics[width=0.21\textwidth]{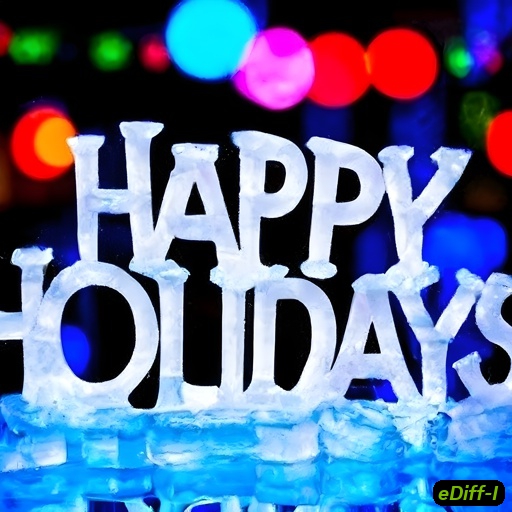} \\

        \makecell[X]{
            \emph{\footnotesize An oil painting of raccoons dressed as three musketeers in an old French town.}
        } & {\;\;\;} &
        \includegraphics[width=0.21\textwidth]{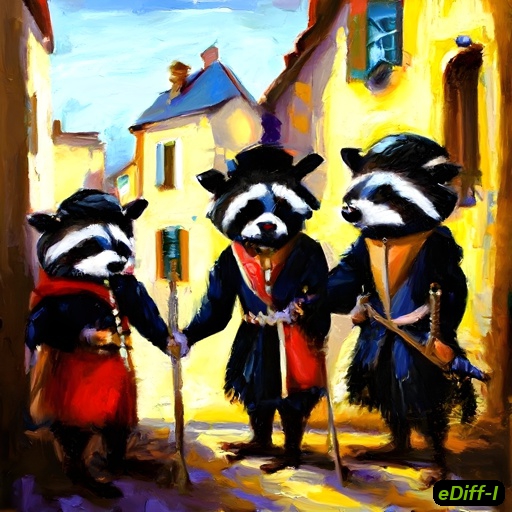} &
        \includegraphics[width=0.21\textwidth]{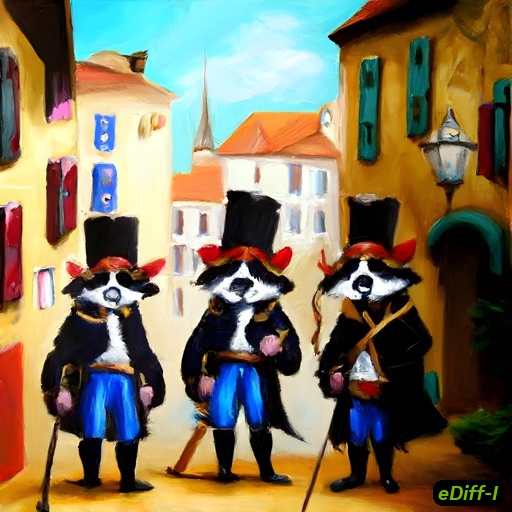} &
        \includegraphics[width=0.21\textwidth]{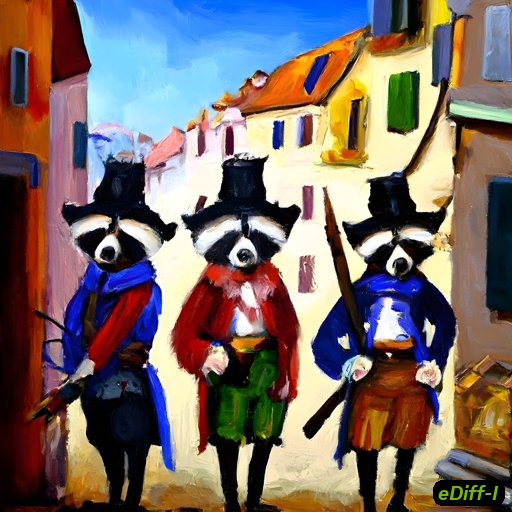} &
        \includegraphics[width=0.21\textwidth]{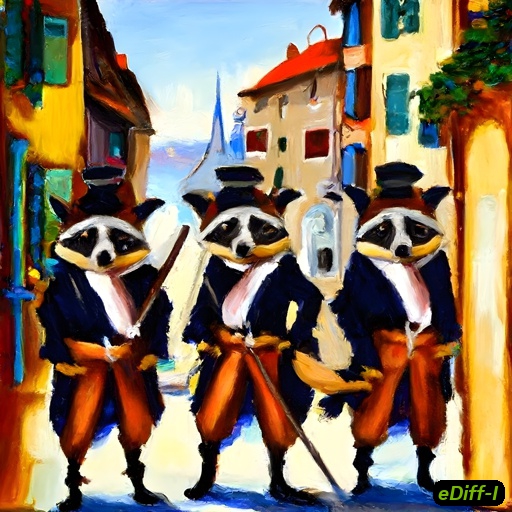} \\
        
        \makecell[X]{
            \emph{\footnotesize A photo of a lemur wearing a red magician hat and a blue coat in a garden. The lemur is performing a magic trick with cards.}
        } & {\;\;\;} &
        \includegraphics[width=0.21\textwidth]{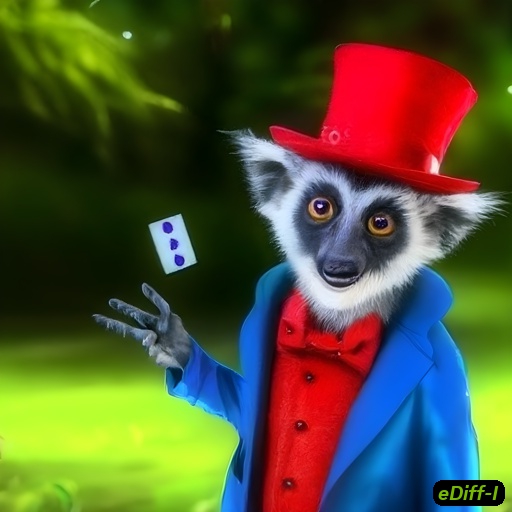} &
        \includegraphics[width=0.21\textwidth]{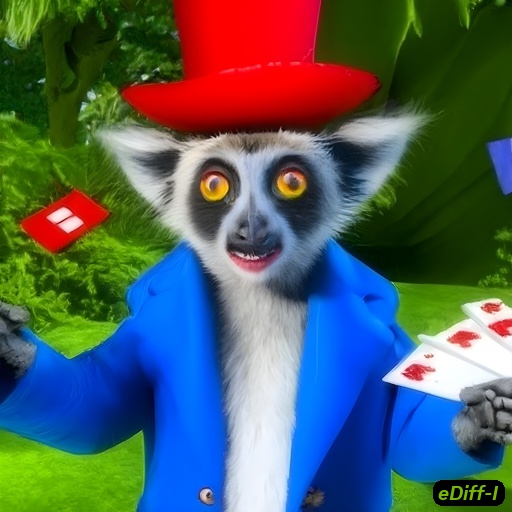} &
        \includegraphics[width=0.21\textwidth]{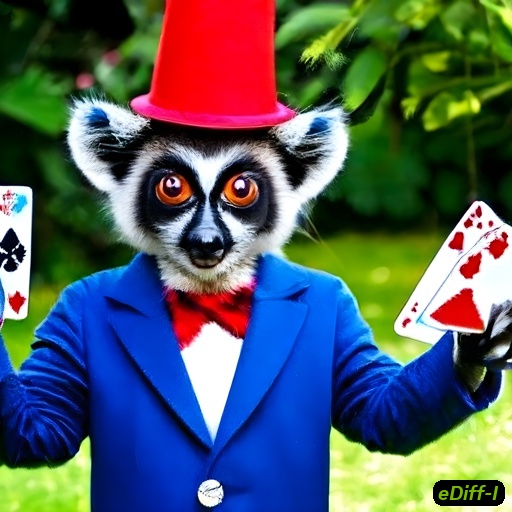} &
        \includegraphics[width=0.21\textwidth]{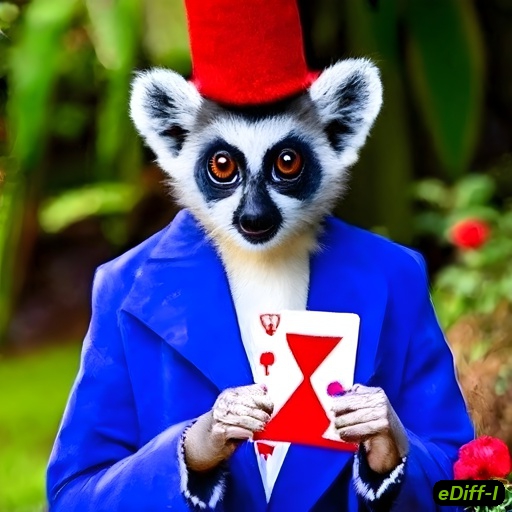} \\

    \end{tabularx}
    \caption{Image variations generated by our method. For each text prompt, we show four random sample variations generated by our method. We find that our model can generate diverse samples for each prompt. In the first row, our model can generate different breeds of dogs and cats, whereas, in the third row, we see that our model can generate the text ``Happy Holidays'' with different styles.}
    \label{fig:variations}
    \vspace{5mm}
\end{figure*}

%% file: figures/clip_img_conditioning.tex
\setlength{\tabcolsep}{0.5pt}
\renewcommand{\arraystretch}{0.5}
\begin{figure*}[ht!]
    \centering
    \begin{tabularx}{\textwidth}{c c M{0.27\textwidth} M{0.27\textwidth} M{0.27\textwidth}}
         & & Reference image & With CLIP image conditioning & Without CLIP image conditioning \\
        \makecell[X]{
            \emph{A photo of two pandas walking on a road.}
        } & {\;\;\;} &
        \includegraphics[width=0.27\textwidth]{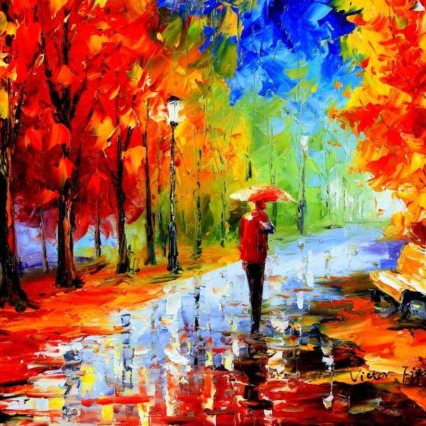} &
        \includegraphics[width=0.27\textwidth]{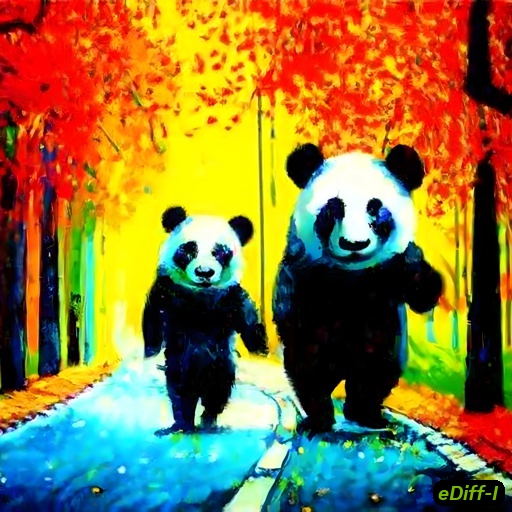} &
        \includegraphics[width=0.27\textwidth]{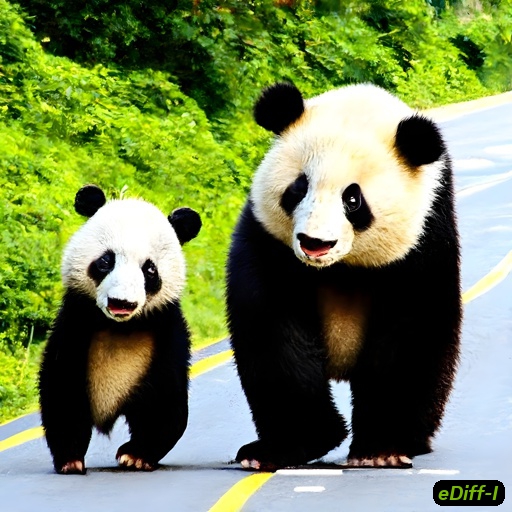} \\

        \makecell[X]{
            \emph{A detailed oil painting of a beautiful rabbit queen wearing a royal gown in a palace. She is looking outside the window, artistic.}
        } & {\;\;\;} &
        \includegraphics[width=0.27\textwidth]{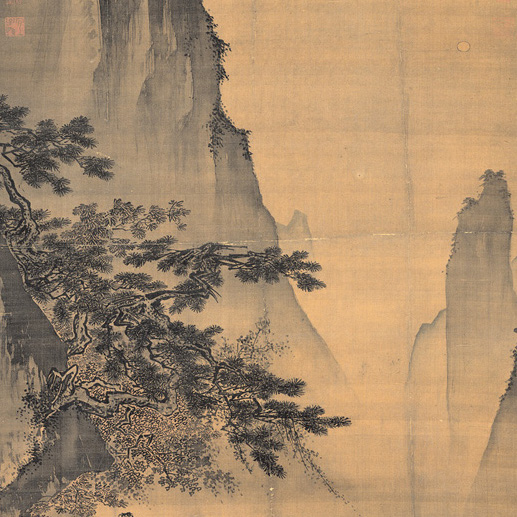} &
        \includegraphics[width=0.27\textwidth]{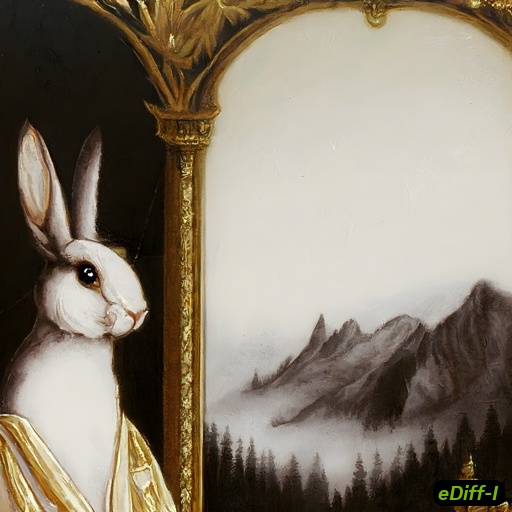} &
        \includegraphics[width=0.27\textwidth]{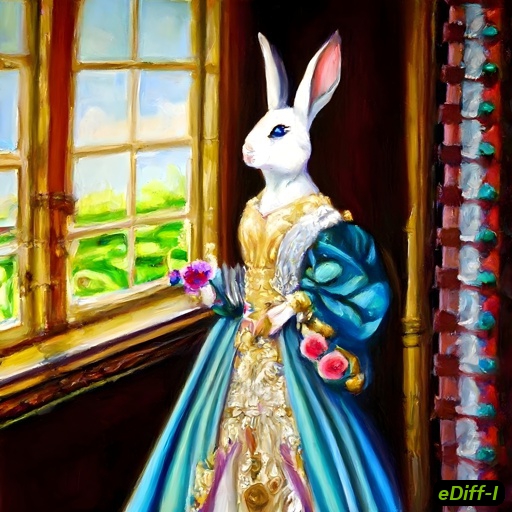} \\

        \makecell[X]{
            \emph{A dslr photo of a dog playing trumpet from the top of a mountain.}
        } & {\;\;\;} &
        \includegraphics[width=0.27\textwidth]{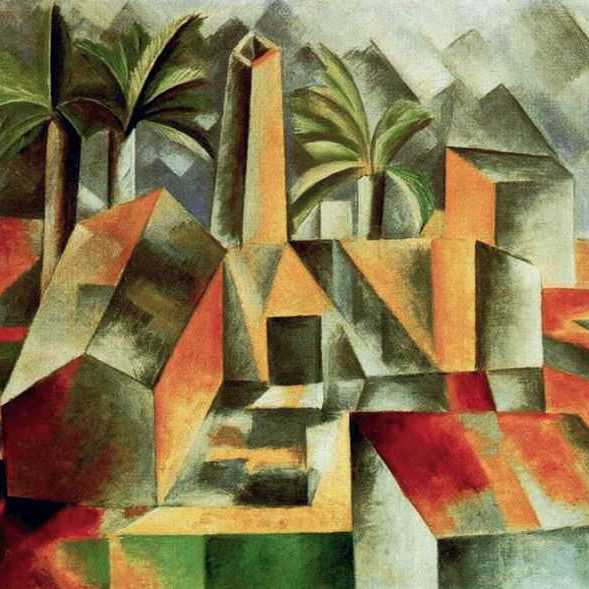} &
        \includegraphics[width=0.27\textwidth]{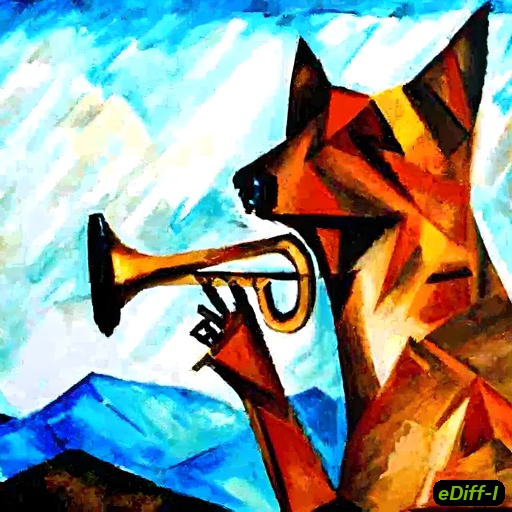} &
        \includegraphics[width=0.27\textwidth]{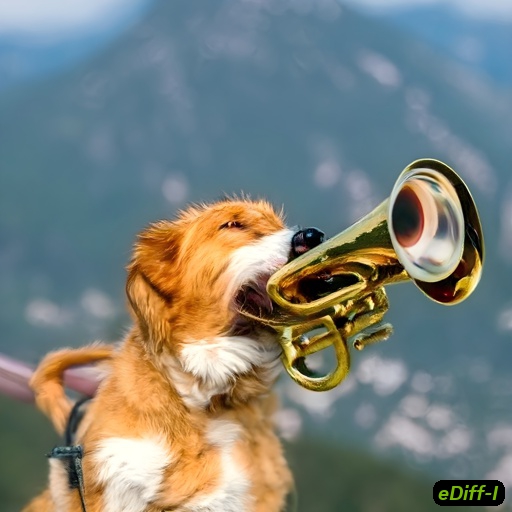} \\

        \makecell[X]{
            \emph{A photo of a teddy bear wearing a casual plain white shirt surfing in the ocean.}
        } & {\;\;\;} &
        \includegraphics[width=0.27\textwidth]{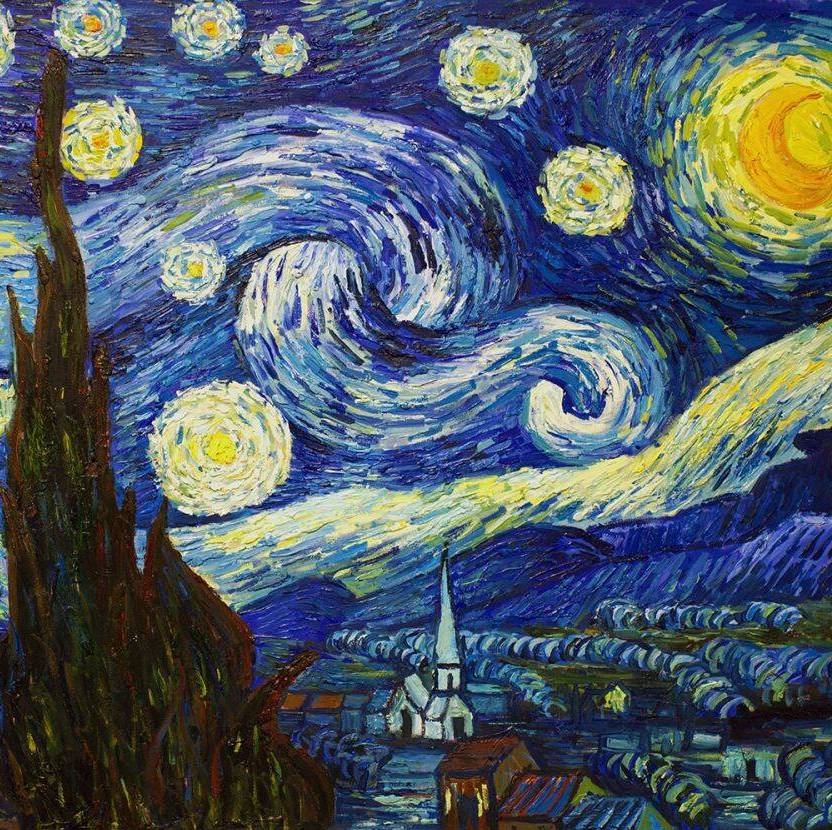} &
        \includegraphics[width=0.27\textwidth]{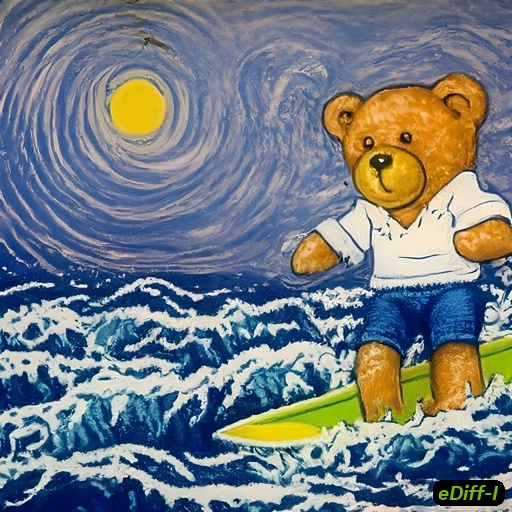} &
        \includegraphics[width=0.27\textwidth]{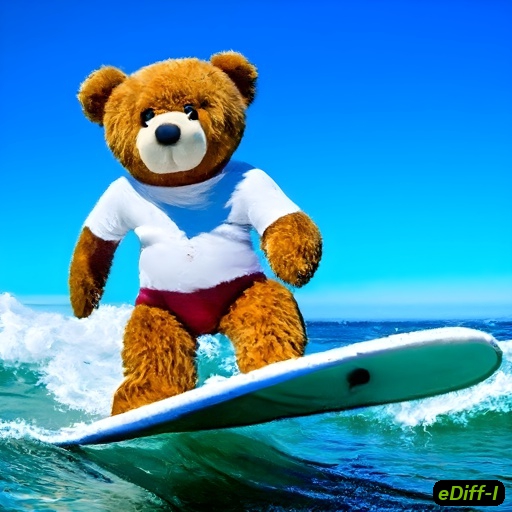} \\

    \end{tabularx}
    \caption{Style transfer with CLIP image conditioning. We use CLIP image embeddings extracted from the reference image shown in the first column to condition our generation. Column 2 shows the generations when CLIP image conditioning is active, while column 3 shows results when CLIP image conditioning is inactive. Our model can effectively synthesize samples consistent with the reference image style.}
    \label{fig:clip_image_conditioning}
\end{figure*}

%% file: figures/paint_with_words_results.tex
\setlength{\tabcolsep}{0.5pt}
\renewcommand{\arraystretch}{0.5}
\begin{figure*}[ht!]
    \newcommand{\sizea}{0.49\textwidth}
    \newcommand{\sizeb}{0.240\textwidth}
    \centering
    
    {
    \setlength{\tabcolsep}{0pt}
    \begin{tabularx}{\sizea}{M{\sizea}}
        \includegraphics[width=\sizeb]{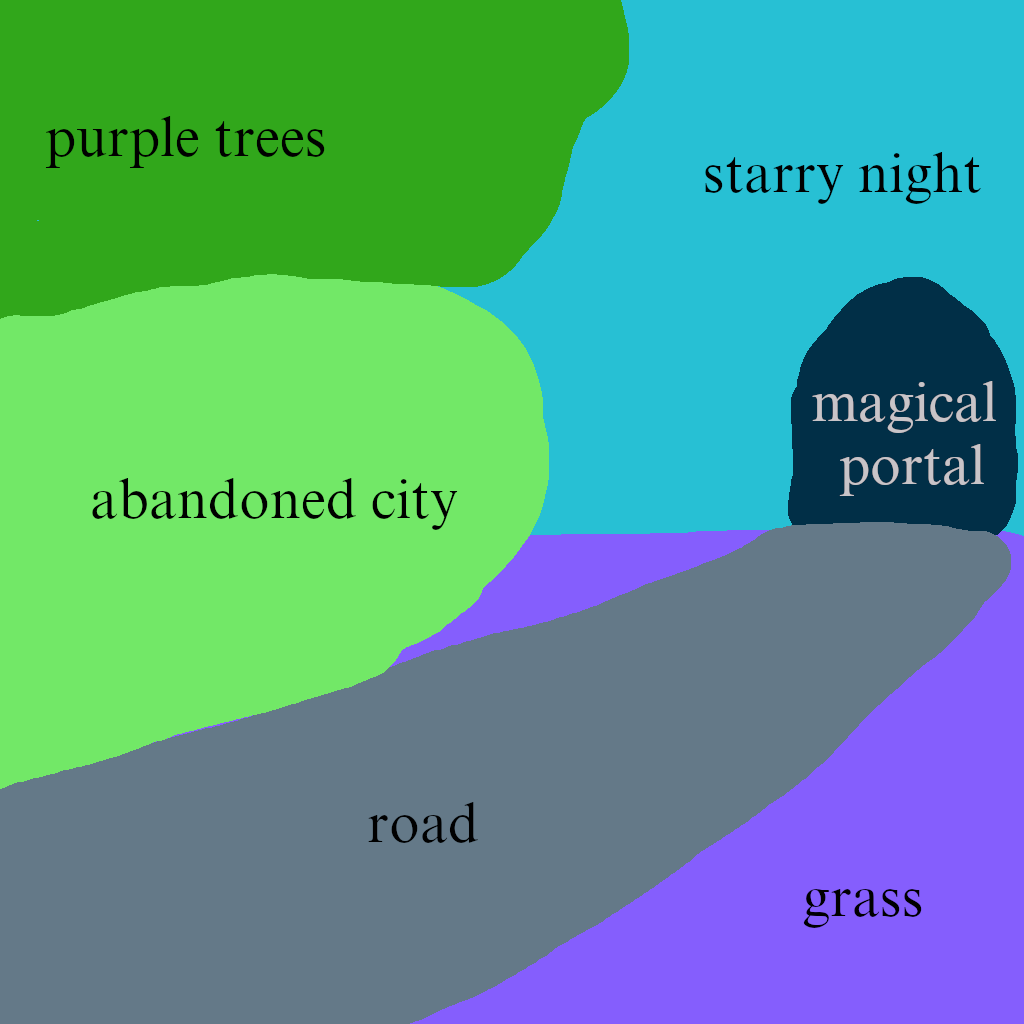} 
        \includegraphics[width=\sizeb]{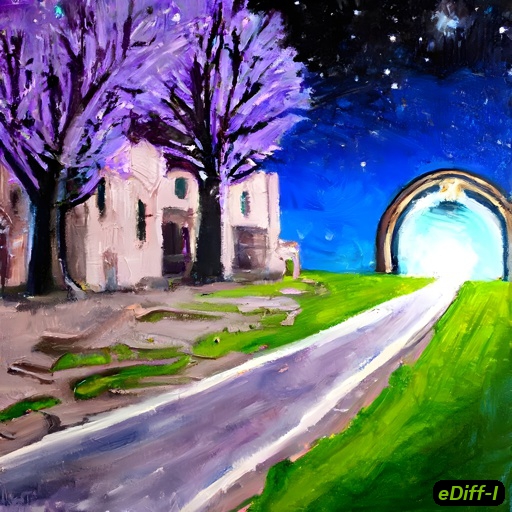} \\
        \emph{ A dramatic oil painting of a road from a magical portal to an abandoned city with purple trees and grass in a starry night.\quad \quad \quad \quad \quad \quad}
    \end{tabularx}
    }    
    {
    \setlength{\tabcolsep}{0pt}
    \begin{tabularx}{\sizea}{M{\sizea}}
        \includegraphics[width=\sizeb]{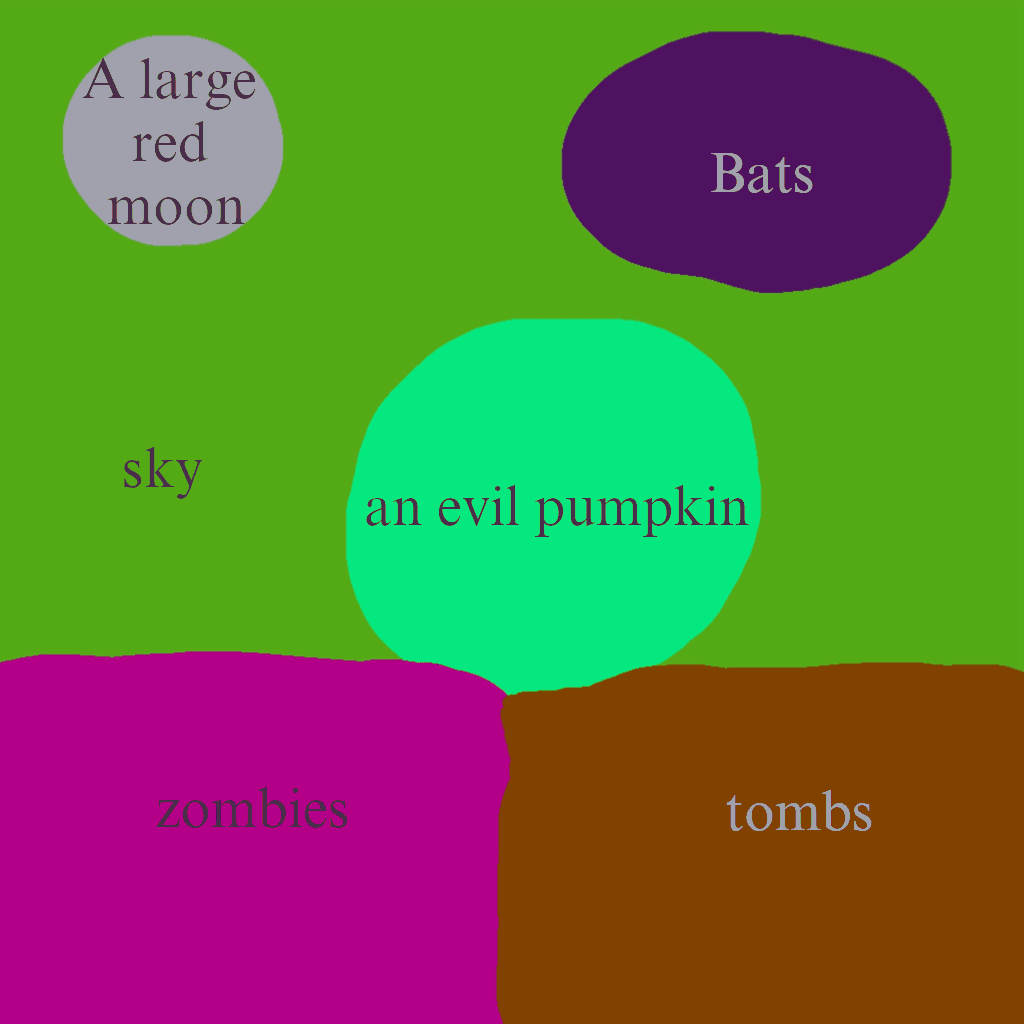} 
        \includegraphics[width=\sizeb]{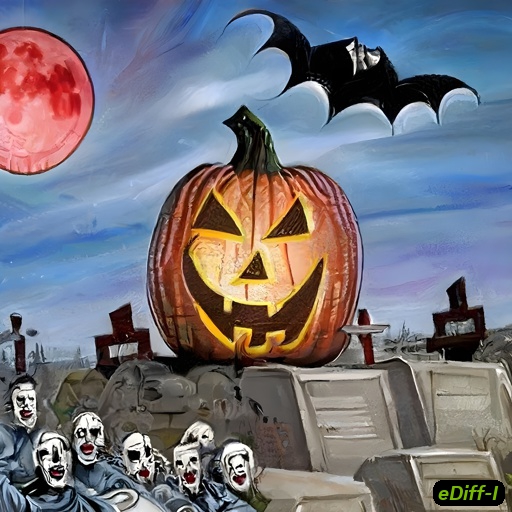} \\
        \emph{ A Halloween scene of an evil pumpkin. A large red moon in the sky. Bats are flying and zombies are walking out of tombs. Highly detailed fantasy art.}
    \end{tabularx}
    }\vspace*{1mm}\\
    {
    \setlength{\tabcolsep}{0pt}
    \begin{tabularx}{\sizea}{M{\sizea}}
        \includegraphics[width=\sizeb]{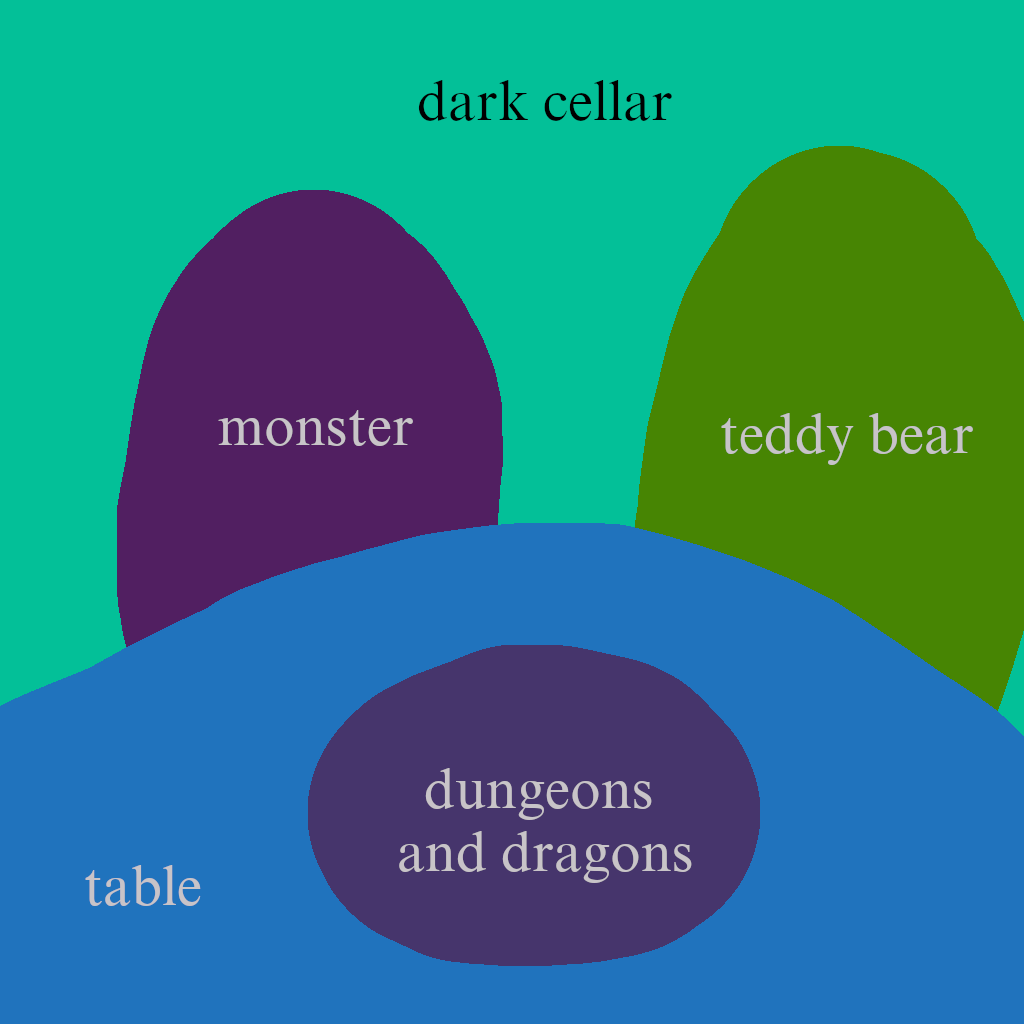} 
        \includegraphics[width=\sizeb]{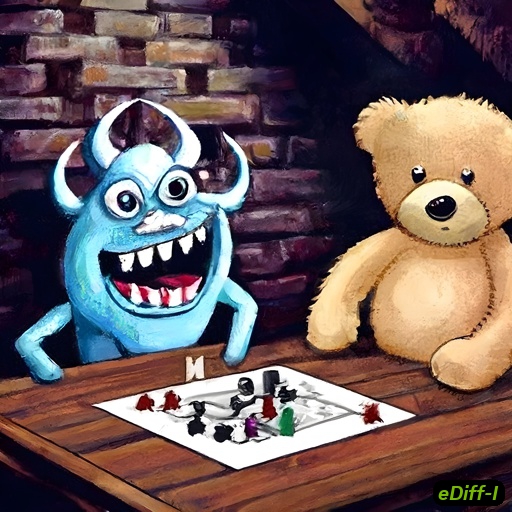} \\
        \emph{ A monster and a teddy bear playing dungeons and dragons around a table in a dark cellar. High quality fantasy art.}
    \end{tabularx}
    }    
    {
    \setlength{\tabcolsep}{0pt}
    \begin{tabularx}{\sizea}{M{\sizea}}
        \includegraphics[width=\sizeb]{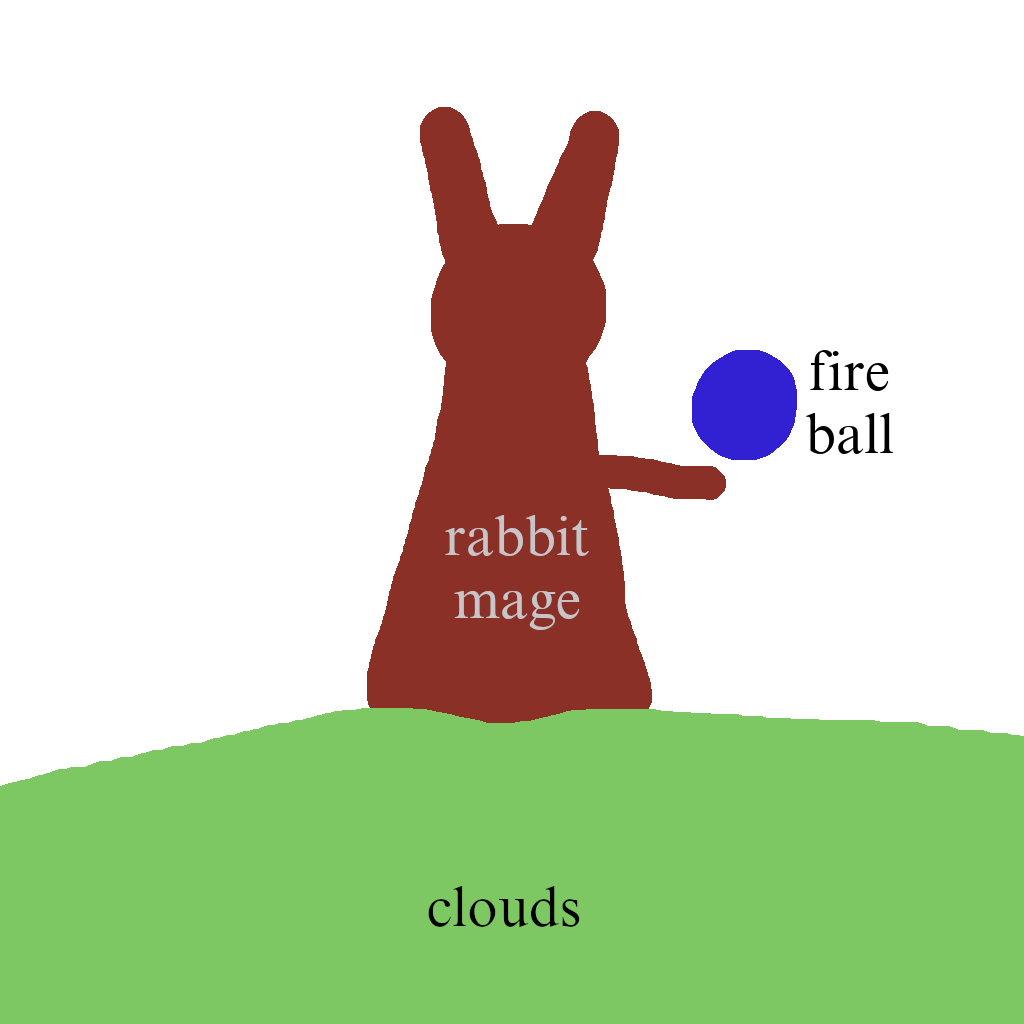} 
        \includegraphics[width=\sizeb]{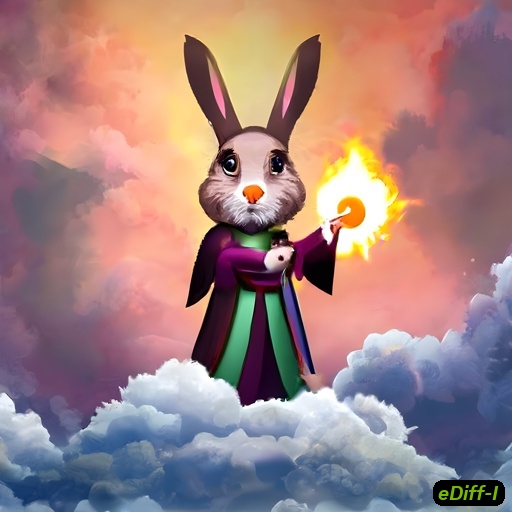} \\
        \emph{ A highly detailed digital art of a rabbit mage standing on clouds casting a fire ball.}
    \end{tabularx}
    }\vspace*{1mm}\\
    {
    \setlength{\tabcolsep}{0pt}
    \begin{tabularx}{\sizea}{M{\sizea}}
        \includegraphics[width=\sizeb]{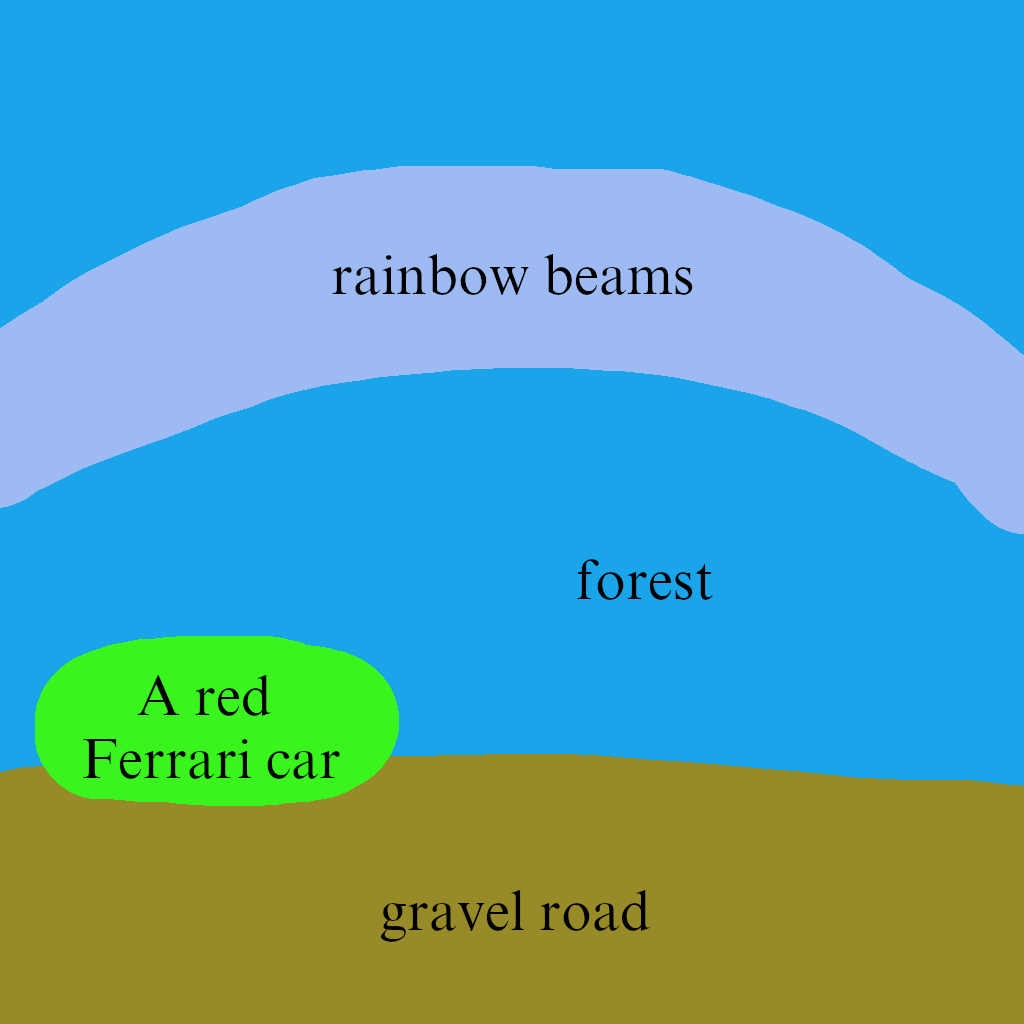} 
        \includegraphics[width=\sizeb]{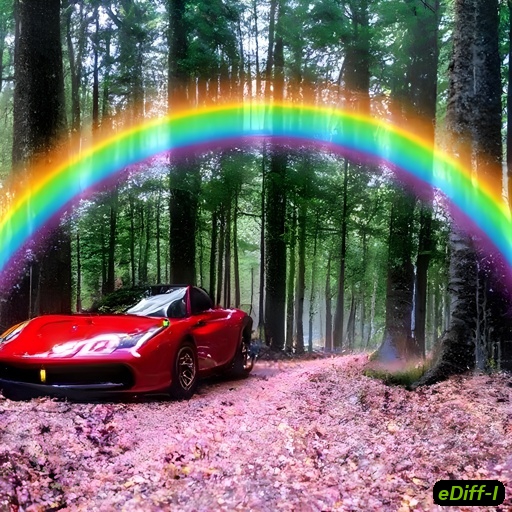} \\
        \emph{ A red Ferrari car driving on a gravel road in a forest with rainbow beams in the distance.}
    \end{tabularx}
    }    
    {
    \setlength{\tabcolsep}{0pt}
    \begin{tabularx}{\sizea}{M{\sizea}}
        \includegraphics[width=\sizeb]{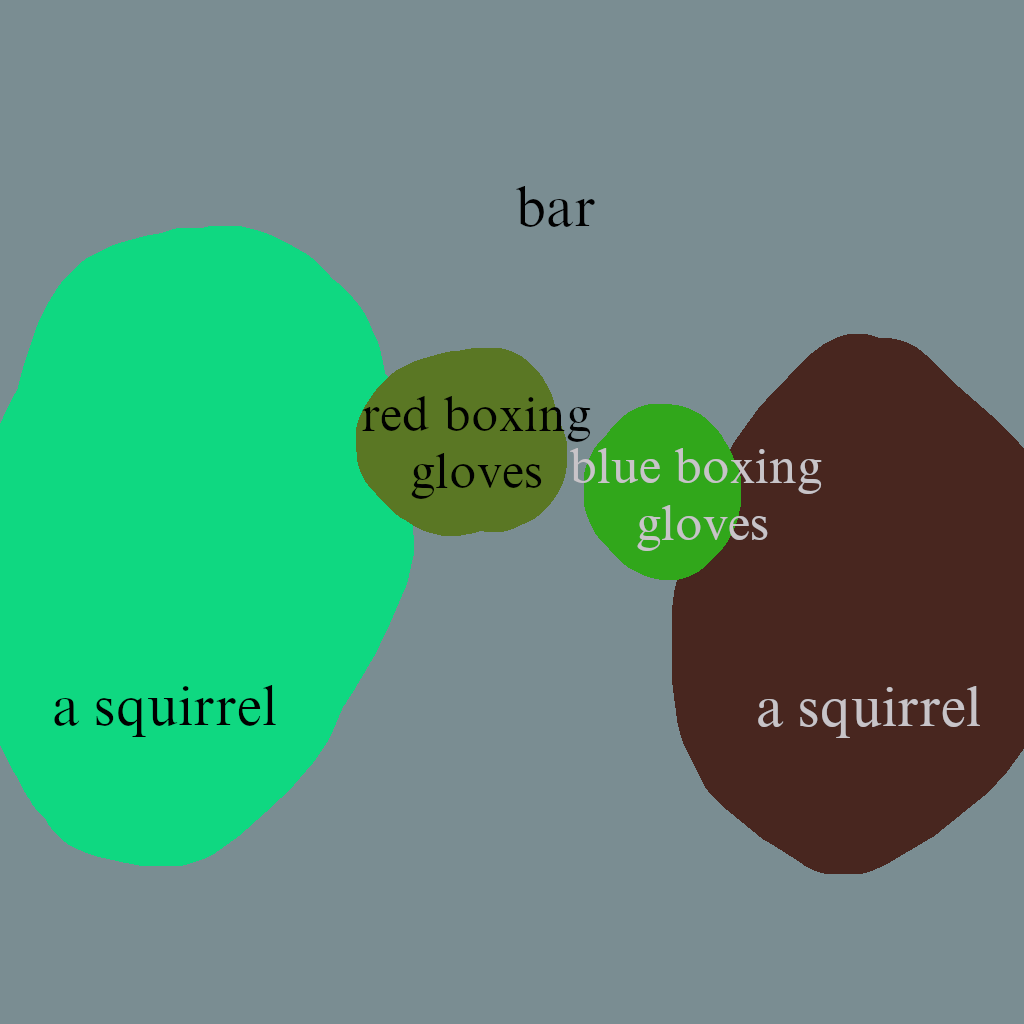} 
        \includegraphics[width=\sizeb]{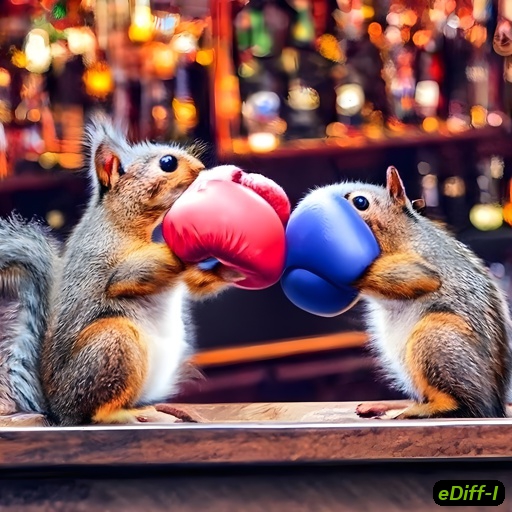} \\
        \emph{ A squirrel with red boxing gloves and a squirrel with blue boxing gloves fighting in a bar.}
    \end{tabularx}
    }\vspace*{1mm}\\
    
    \caption{Samples generated by \pww{}. Our method allows users to control the location of objects mentioned in the text prompt by selecting phrases and scribbling them on the image. Each phrase can contain multiple words. We show two examples in each row. Each example has the text prompt on the bottom, the scribbles on the left, and the output image on the right.}
    \label{fig:paint_with_words_res}
\end{figure*}

%% file: figures/t5_vs_clip.tex
\setlength{\tabcolsep}{0.5pt}
\renewcommand{\arraystretch}{0.5}
\begin{figure*}[ht!]
    \centering
    \begin{tabularx}{\textwidth}{c c M{0.27\textwidth} M{0.27\textwidth} M{0.27\textwidth}}
         & & (a) CLIP only & (b) T5 only & (c) CLIP + T5 \\

         \makecell[X]{
            \emph{A photo of a raccoon wearing a brown sports jacket and a hat. He is holding a whip in his hand.}
        } & {\;\;\;} &
        \includegraphics[width=0.27\textwidth]{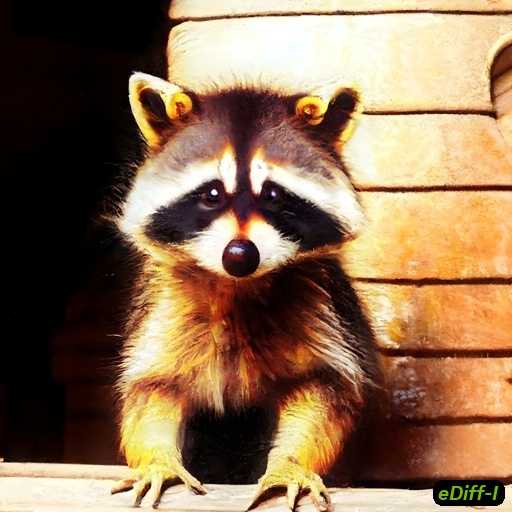} &
        \includegraphics[width=0.27\textwidth]{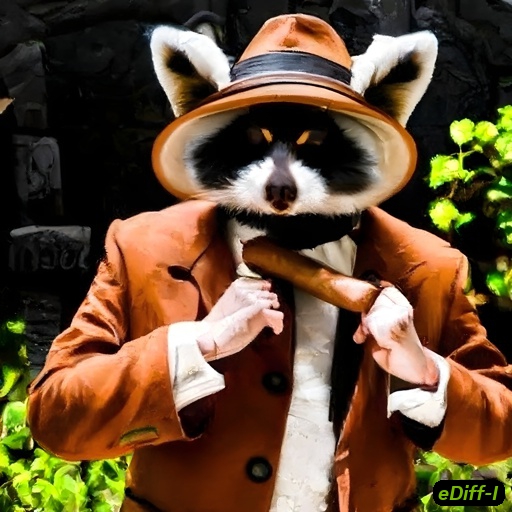} &
        \includegraphics[width=0.27\textwidth]{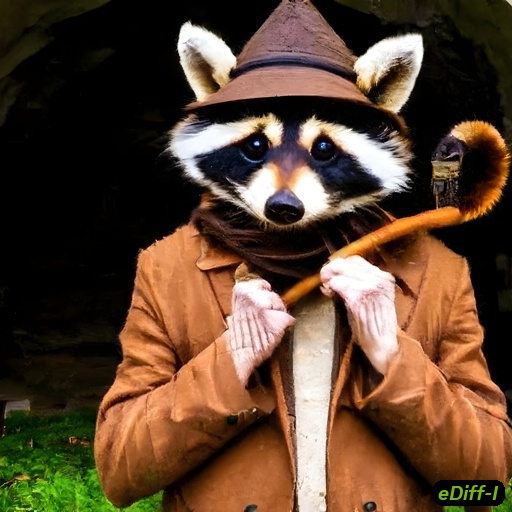} \\

        \makecell[X]{
            \emph{A photo of a red parrot, a blue parrot and a green parrot singing at a concert in front of a microphone. Colorful lights in the background.}
        } & {\;\;\;} &
        \includegraphics[width=0.27\textwidth]{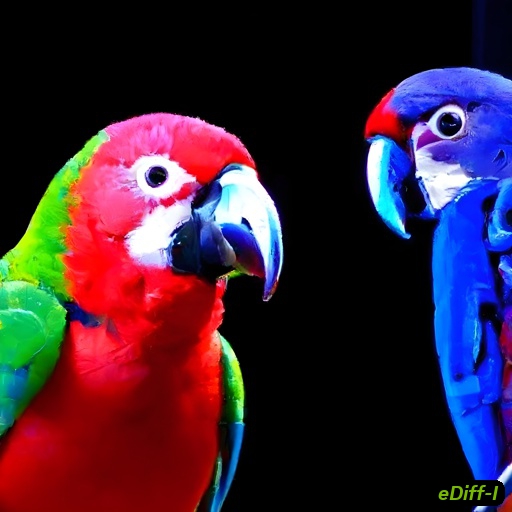} &
        \includegraphics[width=0.27\textwidth]{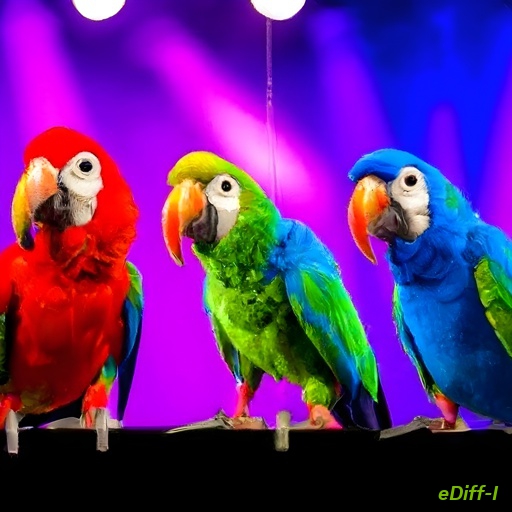} &
        \includegraphics[width=0.27\textwidth]{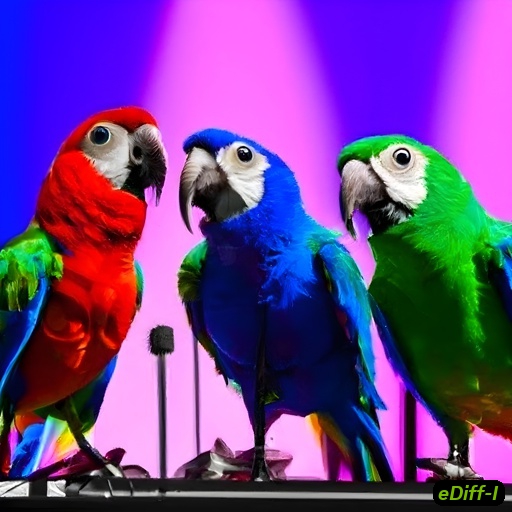} \\

        \makecell[X]{
            \emph{A photo of a cute corgi wearing a beret holding a sign that says ``Diffusion Models''. There is Eiffel tower in the background.}
        } & {\;\;\;} &
        \includegraphics[width=0.27\textwidth]{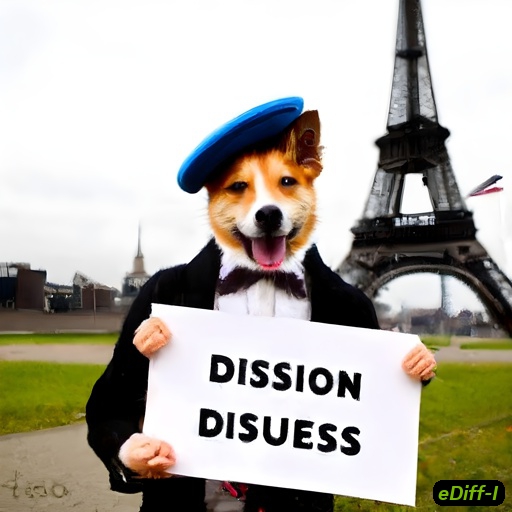} &
        \includegraphics[width=0.27\textwidth]{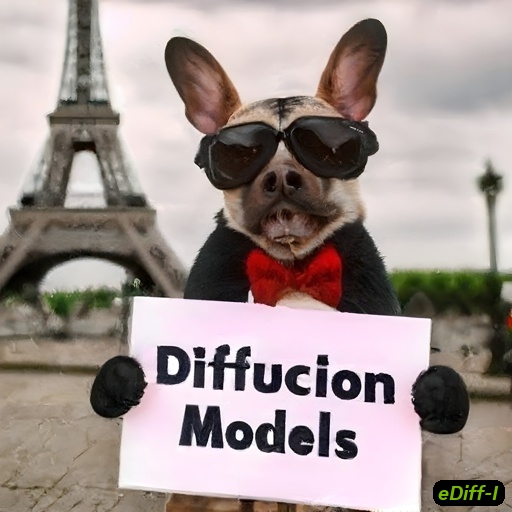} &
        \includegraphics[width=0.27\textwidth]{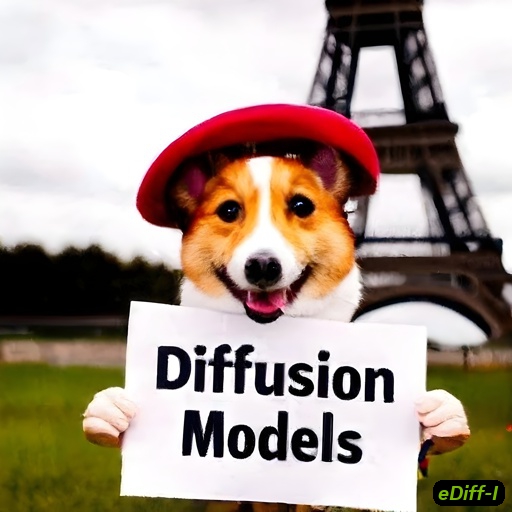} \\

        \makecell[X]{
            \emph{A photo of a lion and a panda competing in the Olympics swimming event.}
        } & {\;\;\;} &
        \includegraphics[width=0.27\textwidth]{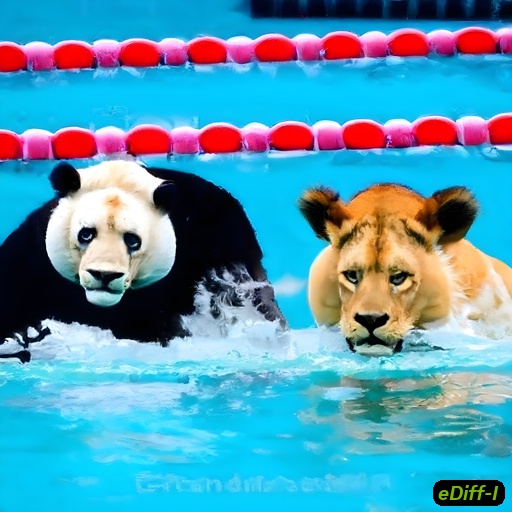} &
        \includegraphics[width=0.27\textwidth]{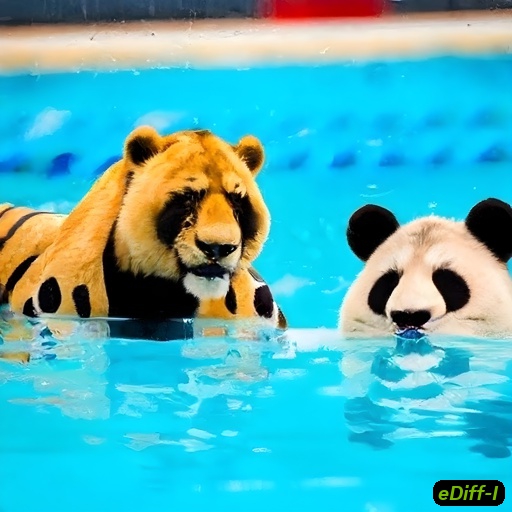} &
        \includegraphics[width=0.27\textwidth]{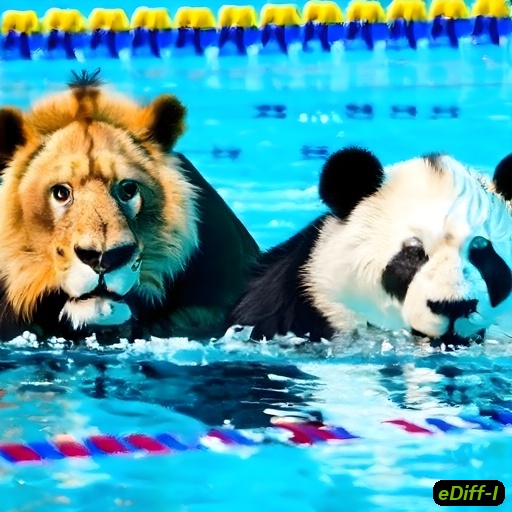} \\
        
    \end{tabularx}
    \vspace{-3mm}
    \caption{Comparison between images generated by our approach using (a) CLIP text encoder alone (our CLIP-text-only setting), (b) T5 text encoder alone (our T5-text-only setting), and (c) Both CLIP and T5 text encoders (our default setting). While images generated in the CLIP-text-only setting often contain correct foreground objects, they tend to miss fine-grain details, such as missing the brown jacket and hat in row 1 and the green parrot and colorful lighting in the background in row 2. Images generated in the T5-text-only setting are of higher quality, but they sometimes contain incorrect objects, such as incorrect dog breed in row 3 and tiger rendering in row 4. Our default setting that combines the strength of the CLIP text encoder and T5 text encoder produces images that better match the input text descriptions and contain the least amount of artifacts.}
    \label{fig:t5_vs_clip}
\end{figure*}

%% file: sections/appendix.tex
\clearpage
\appendix

\section{Network Architecture}\label{sec:network_archtecture}

For our diffusion models, we modify the U-net architecture proposed in Dhariwal~\etal~\cite{dhariwal2021diffusion} with the following changes:
\begin{enumerate}
    \item Global conditioning: We add the projected pooled CLIP text embedding and CLIP image embedding along with the time step embedding in our model. Different from Saharia~\etal~\cite{saharia2022photorealistic}, we do not use pooled T5 embeddings. CLIP text embeddings are trained to be well aligned with images, and hence, using them as global conditioning embeddings are more informative than using T5.
    \item Attention blocks: After every self-attention block in the U-net model of Dhariwal~\etal~\cite{dhariwal2021diffusion}, we add a cross-attention block to perform cross-attention between image embeddings and the conditioning embeddings. The keys in the cross-attention layers are the concatenation of pre-pooled CLIP text embeddings ($77$ tokens), T5 embeddings ($113$ tokens), and pooled CLIP image embedding ($1$ token). In addition to these, we also add a learnable null embedding, which the model can attend to when it does not need to use any of the conditioning embeddings.
\end{enumerate}

In addition, to make the super-resolution models more efficient during training and inference, we use the block structure of Efficient U-net architecture proposed in Saharia~\etal~\cite{saharia2022photorealistic}. Following the prior works~\cite{saharia2022photorealistic, ramesh2022hierarchical}, we train the SR1024 model using random patches of size $256{\times}256$ during training and apply it on $1024{\times}1024$ resolution during inference. We also remove the self-attention layers and only have the cross-attention layers in this network, as computing self-attention during inference is very expensive. The U-net configurations we use for all our models are provided in Tables~\ref{tab:arch_base}, \ref{tab:arch_sr_256}, and \ref{tab:arch_sr_1024} respectively.

\newpage
\input{tables/architecture_base}
\input{tables/architecture_sr_256}
\input{tables/architecure_sr_1024}

\clearpage
\section{Ensemble training schedule}\label{sec:hyperparams}
\label{sec:app_branching}
As discussed in Sec~\ref{sec:body_branching}, we use a binary-tree-based branching strategy for training our ensemble model. In Tables~\ref{tab:training_progress} and \ref{tab:training_progress_sr}, we list the exact training schedule we use to train our models. Each entry in the configuration is a tuple containing $E^{level}_{id}$ in the binary tree. This corresponds to the training out model with the noise distribution $p^{\text{level id}}_{\text{interval id}}(\sigma)$. Models denoted as $M^{C}$ are the intermediate noise models. For these models, the configuration with a negative sign indicates all noise levels other than the one indicated. For instance, $-(9, 511)$ denotes all noise levels other than the one included in $p^{9}_{511}(\sigma)$. That is the noise is sampled from the complementary distribution $p(\sigma) \symbol{92} p^{\text{9}}_{\text{511}}(\sigma)$.

\input{tables/training_progress}

\newpage
\subsection{Hyper-parameters}
The hyperparameters we use for training all our models are provided in Table \ref{tab:hyperparams}.

\input{tables/hyperparameters}

%% file: tables/architecture_base.tex
\begin{table}[ht!]
    \centering
    \caption{Base model architecture}
    \label{tab:arch_base}
    \setlength{\extrarowheight}{10pt}
    \begin{tabularx}{0.5\textwidth}{Y Y}
        \toprule
        Channel mult & $[1, 2, 4, 4]$ \\
        Dropout & $0$ \\
        Number of channels & $256$ \\
        Number of residual blocks & $3$ \\
        Self attention resolutions & $[32, 16, 8]$ \\
        Cross attention resolutions & $[32, 16, 8]$ \\
        Use scale shift norm & True \\
        \bottomrule
    \end{tabularx}
\end{table}

%% file: tables/architecture_sr_256.tex
\begin{table}[ht!]
    \centering
    \caption{SR256 model architecture}
    \label{tab:arch_sr_256}
    \setlength{\extrarowheight}{10pt}
    \begin{tabularx}{0.5\textwidth}{Y Y}
        \toprule
        Channel multiplier & $[1, 2, 4, 8]$ \\
        Block multiplier & $[1, 2, 4, 4]$ \\
        Dropout & $0$ \\
        Number of channels & $128$ \\
        Number of res blocks & $2$ \\
        Self attention resolutions & $[32]$ \\
        Cross attention resolutions & $[32]$ \\
        Use scale shift norm & True \\
        \bottomrule
    \end{tabularx}

\end{table}

%% file: tables/architecure_sr_1024.tex
\begin{table}[ht!]
    \centering
    \caption{SR1024 model architecture}
    \label{tab:arch_sr_1024}
    \setlength{\extrarowheight}{10pt}
    \begin{tabularx}{0.5\textwidth}{Y Y}
        \toprule
        Patch size & $256 \times 256$\\
        Channel multiplier & $[1, 2, 4, 4]$ \\
        Block multiplier & $[1, 2, 4, 4]$ \\
        Dropout & $0$ \\
        Number of channels & $128$ \\
        Number of res blocks & $2$ \\
        Cross attention resolutions & $[32]$\\
        Use scale shift norm & True \\
        \bottomrule
    \end{tabularx}

\end{table}

%% file: tables/training_progress.tex
\begin{table}[hbt!]
    \centering
    \caption{Training schedule for the base model. Each configuration is a tuple of (level id, interval id) in the binary tree. This configuration denotes the model trained with the noise distribution $p^{\text{level id}}_{\text{interval id}}(\sigma)$. The configuration denoted with a negative sign stands for complementary distribution i.e., $p(\sigma) \symbol{92} p^{\text{level id}}_{\text{interval id}}(\sigma)$. The third column denotes the model from which the current model is initialized. The fourth column denotes the number of iterations we train our model.}
    \label{tab:training_progress}

    \setlength{\extrarowheight}{10pt}
    \begin{tabularx}{0.5\textwidth}{Y Y Y Y}
        \toprule
        Model id & Config & Initialized from & \# iterations \\
        \midrule
        $M(0, 0)$ & $(0, 0)$ & $-$ & $500K$ \\
        $M(1, 0)$ & $(1, 0)$ & $M(0, 0)$ & $50K$ \\
        $M(1, 1)$ & $(1, 1)$ & $M(0, 0)$ & $50K$ \\
        $M(2, 0)$ & $(2, 0)$ & $M(1, 0)$ & $120K$ \\
        $M(2, 1)$ & $(2, 1)$ & $M(1, 0)$ & $50K$ \\
        $M(2, 2)$ & $(2, 2)$ & $M(1, 1)$ & $50K$ \\
        $M(2, 3)$ & $(2, 3)$ & $M(1, 1)$ & $130K$ \\
        $M(3, 0)$ & $(3, 0)$ & $M(2, 0)$ & $160K$ \\
        $M(3, 7)$ & $(3, 7)$ & $M(3, 0)$ & $40K$ \\
        $M(4, 0)$ & $(4, 0)$ & $M(3, 0)$ & $110K$ \\
        $M(4, 15)$ & $(4, 15)$ & $M(3, 7)$ & $190K$ \\
        $M(5, 31)$ & $(5, 31)$ & $M(4, 15)$ & $100K$ \\
        $M(9, 511)$ & $(9, 511)$ & $M(5, 31)$ & $50K$ \\
        \midrule
        \multirow{2}{*}{$M^{C}(9, 511)$} & \multirow{2}{*}{-$(9, 511)$} & Base model & \multirow{2}{*}{$50K$} \\
        & & at $1.45M$ & \\
        \multirow{2}{*}{$M^{C}_{2}(9, 511)$} & -$(9, 511)$ & Base model & \multirow{2}{*}{$50K$} \\
        & -$(4, 0)$ & at $1.45M$ & \\
        \bottomrule
    \end{tabularx}

\end{table}

\begin{table}[hbt!]
    \centering
    \caption{Training schedule for the super-resolution models. Each configuration is a tuple of (level id, interval id) in the binary tree. The third column denotes the model from which the current model is initialized. The fourth column denotes the number of iterations we train our model.}
    \label{tab:training_progress_sr}

    \setlength{\extrarowheight}{10pt}
    \begin{tabularx}{0.5\textwidth}{Y Y Y Y}
        \toprule
        Model id & Config & Initialized from & \# iterations \\
        \midrule
        $M(0, 0)$ & $(0, 0)$ & $-$ & $2M$ \\
        $M(1, 0)$ & $(1, 0)$ & $M(0, 0)$ & $300K$ \\
        $M(1, 1)$ & $(1, 1)$ & $M(0, 0)$ & $300K$ \\
        \bottomrule
    \end{tabularx}

\end{table}

%% file: tables/hyperparameters.tex
\begin{table}[htb!]
    \centering
    \caption{Hyperparameters}
    \label{tab:hyperparams}
    \setlength{\extrarowheight}{10pt}
    \begin{tabularx}{0.5\textwidth}{X r}
        \toprule
        Optimizer & AdamW \\
        Learning rate & $0.0001$ \\
        Weight decay & $0.01$ \\
        Betas & (0.9, 0.999) \\
        EMA & $0.9999$ \\
        CLIP text embedding dropout rate & $0.2$ \\
        T5 text embedding dropout rate & $0.25$ \\
        CLIP image embedding dropout rate & $0.9$ \\
        Gradient checkpointing & Enabled \\
        Number of iterations for base model & $1.9M$ \\
        Number of iterations for SR256 model & $2M$ \\
        Number of iterations for SR1024 & $1.7M$ \\
        \multirow{2}{*}{}Sampler for base model & DEIS~\cite{zhang2022fast}, 3kutta \\
        & Order 6, 25 steps \\
        \multirow{2}{*}{}Sampler for super-resolution models & DEIS, 3kutta \\
        & Order 3, 10 steps \\
        
        \bottomrule
    \end{tabularx}
\end{table}